\documentclass{article} %
\usepackage{iclr2027_conference}
\usepackage{times}

\usepackage{amsmath,amsfonts,bm}

\def\eqref#1{equation~\ref{#1}}

\def\1{\bm{1}}

\def\vb{{\bm{b}}}

\DeclareMathAlphabet{\mathsfit}{\encodingdefault}{\sfdefault}{m}{sl}
\SetMathAlphabet{\mathsfit}{bold}{\encodingdefault}{\sfdefault}{bx}{n}

\usepackage{graphicx}
\usepackage{booktabs}
\usepackage{placeins}
\usepackage{amsmath,amssymb}
\usepackage{physics}   %
\usepackage{multirow}
\usepackage{nicefrac}   %
\usepackage{float}
\usepackage{xcolor}
\usepackage{titletoc}
\usepackage{xparse}
\usepackage{etoolbox}
\usepackage{silence}
\usepackage{hyperref} %
\usepackage{url}
\hypersetup{
  colorlinks=true, linkcolor=red, citecolor=blue, urlcolor=magenta,
  pdftitle={Mind the Spike: Mechanisms and Brittleness of Visual Massive Activations in Large Vision-Language Models},
  pdfauthor={Jonas Ngnawé, Yann Pequignot, Sabyasachi Sahoo, Christian Gagné, Frédéric Precioso, Sanmi Koyejo}
}
\makeatletter
\newcommand{\backanchor}[1]{%
\ifcsname backfrom@#1\endcsname\else
  \label{back:#1}\expandafter\gdef\csname backfrom@#1\endcsname{}%
\fi}
\DeclareRobustCommand{\aref}[1]{Appendix~\ref{#1}\backanchor{#1}}   %
\DeclareRobustCommand{\tref}[1]{Table~\ref{#1}\backanchor{#1}}      %
\DeclareRobustCommand{\fref}[1]{Figure~\ref{#1}\backanchor{#1}}     %
\makeatother
\makeatletter
\NewDocumentCommand{\backto}{s m}{%
  \IfBooleanTF{#1}{%
    \par\noindent
    \ifcsname r@back:#2\endcsname
      \hfill{\footnotesize\hyperref[back:#2]{$\hookleftarrow$~back to p.~\pageref{back:#2}}}%
    \fi
    \par\noindent
  }{%
    \ifcsname r@back:#2\endcsname
      \hfill{\footnotesize\hyperref[back:#2]{$\hookleftarrow$~back to p.~\pageref{back:#2}}}%
    \fi
  }%
}
\makeatother

\NewDocumentEnvironment{appfigure}{}{\begin{figure}[t]}{\end{figure}}
\NewDocumentEnvironment{apptable}{}{\begin{table}[t]}{\end{table}}
\definecolor{lens}{RGB}{31,90,160} %
\definecolor{spikec}{HTML}{D93025}\definecolor{sharedc}{HTML}{2A78D6}\definecolor{leastc}{HTML}{6B7076} %
\newcommand{\decodeimg}[1]{\raisebox{\dimexpr-\height+\ht\strutbox\relax}{\includegraphics[width=2.3cm,height=2.3cm,keepaspectratio,page=#1]{figures/fig_decode_tokens.pdf}}} %

\graphicspath{{figures/}}

\title{Mind the Spike: Mechanisms and Brittleness\\ of Visual Massive Activations in Large\\ Vision--Language Models}

\author{%
  Jonas Ngnawé\textsuperscript{1,2,5}\thanks{Corresponding author: \texttt{jonas.ngnawe.1@ulaval.ca}} \quad
  Yann Pequignot\textsuperscript{1,2} \quad
  Sabyasachi Sahoo\textsuperscript{1,2} \\
  \bf Christian Gagné\textsuperscript{1,2,3} \quad
  Frédéric Precioso\textsuperscript{4} \quad
  Sanmi Koyejo\textsuperscript{5} \\[0.5em]
  \textsuperscript{1}Université Laval (IID) \quad
  \textsuperscript{2}Mila -- Quebec AI Institute \quad
  \textsuperscript{3}Canada CIFAR AI Chair \\
  \textsuperscript{4}Université Côte d'Azur, CNRS, Inria, I3S, Maasai \quad
  \textsuperscript{5}Stanford University
}

\iclrfinalcopy %

\begin{document}
\maketitle
\lhead{Preprint}

\begin{abstract}
Large vision--language models (LVLMs) inherit massive activations from their text-only bases: spikes where a few fixed hidden channels receive values thousands of times above the typical magnitude.
The text spike systematically appears in early layers at a fixed initial position, independently of input content.
Visual spikes vary across images, but whether their formation follows a consistent pattern across LVLMs and how they respond to image perturbations remain open questions.
We find that some LVLMs do not form visual spikes, while others spike at different rates, typically in deeper layers.
We identify the trigger direction from model weights and an interpretable location rule: before the language model decoder runs, eventual spike tokens are largely restricted to those sharing least with the rest of the image.
Crucially, visual spikes are strikingly brittle.
Common corruptions frequently create and relocate spikes, and less often remove them, raising overall incidence.
Our trigger-guided spike attack deliberately creates or removes spikes under a small $\ell_\infty$ budget, with $\nicefrac{1}{255}$ enough in nine of the ten models that spike.
Finally, our preventive intervention removes only the trigger component before spikes erupt, eliminating or substantially reducing spikes on clean and perturbed images while leaving the other image tokens nearly unchanged.
Our study spans 25 adapter-based LVLMs built on 18 released text-only bases from 10 families, ranging from 2B to 72B parameters.
\end{abstract}

\suppressfloats[t]

\section{Introduction}

Large language models exhibit idiosyncratic internal phenomena that arise during pretraining and persist across scales and adaptations, such as massive activations \citep{sun2024massive, yu2024super, gu2025when}. A massive activation is a spike on a few fixed hidden channels, with values thousands of times above the typical magnitude.
In most modern text-only language models, it appears on every input at the first token or an early delimiter.
It is written abruptly in the first few blocks and persists until the last layers, a pattern observed across models \citep{sun2024massive, gu2025when, su2026survey}.
Recent work links its formation to alignment with a high-gain direction at a feed-forward block's input \citep{sun2026spike}.
The spike can act as an attention bias, making its token an attention sink, a token that attracts an unusually large share of attention while not contributing much toward the output \citep{sun2024massive, xiao2023streamingllm, clark2019what}.

Large vision--language models (LVLMs) built by adapting these language models retain the inherited spike channels, and image tokens can become massive on the same dimensions \citep{kang2025see, su2026survey}.
Visual spikes, however, are different from text spikes.
Their location varies from one image to another, usually lies in background regions, and is independent of any text prompt following the image \citep{kang2025see}.
Observations on a small set of LVLMs place their formation in the first few decoder layers \citep{kang2025see, choi2026sinks}, and current accounts take the visual mechanism to mirror the text spike's \citep{kang2025see, su2026survey}.
Do visual spikes form in all LVLMs and on every image?
Where in the model do they form, how are they triggered, and what distinguishes the eventual spike token before it becomes massive?
More crucially, how do visual spikes respond to image perturbations?

Understanding visual spikes has practical implications. Attention sinks are already exploited in long-context and streaming inference \citep{xiao2023streamingllm, jolicoeur2026sliding}, as well as in quantization and KV-cache compression \citep{su2026survey}.
In LVLMs, visual attention sinks inform KV caching, token pruning, attention redistribution, and watermarking \citep{ning2025livevlm, xu2025streamingvlm, kang2025see, choi2026sinks, li2026sinkpruner, chistyakova2025activemark}.
These applications depend on whether the spike forms at all in a model, where it occurs, and how it changes under image perturbations.
Massive values can also dominate norms, distances, and similarities in hidden representations, affecting analyses of projector geometry \citep{li2025lost}, token sparsity \citep{fan2026encode}, concept directions \citep{ravfogel2020null, belrose2023leace}, and directions associated with refusal in safety analyses \citep{arditi2024refusal}.
Changes in the spike can therefore affect these analyses without substantially changing the model's answer.

\begin{figure}[t]
\vspace{-25pt}
\centering
\includegraphics[width=\textwidth]{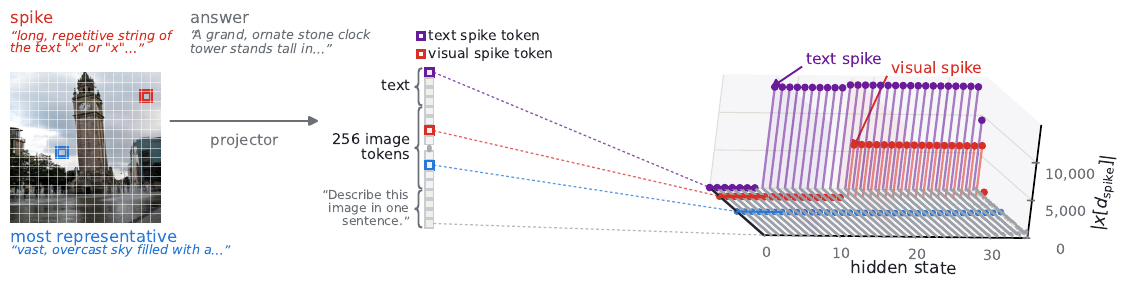}
\vspace{-15pt}
\caption{\textbf{Visual spikes share the text spike's channels but differ in when and where they form.}
An image processed by Qwen3-VL-8B illustrates this distinction.
Red and blue mark two sky patches and their corresponding image tokens.
Right, each token's activation magnitude on the inherited spike channel, $|x[d_{\mathrm{spike}}]|$, across hidden states.
The text token (purple) becomes massive early, at block~6.
The red image token initially has an ordinary activation, which abruptly rises at block~16, making it a \emph{visual spike token}.
Although both tokens come from the sky background, the blue token decodes to sky and remains non-massive, while the red token decodes to a repetitive pattern and later spikes.
}
\label{fig:one}
\vspace{-15pt}
\end{figure}

\begin{figure}[t]
\centering
\vspace{-25pt}
\includegraphics[width=\textwidth]{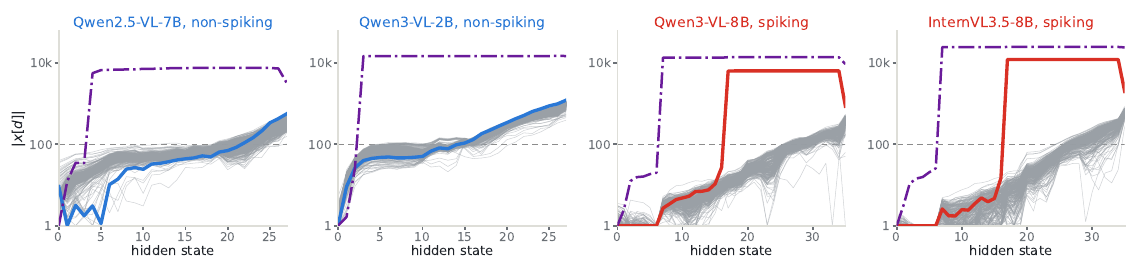}
\vspace{-10pt}
\caption{\textbf{A visual switch writes the spike in only some LVLMs.}
The spike-channel magnitude $|x[d_{\mathrm{spike}}]|$ across hidden states on a log scale for four LVLMs, one image each. Grey lines are the image tokens, the dash-dot line the text spike token. Blue is the largest image token in two non-spiking models, and red the spike token in two spiking ones. \fref{fig:switchall} shows all 25 LVLMs.}
\label{fig:switchfour}
\end{figure}

\begin{figure}[t]
\centering
\vspace{-25pt}
\includegraphics[width=\textwidth]{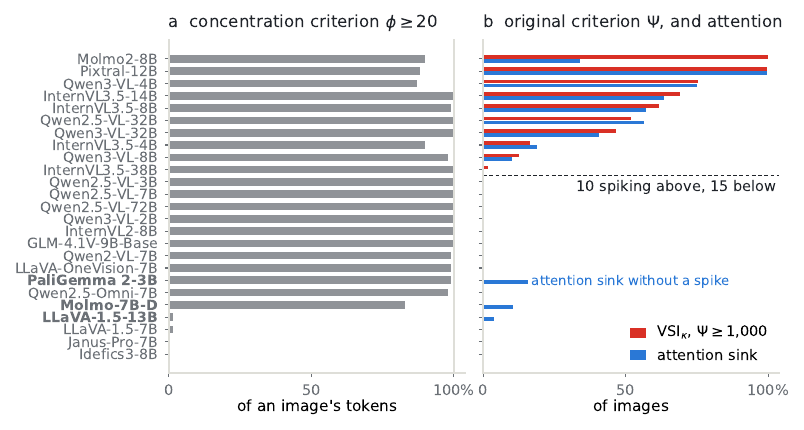}
\vspace{-20pt}
\caption{\textbf{The common definition flags most image tokens in nearly every LVLM, the original flags spikes in only ten, and attention sinks largely track the original.}
Left, the mean share of image tokens passing $\phi \geq 20$ over 300 images for each model.
For 23 of the 25 models, every image has a token passing this threshold.
Right, $\mathrm{VSI}_\kappa$ at $\kappa = 1{,}000$, the fraction of images that spike under the original massive-activation criterion, alongside independently measured attention-sink incidence. Ten models spike and fifteen do not.
The numerical values, confidence intervals, and related measurements are in Tables~\ref{tab:fullroster} and~\ref{tab:regimes}\backanchor{tab:fullroster}\backanchor{tab:regimes} and Appendices~\ref{app:models} and~\ref{app:census}\backanchor{app:models}\backanchor{app:census}.}
\label{fig:crit}
\vspace{-10pt}
\end{figure}

\begin{figure}[t]
\centering
\vspace{-25pt}
\begin{minipage}[t]{0.30\linewidth}
\centering
{\footnotesize\textbf{Predicting the spike token}}\par\smallskip
\includegraphics[width=\linewidth]{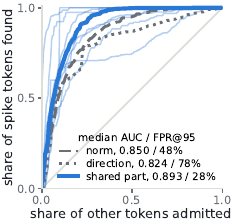}
\end{minipage}\hspace{0.04\linewidth}
\begin{minipage}[t]{0.40\linewidth}
\centering
{\footnotesize\textbf{Relocating through 10\% of the tokens}}\par\smallskip
\includegraphics[width=\linewidth]{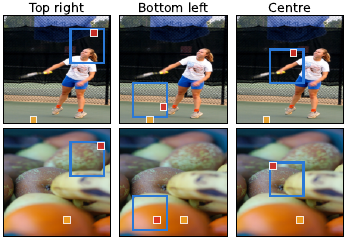}
\end{minipage}
\vspace{-4pt}
\caption{\textbf{Eventual spike tokens lie among the least-shared tokens, and editing a small region can relocate the spike.}
Left, for each spike token, the share of its image's other tokens ranked above it against the share of spike tokens found, with the shared part, the norm or the direction at the decoder's input. Thin blue lines are the shared part in each of the ten spiking models, bold lines the medians over models, and the legend gives the median AUC / FPR@95.
Right, two images under InternVL3.5-8B with the same $5\times5$ block placed at three locations.
The block covers $25/256\approx10\%$ of image tokens, whose decoder-input vectors are scaled to a tenth.
Red marks the resulting spike, amber the clean spike, and blue outlines the edited block.
Tables~\ref{tab:decile} and \ref{tab:locator} give the location statistics; \fref{fig:positions} and Tables~\ref{tab:steer} and \ref{tab:blocks} give further relocation results.}
\label{fig:loc}
\vspace{-10pt}
\end{figure}

\begin{figure}[t]
\centering
\vspace{-25pt}
\includegraphics[width=\textwidth]{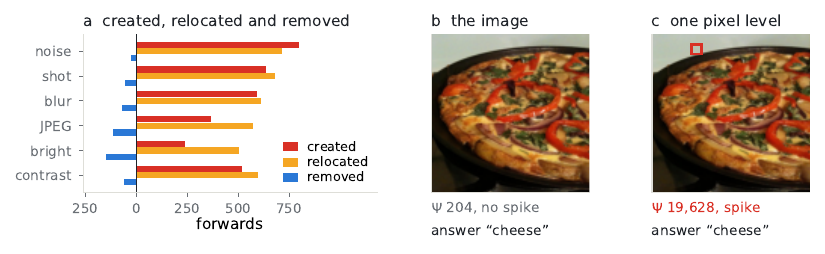}
\vspace{-22pt}
\caption{\textbf{Visual spikes are brittle to image changes, even when the answer stays the same.}
(a) Across ten spiking models and six corruptions, with 200 questions per model and operator, the forwards on which the corruption creates a spike, removes one, or relocates it, the last among forwards that spike both clean and corrupted, on a different image token.
(b, c) One VQAv2 image under InternVL3.5-14B: a trigger-guided perturbation bounded by $\nicefrac{1}{255}$ raises $\Psi_{\mathrm{peak}}$ from $204$ to $19{,}628$ without changing the answer. Red marks the spike token.
Across models, the attack creates or removes spikes at $\nicefrac{1}{255}$ in nine of the ten spiking LVLMs, with yes/no answers unchanged on $93$ to $100\%$ of attempts.
Tables~\ref{tab:corrpanel} and \ref{tab:epssweep} give the corruption results and the full attack sweep.}
\label{fig:britt}
\backanchor{tab:corrpanel}\backanchor{tab:epssweep}
\vspace{-10pt}
\end{figure}

\begin{table}[t]
\centering
\vspace{-30pt}
\caption{\textbf{Removing the trigger component suppresses spikes on clean and perturbed images, with small task-score changes in nine of the ten models.}
$\mathrm{VSI}_\kappa$ (\%), the fraction of images that spike, is shown before $\to$ after intervention: clean POPE uses 200 questions per model, corruption is the mean over six operators on 50 questions each, and attack uses 40 saved creation-attack images where available.
These are separate evaluation sets. Score changes are intervention minus original on 297 VQAv2 questions and 300 COCO captions.
A random direction leaves every model's reported clean spike rate unchanged.
Tables~\ref{tab:prevent}, \ref{tab:preventcorr}, \ref{tab:preventatk} and~\ref{tab:preventfree}\backanchor{tab:prevent}\backanchor{tab:preventcorr}\backanchor{tab:preventatk}\backanchor{tab:preventfree} give the detailed results.}
\label{tab:prevention_main}
\begingroup\footnotesize\setlength{\tabcolsep}{3pt}
\begin{tabular}{lr@{\,}c@{\,}rr@{\,}c@{\,}rr@{\,}c@{\,}rrr}
\toprule
 & \multicolumn{9}{c}{$\mathrm{VSI}_\kappa$ before $\to$ after (\%)} & \multicolumn{2}{c}{Task-score change} \\
\cmidrule(lr){2-10}\cmidrule(lr){11-12}
LVLM & \multicolumn{3}{c}{Clean} & \multicolumn{3}{c}{Corrupted} & \multicolumn{3}{c}{Attacked} & \shortstack{$\Delta$VQA\\(pct. points)} & $\Delta$CIDEr \\
\midrule
Qwen2.5-VL-32B & 49.0 & $\to$ & 0.0 & 90.0 & $\to$ & 0.0 & 100.0 & $\to$ & 0.0 & $+0.0$ & $-0.007$ \\
Qwen3-VL-4B & 71.0 & $\to$ & 0.0 & 89.7 & $\to$ & 0.0 & 95.0 & $\to$ & 0.0 & $-0.6$ & $+0.003$ \\
InternVL3.5-4B & 16.0 & $\to$ & 3.0 & 42.0 & $\to$ & 11.3 & 80.0 & $\to$ & 27.5 & $-0.4$ & $+0.010$ \\
Qwen3-VL-8B & 9.0 & $\to$ & 0.0 & 53.0 & $\to$ & 0.0 & 82.5 & $\to$ & 0.0 & $+0.1$ & $+0.008$ \\
InternVL3.5-8B & 59.5 & $\to$ & 32.5 & 86.7 & $\to$ & 62.0 & 100.0 & $\to$ & 52.5 & $+0.0$ & $+0.000$ \\
Molmo2-8B & 100.0 & $\to$ & 45.5 & 100.0 & $\to$ & 53.3 & \multicolumn{3}{c}{n/a$^{\ast}$} & $-0.3$ & $+0.008$ \\
InternVL3.5-14B & 61.0 & $\to$ & 16.5 & 84.3 & $\to$ & 17.0 & 100.0 & $\to$ & 30.0 & $+0.2$ & $-0.001$ \\
Qwen3-VL-32B & 35.0 & $\to$ & 0.0 & 60.3 & $\to$ & 0.0 & 92.5 & $\to$ & 0.0 & $+0.0$ & $+0.000$ \\
InternVL3.5-38B & 4.5 & $\to$ & 0.0 & 34.3 & $\to$ & 0.3 & 52.5 & $\to$ & 0.0 & $+0.0$ & $-0.005$ \\
Pixtral-12B & 100.0 & $\to$ & 0.0 & 99.7 & $\to$ & 0.0 & \multicolumn{3}{c}{n/a$^{\ast}$} & $+2.6$ & $+0.158$ \\
\bottomrule
\end{tabular}
\par\smallskip
\begin{minipage}{\linewidth}\scriptsize
Corruption means are computed from the rounded per-operator rates reported in \tref{tab:preventcorr}.
Attack images are fixed before intervention. $^{\ast}$Molmo2-8B and Pixtral-12B spike on nearly every clean image, so no attack creates a spike to prevent.
Stable task scores do not imply identical wording: Molmo2-8B and Pixtral-12B retain $53\%$ and $13\%$ of their captions, respectively (\tref{tab:preventfree}).
\end{minipage}
\vspace{-15pt}
\endgroup
\end{table}

In this work, we investigate visual massive activations across 25 adapter-based LVLMs built on 18 released text-only bases from 10 families, spanning 2B to 72B parameters.
Earlier studies characterize visual spikes on one or two LVLMs each, mostly LLaVA-1.5, and test their methods on up to nine \citep{kang2025see, choi2026sinks, luo2025tosink, yoo2026nature}.
The breadth here exposes patterns across models and bases that a few models cannot show.
We analyze each LVLM alongside its text-only base to investigate how spiking mechanisms are inherited through visual adaptation.

Our experiments reveal that visual spikes do not occur in all LVLMs.
Using the original massive-activation criterion, we find that ten models form visual spikes at different rates, while fifteen do not under our measurements. This separates them into two regimes.
This distinction is obscured by the commonly used visual-spike definition, which measures concentration on inherited channels and can mark most image tokens without identifying exceptional activations \citep{kang2025see, yoo2026nature}.
Measuring the unusual attention attracted by the image tokens supports the separation.
To explain these different regimes, we examine how image tokens activate the mechanism inherited from the base.
We find that image tokens reach a block that produces the spike, the \emph{visual switch}, under some visual adaptations and not others, and that this block typically sits deeper than the text switch.
We also identify a trigger direction for this block. Inputs sufficiently aligned with this direction cause the block to produce a spike. We formulate this direction from the block's weights and estimate it while retaining the full nonlinearity, recovering the quadratic eigendirection of \citet{sun2026spike} as a special approximation (Section~\ref{sec:mech}).
The text spike, by contrast, is gated by the norm alone, since at position 0 a sufficiently small input activates the text switch across the tested directions (Section~\ref{sec:switch}).
We also establish that the visual spike depends on the text spike's attention sink, since blocking the eventual spike token's attention to the text spike token suppresses the visual spike on $75$ to $100\%$ of images (Section~\ref{sec:switch}).

With the visual switch and its trigger identified, we examine what distinguishes the image tokens that will spike by analyzing their representations before the decoder runs.
We find that eventual spike tokens are already largely restricted to those sharing least with the rest of the image.
This least shared rule predicts later alignment with the trigger and gives a finer, interpretable location rule than the background preference observed in previous studies.
We further show that scaling down the tokens of a chosen region as small as a tenth of the image, which lowers their shared part and norm, draws the spike into that region (Section~\ref{sec:location}).

We then investigate the sensitivity of visual spikes to image perturbations and find that they are brittle. 
Common corruptions create and move spikes more often than they remove them, which raises the fraction of images that spike.
Even when incidence remains stable, the token that spikes can change.
Our \emph{spike attack} tests this sensitivity deliberately, using the estimated trigger to create or remove spikes under a small pixel budget.
The attack succeeds under a small $\ell_\infty$ budget, with $\nicefrac{1}{255}$ enough in nine of the ten spiking models, while the models' yes/no answers remain unchanged on $93$ to $100\%$ of attempts despite the absence of an explicit answer-preservation objective.
These results reveal internal brittleness that would not show in the model's answers alone (Section~\ref{sec:brittleness}, Figure~\ref{fig:britt}).

Based on our trigger estimation, we propose a preventive intervention that suppresses this large and unstable state before it is written.
Existing interventions that edit massive activations act after the write, whereas ours removes the trigger component from the input read by the visual switch's feed-forward.
This requires no token deletion or weight change.
Across the evaluated models, the intervention reduces visual spikes on both clean and perturbed images, eliminating them on clean images in six of the ten models, while preserving the switch's write on ordinary tokens and, in eight of the ten, the model's outputs (Section~\ref{sec:prevent}).

In sum, we shed more light on how visual spikes form, from the inherited switch and its trigger to the tokens that reach it, and we demonstrate their extreme brittleness under image changes.

\section{Background and setup}
\label{sec:setup}

We defer the discussion of related work to  \aref{app:relwork}.

\textbf{Formalism of text spike formation by \citet{sun2026spike}.}\quad
Pretraining puts the spike on a few model-specific channels, each associated with an outlier row of the down projection matrix of an early feed-forward block \citep{sun2024massive, yu2024super}.
For each \emph{spike channel}, this row acts as a \emph{writer} whose output is amplified when the block's input aligns with a \emph{trigger direction}.
\citet{sun2026spike} describe this mechanism as a directional quadratic amplifier.
An \emph{eruption} occurs when this writer makes a token's spike channel massive within a single block, turning it into a \emph{spike token}.
\mbox{\citet{sun2026spike}} call the responsible block the step-up block. We refer to it as the \emph{switch} to emphasize its role in determining whether the spike state is on or off, and, when the spike is a text spike, as the \emph{text switch}.
At position 0, the first token attends only to itself, which drives its residual toward the trigger direction \citep{sun2026spike}.
 \aref{app:drive} explains the mechanism in full.

\textbf{Spike selection criteria.}\quad
\label{sec:stats}
Let $\vb{x}_i^{(\ell)} \in \mathbb{R}^D$ denote image token $i$'s residual entering block $\ell$, and $D_{\mathrm{spike}}$ the spike channels inherited from the base.
We write the original massive-activation ratio of \citet{sun2024massive} and the channel-concentration ratio of \citet{kang2025see} as
\[
\Psi_i^{(\ell)}
= \frac{\max_{d \in D_{\mathrm{spike}}} |\vb{x}_i^{(\ell)}[d]|}
{\mathrm{median}_{j,k} |\vb{x}_j^{(\ell)}[k]|},
\qquad
\phi_i^{(\ell)}
= \frac{\max_{d \in D_{\mathrm{spike}}} |\vb{x}_i^{(\ell)}[d]|}
{\mathrm{RMS}(\vb{x}_i^{(\ell)})}.
\]
The ratios differ only in their denominator.
$\Psi$ compares the activation with the median magnitude across all tokens and channels of the layer, while $\phi$ measures concentration within the token.
The original massive-activation criterion declares an image to spike when, in at least one measured deep layer, an eligible image token satisfies $\Psi_i^{(\ell)} \ge 1{,}000$ and has an absolute magnitude of at least $100$; the channel-concentration criterion uses $\phi_i^{(\ell)} \ge 20$.
A token is eligible when it enters the language model below $100$ on the spike channels, excluding the ViT-emerged tokens of \citet{choi2026sinks} and \citet{luo2025tosink}.
Further details on spike measurement and the image-level peak $\Psi_{\mathrm{peak}}$ are provided in  \aref{app:measure}.

\textbf{Attention sink.}\quad
We measure attention sinks independently of massive activations.
Let $\bar{a}_i^{(\ell)}$ denote the attention received by image token $i$ at layer $\ell$, averaged over heads and text queries following the image.
Following \citet{gu2025when}, an image token is an \emph{attention sink} if
$\bar{a}_i^{(\ell)} / \sum_{j \in I} \bar{a}_j^{(\ell)} \ge 0.3$ in at least one measured deep layer, where $I$ is the set of image tokens.
Further measurement details are provided in  \aref{app:measure}.

\textbf{Setup.}\quad
We study 25 open adapter-based LVLMs built on 18 released text-only bases from 10 families, spanning 2B to 72B parameters (\aref{app:models}).
Each base carries a text spike and is adapted into an LVLM by connecting a vision encoder through a projector.
We use \emph{visual adaptation} to refer to both the added components and any changes made to the base model.
We also evaluate the text-only bases on text and on the image tokens produced by their corresponding LVLMs, to distinguish inherited mechanisms from the effects of adaptation.
Unless stated otherwise, we evaluate each model on 300 clean COCO validation images \citep{lin2014microsoft}, with the image preceding a short question, and record the image-token residual states after each decoder block.
We assess task performance on POPE \citep{li2023pope}, VQAv2 \citep{goyal2017vqav2}, and COCO captioning \citep{chen2015cococaptions}, using accuracy, VQA accuracy, and CIDEr \citep{vedantam2015cider}, respectively.

\section{When and how visual spikes form}
\label{sec:mech}

While LVLMs inherit spike channels from their text-only bases, visual spikes do not always form.
We first establish which LVLMs form visual spikes, then identify the inherited blocks and input directions that write them, which the spike attack and the preventive intervention target.

\subsection{Visual spikes are not universal across LVLMs}
\label{sec:chanregimes}\label{sec:attn}

\textbf{Ten models spike and fifteen do not.}\quad
We define \emph{visual-spike incidence}, $\mathrm{VSI}_\kappa$, as the fraction of images containing an image token that meets the massive-activation criterion at $\Psi\ge\kappa$, with an absolute floor of $100$.
We use $\kappa=1{,}000$;  \aref{app:measure} gives the full expression and measurement details.
Across 300 clean images per model, fifteen LVLMs spike on none of them, while incidence in the other ten ranges from $1.7$ to $100\%$ (Figure~\ref{fig:crit}).
We call these observed regimes \emph{non-spiking} and \emph{spiking}.
Their largest clean peaks are separated: the non-spiking models reach at most $877$, while every spiking model reaches at least $5{,}278$.
The fifteen non-spiking models also remain below the criterion throughout the Gaussian-noise sweep (\tref{tab:noiseroster}).
Within a shared base, adaptation changes incidence by up to $87$ percentage points without changing the regime (\tref{tab:fullroster}).
Between two attention implementations, up to $35$ of $200$ evaluation forwards change side, and no model changes regime (\tref{tab:kernel}).

\textbf{Channel concentration obscures the two regimes.}\quad
The criterion $\phi\ge20$ of \citet{kang2025see} selects $83$ to $100\%$ of image tokens in 21 of the 25 LVLMs and makes 23 models appear to spike on every image (Figure~\ref{fig:crit}).
This widespread selection agrees with \citet{yoo2026nature}'s observations on two omni-modal models.
By comparison, $\Psi\ge1{,}000$ selects at most $1\%$ of tokens on the same forwards.
 \aref{app:census} gives the model-specific results, the bound $\phi\le\sqrt{D}$, and the threshold comparisons.
Changing or calibrating the concentration threshold does not recover the two regimes.

\textbf{Independently measured attention reflects the separation.}\quad
Across POPE, VQAv2, and captioning, an attention sink forms on $96.7$ to $97.6\%$ of spiking forwards pooled over nine spiking models, almost always on the spike token, and on under $1\%$ of their other forwards.
In Molmo2-8B the sink forms on $24$ to $34\%$ of spiking forwards, and some non-spiking models form attention sinks without massive activations (\tref{tab:attntasks}).
The measurements support the separation while distinguishing the two observables.

\subsection{The visual switch}
\label{sec:switch}

\textbf{A visual switch or none.}\quad
In the ten spiking LVLMs, one block takes an image token's spike channel from the tens to the thousands, after which the value persists until the final layers.
The fifteen non-spiking models have no such write (Figures~\ref{fig:one} and~\ref{fig:switchfour}; \tref{tab:twoswitch}).
The abrupt write and late decline match the pattern observed in text by \citet{sun2024massive}, but the visual switch lies later than the text switch in nine of the ten models.
Pixtral-12B uses the same block for both (\aref{app:extension}).

\textbf{A small norm activates the text switch, direction activates the visual switch.}\quad
When a single vector of chosen norm and direction is fed at position 0, a sufficiently small norm activates the text switch across the tested directions, even without a vocabulary token.
The residual is driven toward the trigger, while ordinary image tokens at their own norm do not follow this route (Tables~\ref{tab:pos0} and~\ref{tab:pos0all}\backanchor{tab:pos0}\backanchor{tab:pos0all}).
Image tokens instead activate a visual switch by aligning with its trigger direction $\vb{t}$.
Across the ten spiking models and their six bases, the switch's block, writer, and trigger are inherited from the base (\tref{tab:sun}).

\textbf{The visual spike depends on the text spike's attention sink.}\quad
The eventual spike token gives the text spike token $5$ to $33\%$ of its attention before the text switch and $51$ to $79\%$ after it, and $43$ and $39\%$ in InternVL3.5-38B (\tref{tab:formcausal}).
Blocking that attention suppresses the visual spike on $75$ to $100\%$ of images, and on two of InternVL3.5-38B's three (\tref{tab:formcausal}).
The blocked attention goes instead to image tokens whose values are $1.9$ to $6.5$ times the text spike token's, and their writes overwrite the eventual spike token's state before it reaches the switch (\tref{tab:overwrite}).
A visual spike therefore needs the attention sink on the text spike token.
 \aref{app:formation} gives the per-model results.

\textbf{The adaptation makes an inherited switch reachable.}\quad
The base provides several blocks capable of writing massive values on the inherited channels. We call them \emph{candidate blocks}. Whether image tokens reach their triggers differs with the adaptation.
In three of the six bases, image tokens pass one to fifteen candidate blocks without spiking, so the visual switch need not be the first (\tref{tab:passed}).
Among these blocks, image-token residuals approach only the visual switch, while Gaussian or zero vectors at image positions reach none (Tables~\ref{tab:capacity} and \ref{tab:reach}).
The text-only base run on an LVLM's decoder-input vectors does not spike on the LVLM's spike token, but given the LVLM's state entering the visual switch it does on most spiking images (Tables~\ref{tab:basefed} and~\ref{tab:basefedall}\backanchor{tab:basefed}\backanchor{tab:basefedall}).
The adaptation can also prevent alignment that occurs in the base (\tref{tab:basefedov}).
When several spiking LVLMs share a base, they nevertheless use the same visual switch.

\subsection{Trigger direction estimation}
\label{sec:trigger}
Let $\vb{w}_d$ be the writer of spike channel $d$, $\vb{h}(\vb{u})$ the block's hidden activation for a unit input $\vb{u}$, and $s_d\in\{-1,+1\}$ the sign of its write on channel $d$.
We formulate the trigger direction as
\begin{equation}
\vb{t} = \arg\max_{\lVert \vb{u} \rVert = 1} \, s_d \, \vb{w}_d^{\top} \vb{h}(\vb{u}),
\label{eq:trigger}
\end{equation}
where $s_d$ is measured on a forward pass at the switch.
Given that sign, we estimate the maximizer by gradient ascent on the unit sphere from several starts, using the block's weights and that sign.
The objective is exact and general, since it retains the full nonlinearity (\aref{app:drive}).

Replacing the SwiGLU activation by the identity recovers the quadratic form of \citet{sun2026spike}, with the norm's weights folded in.
Its maximizer is the leading eigenvector for the measured write sign, making the eigendirection a special approximation.
The quadratic form cannot distinguish opposite directions, whereas the full nonlinear optimization selects the orientation that produces the measured write.
The estimated trigger matches the eigenvector for the measured write sign at cosine $0.999$ to $1.000$ on every checkpoint (\aref{app:drive}).

We estimate the trigger at the visual switch of all six bases.
At the switch input, the eventual spike token has a median cosine of $0.08$ to $0.62$ with $\vb{t}$, while $99\%$ of other tokens lie below $0.05$ (\tref{tab:trigcos}).
InternVL3.5-38B, which spikes on five of its 300 census images, sits at $0.06$, and in Molmo2-8B, where multiple spike tokens align, the bound is $0.11$.
Alignment ranks the eventual spike token first on $75$ to $100\%$ of spiking images, and on $44\%$ in Molmo2-8B.
This identifies the direction the attack can promote or oppose and the component the intervention can remove before the write.

\section{Where the visual spike forms}
\label{sec:location}\label{sec:steer}

Trigger alignment explains which token spikes at the switch.
We find a finer, interpretable restriction before the decoder runs: the eventual spike token is largely drawn from those that share least with the rest of the image.

\textbf{The shared part predicts trigger alignment and the spike token.}\quad
For an image token $\vb{x}_i$, the mean of the other $N-1$ tokens, $\vb{m}_{-i}=\frac{1}{N-1}\sum_{j\neq i}\vb{x}_j$, points along the direction the rest of the image shares.
We define its \emph{shared part} as the scalar projection
$p_i=\nicefrac{\vb{x}_i^{\top}\vb{m}_{-i}}{\lVert\vb{m}_{-i}\rVert}$, which combines the token's norm with its cosine to that direction.
Neither part locates the spike token on its own.
The norm ranks it well in the Qwen3-VL models and Molmo2-8B and fails in Qwen2.5-VL-32B and Pixtral-12B, where the direction alone carries it, while the direction collapses in the Qwen3-VL models (\tref{tab:locator}).
As a within-image ranking, the shared part has the best median AUC and FPR@95, $0.893$ and $28\%$ against $0.850$ and $48\%$ for the norm and $0.824$ and $78\%$ for the direction, admitting that share of other tokens when covering $95\%$ of spike tokens (Figure~\ref{fig:loc}; \tref{tab:locator}).

The eventual spike token falls in the least-shared tenth on $43$ to $97\%$ of spiking images, against $10\%$ by chance, and on all five of InternVL3.5-38B's (\tref{tab:decile}).
It falls in the least-shared quarter on $71$ to $100\%$ of spiking images, against $25\%$ by chance.
Across tokens, the Spearman correlation between this decoder-input shared part and later trigger alignment has a negative median over spiking images in all ten models ($-0.57$ to $-0.05$), and the least-shared tenth has higher mean trigger alignment than the most-shared tenth in every model (\tref{tab:c4corr}).
The least-shared tenth also remains predictive under Gaussian noise (\aref{app:location}).

This rule is finer than the background preference observed by \citet{kang2025see}.
Background covers $60$ to $75\%$ of image tokens and adds little localization once we know the least-shared tenth (Tables~\ref{tab:bgassoc} and~\ref{tab:bgcond}\backanchor{tab:bgassoc}\backanchor{tab:bgcond}, \fref{fig:bgdrive}).
In Figure~\ref{fig:one}, both marked image tokens lie in the sky, but the blue token, which shares most with the rest of the image, decodes to the sky, while the red spike token decodes to a repetitive pattern.
Across models, the spike token rarely decodes to the scene (\tref{tab:decodesum}).
 \aref{app:location} gives further token decodings and the geometric comparisons.

\textbf{The candidates are constrained early and the token is selected near the switch.}\quad
The shared part fixes the candidates before the decoder runs, and the token's state decides among them only near the switch.
Swapping an eventual spike token's state with that of another token carries the spike to the new position on $11$ to $100\%$ of images at the decoder input and on $63$ to $100\%$ one block before the switch, and on none of InternVL3.5-38B's three (\tref{tab:swapdepth}).
The closer the swap is to the switch, the more often the moved state carries the spike, suggesting that the choice among the candidates is completed late.
In eight of ten models, trigger alignment remains near zero until one or two blocks before the switch, where attention and feed-forward writes produce it (\tref{tab:aimorigin}).

\textbf{Scaling a region down moves the spike into it.}\quad
Before the decoder runs, we edit the tokens of a chosen region, either changing their shared part at fixed norm or scaling their full vectors to a tenth.
Scaling a quadrant down draws the spike into it on $33$ to $79\%$ of spiking images in six models, against $9$ to $19\%$ on the clean forward, and on $82\%$ of InternVL3.5-38B's eleven (\tref{tab:steer}).
In Qwen2.5-VL-32B and Pixtral-12B the spike does not move into the scaled quadrant.
We read the pull as two effects of the edit.
The scaled tokens now share least with the image, so they become candidates. They also become the smallest tokens, and in these models the smallest candidate takes the spike.
The effect holds for regions as small as a tenth of the image and even for a single token (Figures~\ref{fig:loc} and~\ref{fig:blocks}\backanchor{fig:blocks}).

\section{Visual spikes are brittle to image perturbations}
\label{sec:brittleness}

Unlike the text spike, which recurs at a fixed initial position, the visual spike depends on the image-token trajectory. Its prompt independence follows from the causal mask and says nothing about stability to image changes. The state is brittle, meaning that a small image change creates, removes, or relocates it, under common corruptions and under a white-box, trigger-guided adversarial attack.
We also record whether the model's answer changes, since the tests target the state and the attack's objective contains no term on the answer or the output.

\textbf{Common corruptions create and relocate spikes more often than they remove them.}\quad
Gaussian noise raises spike incidence in every spiking model below saturation at the strongest tested grade, while the fifteen non-spiking models remain non-spiking (\tref{tab:noiseroster}, \fref{fig:noisesweep}).
Across ten spiking models, six corruptions, and 200 questions per model and operator, we observe $3{,}152$ creations against $476$ removals (Figure~\ref{fig:britt}, \tref{tab:corrpanel}).
Thus, corruption raises overall incidence, although individual model--operator pairs can move the other way.
The answer's yes-or-no decision remains unchanged on $73.5$ to $96.5\%$ of forwards.

Incidence alone understates the brittleness.
On Qwen3-VL-32B, increased brightness changes incidence only from $0.41$ to $0.39$, yet 39 questions gain a spike and 43 lose one.
Among forwards that spike both before and after corruption, the spike changes token on $85$ to $100\%$ of them in Qwen3-VL-4B, Qwen3-VL-8B, and Qwen3-VL-32B.
Molmo2-8B retains a spike on every corrupted input but relocates it on $96$ to $100\%$ of them.
The location is more stable in Qwen2.5-VL-32B and Pixtral-12B, which relocate on $6$ to $14\%$ and $2$ to $20\%$, respectively (\aref{app:corruptions}).
Both the existence and the location of the massive state must therefore be measured to assess its stability.

\textbf{The spike attack.}\quad
For an image $\vb{x}$ and the visual-switch block $\ell^{\ast}$, let $c_i(\vb{x}) = \cos(\vb{x}_i^{(\ell^{\ast})}, \vb{t})$ be the cosine similarity between image token $i$'s residual at the switch's input and the trigger $\vb{t}$.
Let $v_i(\vb{x}) = s_d\,\vb{x}_i^{(\ell^{\ast}+1)}[d]$ be its signed spike-channel value at the switch's output, using the same write sign $s_d$ as in Section~\ref{sec:trigger}.
We formulate the spike attack as
\begin{equation}
\vb*{\delta}^{\ast} = \arg\max_{\substack{\lVert \vb*{\delta} \rVert_\infty \le \varepsilon \\ \vb{x} + \vb*{\delta} \in [0, 1]^{n}}} \; \pm \bigg[ \tau_1 \operatorname*{logsumexp}_{i \in I} \frac{c_i(\vb{x} + \vb*{\delta})}{\tau_1} + \frac{1}{1000} \min\Big\{ \tau_2 \operatorname*{logsumexp}_{i \in I} \frac{v_i(\vb{x} + \vb*{\delta})}{\tau_2}, 2000 \Big\} \bigg],
\label{eq:attack}
\end{equation}
where $\operatorname{logsumexp}_{i \in I} a_i = \log \sum_{i \in I} e^{a_i}$, with $\tau_1 = 0.02$ and $\tau_2 = 50$.
The plus sign creates the state and the minus sign abolishes it.
We optimize with signed-gradient steps of $\nicefrac{\varepsilon}{4}$ projected onto the constraints and keep the first step at which the spike criterion flips.
The model's answer does not enter either the objective or the stopping rule.
 \aref{app:attack} gives the derivation of the objective and the full attack procedure.

At an $\ell_\infty$ budget of $\nicefrac{1}{255}$, equivalent to one 8-bit intensity level per pixel channel, the attack creates the massive state on $53$ to $100\%$ of eligible images in the eight models with non-spiking inputs.
It removes the state on $95$ to $100\%$ of eligible images in those eight models and in Pixtral-12B under the schedules of \tref{tab:epssweep}.
Larger budgets add successes. Molmo2-8B removes the state on $88\%$ of eligible images at $\nicefrac{4}{255}$ and on $98\%$ at $\nicefrac{8}{255}$.
Across these attacks, the yes/no answer remains unchanged on $93$ to $100\%$ of attempts.
On the free-form tasks, VQAv2 accuracy changes by $0.6$ to $4.9$ percentage points in absolute value, while CIDEr moves between $-0.10$ and $+0.11$ (\tref{tab:attackfreeform}).
The strongest answer-preservation result is therefore on POPE, where internal spike changes routinely occur without a changed decision.
\fref{fig:cimg} shows the state being created, abolished, and relocated by the attack and by a corruption, and  \aref{app:attack} gives the full budget sweep and per-model results.
The attack demonstrates control over a large internal state that can remain hidden by stable answers.

\section{Preventive intervention}
\label{sec:prevent}

Our trigger estimate lets us suppress the spike before the switch writes it.
Let $\bar{\vb{x}}_i$ be image token $i$'s residual after the switch's attention sub-block.
We remove the trigger component from the input read by its feed-forward,
\begin{equation}
\bar{\vb{x}}_i' = \bar{\vb{x}}_i - (\vb{t}^{\top} \bar{\vb{x}}_i)\, \vb{t},
\label{eq:null}
\end{equation}
on every image token.
The residual connection still carries $\bar{\vb{x}}_i$, every weight is untouched, and no token is deleted.
Unlike interventions that remove spike tokens after formation \citep{kang2025see}, this edit acts before the massive write.

On clean POPE inputs, the intervention eliminates spikes in six of ten models and reduces their incidence by $45$ to $81\%$ in the others (Table~\ref{tab:prevention_main}).
Suppression persists across all six corruptions. Five models remain at zero incidence, and accuracy stays within $4$ percentage points of the unmodified model.
On the images the attack produced before the intervention, it eliminates induced spikes in five of the eight models and reduces incidence in the other three, with attacked accuracy unchanged (Tables~\ref{tab:preventcorr} and~\ref{tab:preventatk}\backanchor{tab:preventcorr}\backanchor{tab:preventatk}).

Across eight models (all but Molmo2-8B and Pixtral-12B), task outputs are nearly unchanged. Of $1{,}600$ clean POPE answers, $1{,}599$ are identical, VQA accuracy moves by at most $0.006$, and CIDEr by at most $0.010$.
Molmo2-8B and Pixtral-12B change more in their free-form outputs, retaining $53\%$ and $13\%$ of captions, compared with $88\%$ and $63\%$ under a random direction of the same rank (\tref{tab:preventfree}).

\begin{table}[t]
\vspace{-25pt}
\caption{\textbf{The trigger intervention suppresses the spike without disabling the switch.} The ten models that raise a visual spike, 200 evaluation questions each, every edit on all image tokens at the switch's read.
Each cell is the median over the ten models with its range, and \emph{at zero} counts the models left with no spiking forward.
$\cos F$ and the ratio are medians over image tokens of $\cos(F(\bar{\vb{x}}'), F(\bar{\vb{x}}))$ and $\lVert F(\bar{\vb{x}}')\rVert / \lVert F(\bar{\vb{x}})\rVert$.
Bold marks the best among the edits that remove the spike.}
\vspace{-10pt}
\label{tab:baselines}
\begin{center}
\footnotesize
\setlength{\tabcolsep}{3pt}
\resizebox{\linewidth}{!}{\begin{tabular}{lrrrrr}
\toprule
edit at the switch & spikes $\downarrow$ & at zero $\uparrow$ & answers same $\uparrow$ & $\cos F$ $\uparrow$ & $\lVert F \rVert$ ratio \\
\midrule
\emph{no edit} & 0.54 (0.04--1.00) & 0/10 & 1.00 & 1.000 & 1.000 \\
\midrule
ours, the trigger & 0.00 (0.00--0.45) & 6/10 & \textbf{1.00 (0.94--1.00)} & \textbf{0.999 (0.980--1.000)} & \textbf{1.000 (0.948--1.004)} \\
the read zeroed, $\vb{x}'=\vb{0}$ & \textbf{0.00} & \textbf{10/10} & 0.99 (0.94--1.00) & 0.000 & 0.000 \\
random $\vb{r}$ & 0.54 (0.04--1.00) & 0/10 & 1.00 & 1.000 (0.999--1.000) & 1.000 \\
\bottomrule
\end{tabular}}
\end{center}
\end{table}

\textbf{The trigger intervention against less specific edits.}\quad
Zeroing the feed-forward input removes spikes but also zeroes the feed-forward output itself.
Removing a random direction leaves spikes untouched.
Trigger removal largely preserves the feed-forward output on ordinary image tokens, with a token-median cosine of $0.999$ and a norm ratio of $1.000$ across models (Table~\ref{tab:baselines}).
The write changes most on tokens that would have spiked, whose trigger alignment drives the massive output.
 \aref{app:prevent} gives the detailed results and a conditional variant.

\section{Conclusion}
\label{sec:conclusion}
Our study explains how visual massive activations form and establishes control over their creation, relocation, and suppression.
Their functional role, however, remains open.
Suppressing spikes leaves behavior nearly unchanged in eight of the ten spiking models, while substantially changing free-form outputs in the other two.
These changes alone do not establish whether spikes support useful computation, and stable answers do not rule out a role under other conditions.
Whether and when the visual massive state contributes to model behavior, including in multi-image reasoning and video, remains a question for future work.
We discuss the limitations of our study and further research directions in \aref{app:limits} and~\aref{app:future}.

\FloatBarrier

\label{end:body}

\section*{Reproducibility statement}
All 25 models are public checkpoints (\tref{tab:fullroster}, \aref{app:models}).
The images are COCO val2014, the 300-image census drawn by a hash of a fixed seed and the edit set taken from the POPE-derived manifest (\aref{app:data}).
The criterion is defined in Section~\ref{sec:setup} and the image-level peak in \aref{app:measure}, $\mathrm{VSI}_\kappa$ in Section~\ref{sec:chanregimes}, and all three are measured as \aref{app:measure} describes.
The trigger is estimated from a checkpoint's weights and the measured sign of the eruption by the procedure of \aref{app:drive}, the spike attack and its budgets are given in \aref{app:attack}, and the Gaussian noise grades, with their seeds, in \aref{app:ops}.
The hardware, the wall-clock of each kind of run and the compute the paper costs are in \aref{app:compute}, about $53$ B200-hours. The attack accounts for roughly $23$ of these, most of it on Molmo2-8B and InternVL3.5-38B, and the total is a rental equivalent of about \$360 at September 2026 rates.
The project as a whole used considerably more than that, since the exploratory and superseded runs are not counted.
We will release the code that produces every table and figure.

\section*{The use of large language models}
Large language models were used throughout this work, under the authors' direction and with the authors responsible for every claim.
They assisted with and polished the writing, and they drafted text for sections that the authors then revised.
They were used to search for and retrieve related work, whose bibliographic details were verified against the original sources.
They also drafted analysis code, ran and summarized experiments, and maintained the project's research record.
The research questions, the story, the rulings on terminology and claims, and the final text are the authors'.

\section*{Ethics statement}
This work describes a white-box attack that creates or removes an internal activation state in open vision--language models under a small pixel budget. The attack requires the model's weights, the visual switch, and its trigger. It does not optimize the model's answer, and was evaluated only on public checkpoints and public benchmark images. We also provide the preventive intervention that suppresses the state it targets, and we will release both together. We see no direct path from this work to harm beyond what white-box access already permits.

\bibliography{references}
\bibliographystyle{iclr2027_conference}

\clearpage
\appendix
\newif\ifinappendix\inappendixtrue
\setcounter{topnumber}{1}\setcounter{bottomnumber}{0}\setcounter{totalnumber}{1}
\renewcommand{\topfraction}{1}
\renewcommand{\bottomfraction}{0}
\renewcommand{\textfraction}{0}
\renewcommand{\floatpagefraction}{1}
\makeatletter
\def\fps@figure{t}
\def\fps@table{t}
\makeatother
\newcommand{\appcontinued}{\par\vfill\noindent\hfill{\scriptsize(Appendix continues on next page)}\par}
\AtEndEnvironment{figure}{\ifinappendix\appcontinued\clearpage\fi}
\AtEndEnvironment{table}{\ifinappendix\appcontinued\clearpage\fi}

\startcontents[appendix]

\section*{Appendix contents}
\printcontents[appendix]{}{1}[2]{}

\clearpage

\FloatBarrier
\section{Related work}
\label{app:relwork}
\backto*{app:relwork}

\textbf{Attention sinks and massive activations in language models.}\quad
Attention sinks often occur on initial tokens, whose privileged attention depends on position rather than content \citep{xiao2023streamingllm, clark2019what}.
\citet{sun2024massive} link this attention pattern to massive activations that act as bias terms.
These activations appear abruptly in early layers and persist until the last layers, a pattern observed across pretrained models \citep{sun2024massive, gu2025when}.
\citet{yu2024super} connect the massive write to a row containing an outlier weight in the feed-forward down projection and show that instruction tuning preserves this weight.
\citet{sun2026spike} describe the writing mechanism as a directional quadratic amplifier.
They find that nearly the whole vocabulary produces a spike at position 0, where the token attends only to itself, and characterize formation through alignment with a direction fixed by the weights.
Our position-0 probes further show how the input norm determines whether the residual is driven toward that direction (Section~\ref{sec:mech}).
\citet{choi2026textloss} find that visual adaptation weakens the text spike token's hold on attention. \citet{anand2026avsr} tie massive activations on later tokens to their alignment with the first token's direction in audio-visual speech recognition by rotating hidden states toward it. In LVLMs, blocking the eventual spike token's attention to the text spike token suppresses the visual spike on $75$ to $100\%$ of images (Section~\ref{sec:switch}).
We compare LVLMs with their text-only bases to distinguish the inherited writing mechanism from changes introduced by visual adaptation.
Attention sinks are also used in long-context and streaming inference \citep{xiao2023streamingllm, jolicoeur2026sliding}, while massive activations and the attention they attract inform quantization, token pruning, and KV-cache compression \citep{su2026survey}.

\textbf{Visual-spike formation and location.}\quad
\citet{kang2025see} characterize visual attention sinks on clean images, associating them with large values on inherited hidden channels and a preference for background regions.
They attribute visual and text sinks to a shared mechanism inherited from the base language model, based on their common channels and early onset.
The survey of \citet{su2026survey} also presents visual massive activations as following the text pattern.
\citet{choi2026sinks} distinguish two origins.
\emph{ViT-emerged} sinks originate in high-norm vision-transformer tokens \citep{darcet2024vision} carried through the projector, whereas \emph{LLM-emerged} sinks are written inside the language model.
They identify early feed-forward blocks that write the latter in LLaVA-1.5 and LLaVA-OneVision, and \citet{luo2025tosink} make the same distinction between origins.
We study spikes written inside the language model, separating them from tokens already massive at the decoder input (\aref{app:measure}).
Our base--adaptation comparisons identify an inherited visual switch, but in nine of the ten spiking LVLMs it differs from the block that writes the text spike (Section~\ref{sec:mech}).
Beyond the background preference, we identify a restriction on the eventual spike token already present before the decoder runs and test it through token interventions (Section~\ref{sec:location}).

\textbf{Distinguishing massive activations from attention sinks.}\quad
\citet{ipm2025rope} explain visual attention sinks through rotary position decay over the image sequence and massive activations.
\citet{sun2026spike} distinguish the two phenomena in language models, which motivates us to measure them independently on image tokens.
Their relationship also depends on how massive activations are identified.
\citet{kang2025see} threshold a token's concentration on inherited channels, whereas the original criterion of \citet{sun2024massive} compares activation magnitude with the rest of the layer.
On two omni-modal models, \citet{yoo2026nature} find that the concentration criterion selects most tokens in deep layers, including highly attended object tokens, while the original criterion selects a single token.
\citet{choi2026sinks} instead use model-specific thresholds on raw activation magnitude.
We compare these criteria across 25 LVLMs and measure the fraction of images that form a visual massive state.
The original criterion reveals spiking and non-spiking regimes that widespread channel concentration obscures, with independent attention measurements largely supporting the separation (Section~\ref{sec:chanregimes}).
\aref{app:census} gives the concentration bound and threshold comparisons.

\textbf{Visual and text representations.}\quad
Visual projectors produce long streams of continuous, highly correlated token embeddings \citep{lee2026wordlike}.
\citet{chen2024fastv} report inefficient attention over visual tokens in deep layers and show that many can be pruned after the early layers.
\citet{zhang2024treat} identify further redundancy in attention heads and layer computation when processing visual tokens.
At the decoder input, \citet{fan2026encode} distinguish \emph{sink}, \emph{dead}, and \emph{alive} token populations, associating image-specific semantic content with the alive group.
\citet{li2025lost} show that the projector changes local representation geometry, substantially altering nearest-neighbour relationships.
We connect decoder-input geometry to later spike formation.
The shared part of an image token predicts its alignment with the inherited trigger and gives a finer location rule than background membership.
Scaling down a region's tokens before the decoder runs lowers their shared part and norm and moves the eventual spike into that region (Section~\ref{sec:location}).

\textbf{Image perturbations and attacks on attention sinks.}\quad
Beyond characterizing visual spikes on clean images, we examine whether their existence and location remain stable when the image changes.
Common corruptions test this stability without targeting the mechanism, while our spike attack deliberately promotes or opposes its activation.
Related adversarial work uses attention sinks in different ways.
\citet{shang2025forgetting} place training-time backdoor triggers at the sink positions of a language model.
\citet{wang2025mirage} perturb an image to induce an attention sink on a token of the generated answer, producing a columnar attention pattern mainly on non-content tokens.
Their objective combines attention and embedding similarity to induce hallucinated content, and the attack transfers across LVLMs.
Our spike attack instead targets image tokens and the massive-activation state itself.
It uses the inherited trigger to create or remove that state under a small pixel budget.
The answer enters neither the objective nor the stopping rule, allowing us to measure how often large internal changes occur while the answer remains unchanged (Section~\ref{sec:brittleness}).

\textbf{Interventions on the spike and the attention sink.}\quad
In language models, \citet{sun2024massive} show that zeroing massive activations severely degrades perplexity and task accuracy, whereas replacing them with their empirical mean leaves performance nearly unchanged.
\citet{queipo2025sinks} zero the first token's feed-forward write at the layers responsible for its massive activation, eliminating both the attention sink and the associated representational compression in LLaMA3-8B.
In LVLMs, \citet{kang2025see} find that removing visual sink tokens has little effect on the evaluated tasks.
Their Visual Attention Redistribution method reallocates attention from these tokens to other image tokens in selected heads, improving visual understanding and reducing hallucinations.
SinkPruner combines high-norm token filtering with text-guided pruning, alleviating attention sinks and retaining most benchmark performance with substantially fewer visual tokens \citep{li2026sinkpruner}.
\citet{salazar2026pathways} show that blocking direct attention from the answer token to image tokens can leave task performance intact because visual information is rerouted through the query tokens.
For quantization, INSERTQUANT clamps spikes and restores their downstream role through precomputed template vectors, largely preserving full-precision performance and reducing performance loss under low-bit quantization \citep{chen2026biasvectors}.
Related interventions remove information from representations.
INLP and LEACE remove linearly decodable information about selected attributes or concepts \citep{ravfogel2020null, belrose2023leace}, while removing a refusal direction suppresses refusal behaviour \citep{arditi2024refusal}.
Our trigger estimate identifies the input component that activates the massive writer.
Removing this component from the visual switch's feed-forward input suppresses spikes before they are written, while largely preserving the feed-forward output on ordinary image tokens.
The intervention eliminates visual spikes in six of the ten models and reduces their incidence by $45$ to $81\%$ in the other four, on clean and perturbed images, with little change in task behaviour in eight of the ten spiking models and larger changes in free-form outputs in the other two (Sections~\ref{sec:trigger} and~\ref{sec:prevent}).

\FloatBarrier

\section{Experimental setup}
\label{app:protocol}
\backto*{app:protocol}
\subsection{Models}
\label{app:models}
\backto*{app:models}

The base must carry a text spike, since a base without one has no spike channel for its adaptation to inherit.
A model's visual adaptation runs from the vision encoder and the projector to the training data and the training itself, so the base run on its own is the control for all of it.
Table~\ref{tab:fullroster} lists the 25 adapter-based vision--language models.
Among them are these models at 7B and 8B: Qwen2-VL-7B \citep{wang2024qwen2vl}, Molmo-7B-D
\citep{deitke2024molmo} and LLaVA-OneVision-7B \citep{li2024llavaonevision} on Qwen2-7B
\citep{yang2024qwen2}; Qwen2.5-VL-7B \citep{bai2025qwen25vl} and Qwen2.5-Omni-7B
\citep{qwen2025omni} on Qwen2.5-7B \citep{yang2024qwen25}; InternVL2-8B \citep{chen2024internvl}
on InternLM2.5-7B \citep{cai2024internlm2}; Qwen3-VL-8B \citep{qwen2025qwen3vl} and InternVL3.5-8B
\citep{internvl2025v35} on Qwen3-8B \citep{yang2025qwen3}; LLaVA-1.5-7B \citep{liu2024improved} on Vicuna-7B \citep{chiang2023vicuna}.
Four size series extend them: Qwen3-VL from 2B to 32B, InternVL3.5 from 4B to 38B,
Qwen2.5-VL from 3B to 72B, and LLaVA-1.5 at 7B and 13B.
Four text-only
models, Qwen3-8B and Qwen2.5-7B in their base and instruct variants, serve as inheritance probes,
together with the text-only base at every other width where one exists.
Molmo2-8B, Pixtral-12B, GLM-4.1V-9B-Base, Janus-Pro-7B, PaliGemma 2-3B and Idefics3-8B are given with their citations in \aref{app:extension}, and their bases were probed the same way.

Every spike channel in the table is the model's top text-spike dimension as recorded in its own
census, and it coincides with the base model's top text-spike dimension wherever a text-only base was probed.
Channel 4 is the top text-spike dimension of Qwen3-4B and is shared by both 4B LVLMs, channel 731 is the top text-spike dimension of both Qwen3-14B and Qwen3-32B, $\{4675, 3094\}$ are the top two of Qwen2.5-32B, and
$\{4743, 2100\}$ are the top two of Vicuna-13B, the LLaMA2-13B dimensions reported by
\citet{kang2025see}.
The pair $\{458, 2570\}$ shared by Qwen2-7B and Qwen2.5-7B, two distinct
bases, suggests that the two share a weight lineage, though the training base family itself is not
public.

\begin{apptable}
\caption{\textbf{Within a shared base the adaptation moves the rate but never flips the spiking status of a model.} The 25 LVLMs grouped by base language model, with the base's inherited spike channels: each LVLM's model line, clean $\mathrm{VSI}_\kappa$ at $\kappa = 1{,}000$ on $n = 300$ COCO images at the default input size with the Wilson $95\%$ interval, and the largest clean peak.\protect\backto{tab:fullroster}}
\label{tab:fullroster}
\begin{center}
\scriptsize
\setlength{\tabcolsep}{2.5pt}
\begin{tabular}{llllrr}
\toprule
\textbf{base LLM} & \textbf{spike channel(s)} & \textbf{LVLM} & \textbf{model line} & \textbf{clean $\mathrm{VSI}_\kappa$} & \textbf{max $\Psi$} \\
\midrule
Qwen2-7B & $\{458, 2570\}$ & Qwen2-VL-7B & Qwen2-VL & 0.0\% [0.0, 1.3] & 388 \\
 & & Molmo-7B-D & Molmo & 0.0\% [0.0, 1.3] & 392 \\
 & & LLaVA-OneVision-7B & LLaVA-OV & 0.0\% [0.0, 1.3] & 378 \\
\addlinespace[2pt]
Qwen2.5-3B & $\{318, 1874\}$ & Qwen2.5-VL-3B & Qwen2.5-VL & 0.0\% [0.0, 1.3] & 171 \\
\addlinespace[2pt]
Qwen2.5-7B & $\{458, 2570\}$ & Qwen2.5-VL-7B & Qwen2.5-VL & 0.0\% [0.0, 1.3] & 237 \\
 & & Qwen2.5-Omni-7B & Qwen2.5-Omni & 0.0\% [0.0, 1.3] & 172 \\
\addlinespace[2pt]
Qwen2.5-32B & $\{4675, 3094\}$ & Qwen2.5-VL-32B & Qwen2.5-VL & 52.0\% [46.4, 57.6] & 5{,}278 \\
\addlinespace[2pt]
Qwen2.5-72B & $\{5060, 2262\}$ & Qwen2.5-VL-72B & Qwen2.5-VL & 0.0\% [0.0, 1.3] & 520 \\
\addlinespace[2pt]
Qwen3-1.7B & $\{1793, 1999\}$ & Qwen3-VL-2B & Qwen3-VL & 0.0\% [0.0, 1.3] & 143 \\
\addlinespace[2pt]
Qwen3-4B & $4$ & Qwen3-VL-4B & Qwen3-VL & 75.3\% [70.2, 79.9] & 23{,}579 \\
 & & InternVL3.5-4B & InternVL3.5 & 16.3\% [12.6, 20.9] & 18{,}579 \\
\addlinespace[2pt]
Qwen3-8B & $2276$ & Qwen3-VL-8B & Qwen3-VL & 12.7\% [9.4, 16.9] & 21{,}292 \\
 & & InternVL3.5-8B & InternVL3.5 & 61.7\% [56.1, 67.0] & 32{,}050 \\
 & & Molmo2-8B & Molmo2 & 100.0\% [98.7, 100.0] & 27{,}449 \\
\addlinespace[2pt]
Qwen3-14B & $731$ & InternVL3.5-14B & InternVL3.5 & 69.0\% [63.6, 74.0] & 22{,}403 \\
\addlinespace[2pt]
Qwen3-32B & $731$ & Qwen3-VL-32B & Qwen3-VL & 46.7\% [41.1, 52.3] & 22{,}665 \\
 & & InternVL3.5-38B & InternVL3.5 & 1.7\% [0.7, 3.8] & 13{,}487 \\
\addlinespace[2pt]
InternLM2.5-7B & $\{3584, 3286\}$ & InternVL2-8B & InternVL2 & 0.0\% [0.0, 1.3] & 212 \\
\addlinespace[2pt]
Vicuna-7B & $\{2533, 1415\}$ & LLaVA-1.5-7B & LLaVA-1.5 & 0.0\% [0.0, 1.3] & 877 \\
\addlinespace[2pt]
Vicuna-13B & $\{2100, 4743\}$ & LLaVA-1.5-13B & LLaVA-1.5 & 0.0\% [0.0, 1.3] & 370 \\
\addlinespace[2pt]
Mistral-NeMo-12B & $\{3448, 3357\}$ & Pixtral-12B & Pixtral & 99.7\% [98.1, 99.9] & 23{,}130 \\
\addlinespace[2pt]
GLM-4-9B & $2477$ & GLM-4.1V-9B-Base & GLM-4.1V & 0.0\% [0.0, 1.3] & 236 \\
\addlinespace[2pt]
DeepSeek-LLM-7B & $\{655, 3557\}$ & Janus-Pro-7B & Janus-Pro & 0.0\% [0.0, 1.3] & 60 \\
\addlinespace[2pt]
Gemma-2-2B & $334$ & PaliGemma 2-3B-mix-448 & PaliGemma 2 & 0.0\% [0.0, 1.3] & 200 \\
\addlinespace[2pt]
Llama-3.1-8B & $\{788, 1384\}$ & Idefics3-8B & Idefics3 & 0.0\% [0.0, 1.3] & 29 \\
\bottomrule
\end{tabular}
\end{center}
\end{apptable}
\subsection{Compute}
\label{app:compute}
\backto*{app:compute}

Every run is single-GPU, on one NVIDIA B200 with 183\,GB, driver 580.82.07, PyTorch 2.12 and CUDA 12.8.
Weights are loaded in bfloat16, so the largest model of the study, Qwen2.5-VL-72B, fits on one card without sharding.
Readings are taken with hidden states returned, which dominates the cost of a forward at these sizes.

The wall-clock figures below are representative, measured on the runs reported here.
A forward with hidden states takes about $0.1$\,s on an 8B model, so the corruption panel of Table~\ref{tab:corrpanel}, 200 questions under seven conditions, takes under three minutes a model.
The intervention under corruption of Table~\ref{tab:preventcorr}, 50 questions on the clean image and under six operators, read with and without the edit, takes about one minute.
The region edits of Tables~\ref{tab:steer} and~\ref{tab:blocks} take about 90\,s over 150 images.
The free-form intervention of Table~\ref{tab:preventfree}, 297 VQAv2 questions and 300 captions each generated four times, takes about an hour a model, generation being the cost.

The attack dominates the total, and its cost is set by the step budget rather than by the image.
A step is one forward and one backward, at most 300 of them and 600 in the fine-step runs of Molmo2-8B and Pixtral-12B (\aref{app:attack}), and the run stops at the first success.
A success therefore costs a few steps and a failure costs the whole budget, which is why a model whose state does not move is the expensive one to establish.
On a 4B model a step takes about $0.07$\,s, so a failed image costs about $20$\,s and a successful one a second or two.
Over the attack runs reported in the tables, $3{,}900$ attacked images and $218{,}259$ steps, the attack takes about $23$ GPU-hours of measured wall-clock, each run counted once.
The median successful attack takes six steps.
Molmo2-8B and InternVL3.5-38B take $18$ of the $23$ hours, and the images on which the attack fails take $9$ of them, since a failure runs to the budget where a success stops.
The measurement and intervention runs add roughly $30$ more, so reproducing the results in this paper costs on the order of $53$ B200-hours, excluding the model downloads and the exploratory runs that the final numbers replace.
The rental equivalent is about \$360 at the median on-demand B200 rate of September 2026, \$6.79 an hour, of which the attack is about \$150.
Specialist clouds quote \$5 to \$8 an hour and the hyperscalers \$14 to \$16, so the figure runs from roughly \$270 to \$740 depending on the provider.
\subsection{Data and preprocessing}
\label{app:data}
\backto*{app:data}

Images come from COCO val2014 \citep{lin2014microsoft}, sampled in a fixed pseudo-random order given by a hash of a fixed seed and the filename, and the 300 images so drawn per model are the census.
The POPE-derived evaluation manifest, disjoint from the census, carries the perturbation and location runs.
The location runs take its first 300 distinct images, the edit set, and the perturbation and noise runs its first 200 question rows, over 161 distinct images.
Every run recomputes $\Psi_{\mathrm{peak}}$ on its own forward, and each table names its set.
The input size is the square pixel size an image is resized to before the model's processor runs.
It fixes the pixel count across models and the image-token count within one, Qwen3-VL merging to 32 pixels per token so that 384, 512 and 768 give 144, 256 and 576 tokens.
The censuses use each model's default input size, and every number in the paper names the input size it was measured at.
\subsection{Measurement protocol}
\label{app:measure}
\backto*{app:measure}

For every forward pass we read the residual stream after each decoder block, restricted to the image tokens.
An image token already at $100$ or above on an inherited spike channel when it enters the language model is ViT-emerged, and is not counted as a spike token.
The eligible set is $I_{\mathrm{eligible}} = \{i \in I : \max_{d \in D_{\mathrm{spike}}} \lvert \vb{x}_i^{(0)}[d] \rvert < 100\}$.
The last entry the \texttt{transformers} library returns is the final norm's output rather than the last block's, and we drop it, which leaves $L$ states, the embeddings and the outputs of blocks $0$ to $L-2$.
In principle the scan covers every state.
To save compute, we compute $\Psi_i$ for every eligible image token only at the states in the deep $60\%$ of those, $\mathcal{L} = \{\lfloor 0.4 L \rfloor,\ldots,L-1\}$.
Reading every state on the 300 census images of the 25 models confirms that the deep range is enough, since it finds every spiking image that the full scan finds except one of $7{,}500$.
In 24 models, no image token first reaches $\Psi \ge 1{,}000$ below the range.
Pixtral-12B writes its spike at block 2, and the spike holds into the range on 299 of its 300 images, so its incidence reads $99.7\%$ rather than $100\%$.
$\Psi_{\mathrm{peak}}$ maximizes over $I_{\mathrm{eligible}}$ and $\mathcal{L}$, on the base's inherited dimensions.
Reading it on whichever channel of the token is largest moves no image across $1{,}000$ in any of the 25 models.
The level sits in a measured gap.
The fifteen non-spiking models top out at $877$ against $5{,}278$ at the bottom of the spiking ten, and across the $7{,}500$ clean census images none falls between $970$ and $1{,}186$.
The level sits above the band $500 \le \Psi < 1{,}000$ rather than at its foot because LLaVA-1.5-7B, whose text switch writes image tokens in the hundreds, has 294 of its 300 images in that band (Table~\ref{tab:twoswitch}).

\textbf{Visual-spike incidence.}\quad
For a model $M$, a dataset $D$, and a level $\kappa$, the incidence defined in Section~\ref{sec:chanregimes} is
\begin{equation}
\label{eq:vsi}
\mathrm{VSI}_\kappa(M, D)
=
\frac{1}{\lvert D \rvert}
\sum_{\vb{x} \in D}
\mathbf{1}\!\left[
\exists\, i \in I_{\mathrm{eligible}},\, \ell \in \mathcal{L}
\ \text{s.t.}\
\Psi_i^{(\ell)} \ge \kappa,\quad
\max_{d \in D_{\mathrm{spike}}}
\left|\vb{x}_i^{(\ell)}[d]\right|
\ge 100
\right].
\end{equation}

\textbf{What the criterion singles out.}\quad At $\Psi \ge 1{,}000$ the criterion usually names one token.
At each spiking image's peak state the peak token's $\Psi$ stands $18$ to $378$ times above the runner-up's in nine of the ten spiking models, exactly one token passes on $51$ to $100\%$ of images in those nine, and the ordinary tokens' 99th percentile is $31$ to $203$ (Table~\ref{tab:separation}).
Molmo2-8B spikes on a median of seven tokens per image.
In the fifteen non-spiking models the peak stands at most $2.1$ times the runner-up, so no token stands apart at the hundreds either, the two LLaVA-1.5 models included, with medians of four and zero tokens per image above $100$.

\begin{apptable}
\caption{\textbf{At the criterion one token stands apart, except in Molmo2-8B.
At the hundreds none does.} 300 census images per model at each image's peak state, spiking images in the spiking models.
Columns: peak token's $\Psi$ over the runner-up's and over the 99th percentile of the other tokens; tokens at or above the criterion per image, median and maximum; share with exactly one; ordinary tokens' $\Psi$, median, 99th percentile and maximum.
Entries are per-image values, median over images.
The criterion is $100$ in the first block, $1{,}000$ in the second.\protect\backto{tab:separation}}
\label{tab:separation}
\centering\footnotesize\setlength{\tabcolsep}{3pt}
\begin{tabular}{lrrrrr}
\toprule
model & peak / runner-up & peak / 99th pct & tokens above & exactly one & ordinary $\Psi$ \\
\midrule
Qwen2-VL-7B & 1.0 & 1.3 & 7, 12 & 0\% & 39, 261, 325 \\
Molmo-7B-D & 1.0 & 4.9 & 6, 14 & 1\% & 44, 71, 335 \\
LLaVA-OV-7B & 1.0 & 5.4 & 4, 4 & 0\% & 23, 62, 328 \\
Qwen2.5-VL-3B & 1.1 & 1.4 & 2, 36 & 31\% & 58, 100, 131 \\
Qwen2.5-VL-7B & 1.0 & 1.1 & 233, 425 & 2\% & 110, 172, 184 \\
Qwen2.5-Omni-7B & 1.1 & 1.3 & 0, 80 & 0\% & 51, 108, 128 \\
Qwen2.5-VL-72B & 1.5 & 2.1 & 215, 390 & 0\% & 111, 199, 267 \\
Qwen3-VL-2B & 1.0 & 1.1 & 24, 54 & 0\% & 81, 123, 127 \\
LLaVA-1.5-7B & 1.1 & 5.0 & 4, 8 & 2\% & 4, 125, 564 \\
LLaVA-1.5-13B & 1.2 & 9.8 & 0, 2 & 7\% & 5, 22, 184 \\
InternVL2-8B & 1.1 & 1.7 & 2, 72 & 17\% & 49, 108, 127 \\
GLM-4.1V-9B-Base & 2.1 & 2.3 & 2, 68 & 34\% & 52, 93, 103 \\
Janus-Pro-7B & 1.1 & 1.7 & 0, 0 & 0\% & 7, 23, 37 \\
PaliGemma 2-3B & 1.0 & 2.4 & 8, 292 & 0\% & 30, 75, 180 \\
Idefics3-8B & 1.1 & 1.3 & 0, 0 & 0\% & 7, 13, 16 \\
\midrule
Qwen2.5-VL-32B & 17.6 & 26.4 & 1, 1 & 100\% & 75, 147, 223 \\
Qwen3-VL-4B & 25.9 & 447 & 1, 4 & 51\% & 22, 40, 639 \\
InternVL3.5-4B & 378.3 & 430 & 1, 2 & 88\% & 15, 33, 37 \\
Qwen3-VL-8B & 271.1 & 318 & 1, 2 & 97\% & 26, 46, 50 \\
InternVL3.5-8B & 320.0 & 700 & 1, 4 & 61\% & 13, 31, 40 \\
InternVL3.5-14B & 34.5 & 113 & 1, 7 & 55\% & 78, 114, 134 \\
Qwen3-VL-32B & 86.9 & 102 & 1, 3 & 86\% & 123, 160, 181 \\
InternVL3.5-38B & 47.6 & 51.7 & 1, 1 & 100\% & 103, 137, 142 \\
Molmo2-8B & 1.1 & 538 & 7, 12 & 0\% & 22, 40, 19{,}266 \\
Pixtral-12B & 21.7 & 84.7 & 1, 5 & 72\% & 25, 203, 804 \\
\bottomrule
\end{tabular}
\end{apptable}

\textbf{Attention received.}\quad Attention is read on three tasks at a $512$ input size.
The 200 POPE questions of the evaluation manifest carry the suffix ``Answer yes or no.''.
One VQAv2 validation question per census image, the one with the lowest identifier whose answer type is not yes/no, carries ``Answer the question using a single word or phrase.''.
The census images under ``Describe this image in one short sentence.'' are scored by CIDEr against COCO's five references.
The model answers greedily, up to 8, 10 and 40 new tokens, and one teacher-forced forward over the prompt and the answer is read with eager attention.
The share defined in Section~\ref{sec:setup} is the head mean over the text queries after the image, renormalized over the image tokens, and a forward has an attention sink when that share reaches $0.3$ at some deep block.

\textbf{Attention implementation and the criterion.}\quad Attention implementations differ numerically and the criterion sits on a threshold, so a forward can change side between them.
Across the eight spiking models whose images do not all spike, $2$ to $35$ of the 200 evaluation forwards change side between PyTorch's scaled-dot-product implementation and the eager one, and none in Molmo2-8B and Pixtral-12B, which spike on every forward under both (Table~\ref{tab:kernel}). The difference decides whether any token is aimed at the trigger: over the forwards that change side, the best-aligned image token reaches the switch at a median cosine of $0.12$ to $0.38$ under the implementation where the image spikes and $0.02$ to $0.23$ under the other.
The census incidences are read with the default implementation and the attention readings with the eager one.
Table~\ref{tab:preventatk} re-reads the attack's saved images on the intervention's own forward, where $38$ of $40$, $33$ of $38$ and $37$ of $40$ of the attack's successes on the three Qwen3-VL models spike, and up to four of the 40 eligible images spike clean.

\begin{table}[!htbp]
\caption{\textbf{Attention implementations change the spiking status of a few forwards, through the alignment with the trigger.} Per LVLM, the 200 POPE evaluation questions, each read under scaled-dot-product and eager attention: forwards that spike under each, those that spike under one alone, over the forwards that change side the median largest cosine of an image token with the trigger at the switch's feed-forward input where the image spikes and under the other implementation, and the median $\Psi$ where it spikes.\protect\backto{tab:kernel}}
\label{tab:kernel}
\begin{center}
\footnotesize
\setlength{\tabcolsep}{4pt}
\resizebox{\linewidth}{!}{\begin{tabular}{lrrrrr}
\toprule
LVLM & questions & spike, SDPA, eager & one alone, SDPA, eager & alignment, spiking, other & $\Psi$ \\
\midrule
Qwen3-VL-4B & 200 & 142, 136 & 7, 1 & 0.38, 0.04 & 14{,}605 \\
Qwen3-VL-8B & 200 & 16, 16 & 5, 5 & 0.27, 0.02 & 11{,}320 \\
Qwen3-VL-32B & 200 & 70, 81 & 12, 23 & 0.25, 0.05 & 15{,}329 \\
Qwen2.5-VL-32B & 200 & 98, 100 & 0, 2 & 0.16, 0.06 & 2{,}891 \\
InternVL3.5-4B & 200 & 30, 30 & 2, 2 & 0.27, 0.23 & 2{,}509 \\
InternVL3.5-8B & 200 & 106, 105 & 8, 7 & 0.18, 0.09 & 2{,}990 \\
InternVL3.5-14B & 200 & 113, 110 & 3, 0 & 0.17, 0.10 & 6{,}074 \\
InternVL3.5-38B & 200 & 10, 8 & 2, 0 & 0.12, 0.07 & 2{,}816 \\
Molmo2-8B & 200 & 200, 200 & 0, 0 & -- & -- \\
Pixtral-12B & 200 & 200, 200 & 0, 0 & -- & -- \\
\bottomrule
\end{tabular}}
\end{center}
\end{table}

\subsection{Corruptions and their grades}
\label{app:ops}
\backto*{app:ops}

Gaussian noise is added in $[0, 1]$ pixel units before the
model's processor runs.
Noise is a value-space operator, so its $\sigma$ is never rescaled with the
input size.
The noise conditions use five grades, s1 to s5, at $\sigma = 0.05$, $0.10$, $0.19$, $0.30$ and $0.45$.
Each image draws one noise seed, shared by the five grades, so a grade differs from the next by scale alone.
The panel of Table~\ref{tab:corrpanel} uses six operators at one grade each.
They are Gaussian noise at $\sigma = 0.45$, shot noise drawn as Poisson counts at 3 per unit of intensity, Gaussian blur at $\sigma = 3.2$ pixels scaled with the input size, JPEG compression at quality 10, a brightness rise of $0.4$ in the HSV value channel, and contrast reduced to $0.1$ of each channel's spread about its mean.

\FloatBarrier
\section{When and how visual spikes form}
\label{app:mechanism}
\backto*{app:mechanism}
This appendix gives the additional analyses for Section~\ref{sec:mech}.

\FloatBarrier
\subsection{Spike criteria and attention}
\label{app:census}
\backto*{app:census}

\begin{appfigure}
\begin{center}
\includegraphics[width=0.85\textwidth]{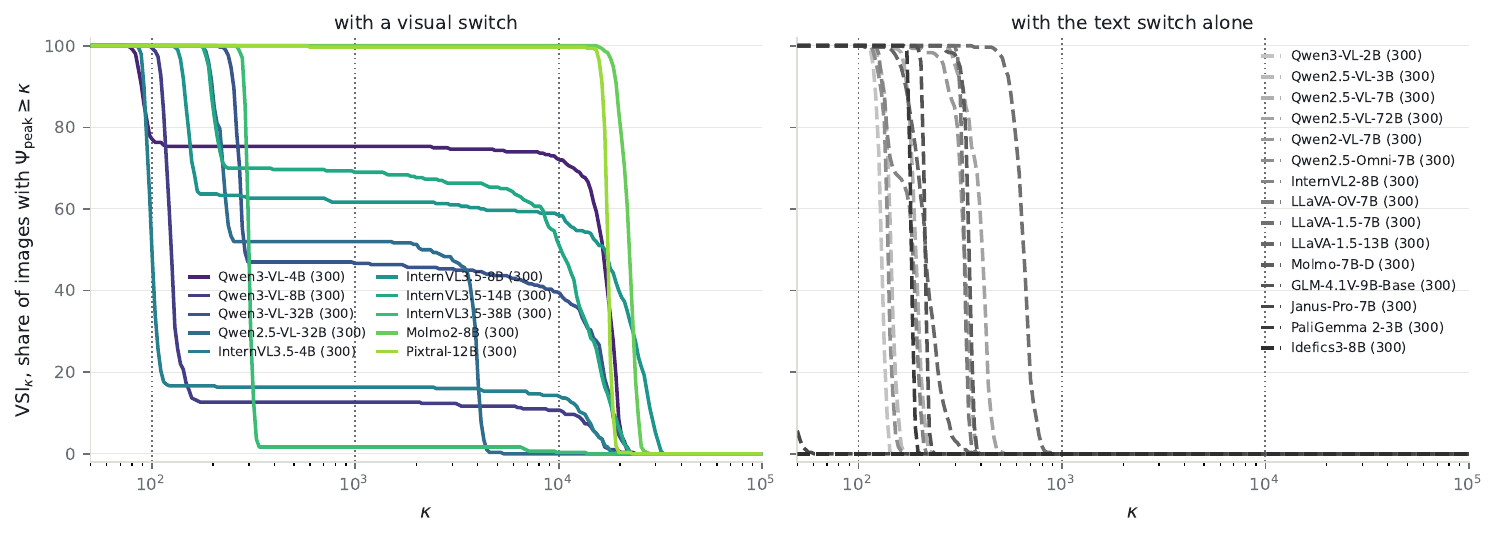}
\end{center}
\caption{\textbf{The two groups are separated by a wide range of levels, not by a tuned one.} $\mathrm{VSI}_\kappa$, the share of census images whose peak reaches $\Psi \ge \kappa$, over $\kappa$ from 50 to $10^5$, the 300 census images of each of the 25 models, the ten with a visual switch on the left and the fifteen without one on the right, dotted rules at $10^2$, $10^3$ and $10^4$.\protect\backto{fig:regimes}}
\label{fig:regimes}
\end{appfigure}

\begin{apptable}
\caption{\textbf{Only the criterion at $1{,}000$ separates the two groups.
The two lower criteria pass in 23 of the 25 models.} Four criteria on the 300 census images per model: the share of images on which some image token passes, with the mean share of the image's tokens passing in parentheses.
The three activation criteria are read on the census forward, the attention on the captioning forward over the same images, and on the census forward for Idefics3-8B.\protect\backto{tab:regimes}}
\label{tab:regimes}
\begin{center}
\scriptsize
\setlength{\tabcolsep}{3pt}
\begin{tabular}{lllrrrrl}
\toprule
\textbf{LVLM} & \textbf{base LLM} & \textbf{spike channel(s)} & $\phi \ge 20$ & $\Psi \ge 100$ & $\Psi \ge 1{,}000$ & \textbf{attention} & \textbf{$\mathrm{VSI}_\kappa$, 300 images} \\
\midrule
Qwen2-VL-7B & Qwen2-7B & $\{458, 2570\}$ & 100 (99) & 100 (11) & 0 & 0 & 0.0 [0.0, 1.3] \\
Molmo-7B-D & Qwen2-7B & $\{458, 2570\}$ & 100 (83) & 100 (1.5) & 0 & 10 ($<$0.1) & 0.0 [0.0, 1.3] \\
LLaVA-OneVision-7B & Qwen2-7B & $\{458, 2570\}$ & 100 (99) & 100 (19) & 0 & 0 & 0.0 [0.0, 1.3] \\
Qwen2.5-VL-3B & Qwen2.5-3B & $\{318, 1874\}$ & 100 (100) & 100 (6.8) & 0 & 0 & 0.0 [0.0, 1.3] \\
Qwen2.5-VL-7B & Qwen2.5-7B & $\{458, 2570\}$ & 100 (100) & 100 (68) & 0 & 0 & 0.0 [0.0, 1.3] \\
Qwen2.5-Omni-7B & Qwen2.5-7B & $\{458, 2570\}$ & 100 (98) & 100 (9.3) & 0 & 0 & 0.0 [0.0, 1.3] \\
Qwen2.5-VL-72B & Qwen2.5-72B & $\{5060, 2262\}$ & 100 (100) & 100 (100) & 0 & 0 & 0.0 [0.0, 1.3] \\
Qwen3-VL-2B & Qwen3-1.7B & $\{1793, 1999\}$ & 100 (100) & 100 (17) & 0 & 0 & 0.0 [0.0, 1.3] \\
InternVL2-8B & InternLM2.5-7B & $\{3584, 3286\}$ & 100 (100) & 100 (16) & 0 & 0 & 0.0 [0.0, 1.3] \\
LLaVA-1.5-7B & Vicuna-7B & $\{2533, 1415\}$ & 100 (1.3) & 100 (0.7) & 0 & 0 & 0.0 [0.0, 1.3] \\
LLaVA-1.5-13B & Vicuna-13B & $\{2100, 4743\}$ & 100 (1.4) & 23 (0.1) & 0 & 4 ($<$0.1) & 0.0 [0.0, 1.3] \\
GLM-4.1V-9B-Base & GLM-4-9B & $2477$ & 100 (100) & 100 (45) & 0 & 0 & 0.0 [0.0, 1.3] \\
Janus-Pro-7B & DeepSeek-LLM-7B & $\{655, 3557\}$ & 0 & 0 & 0 & 0 & 0.0 [0.0, 1.3] \\
PaliGemma 2-3B & Gemma-2-2B & $334$ & 100 (99) & 100 (14) & 0 & 16 ($<$0.1) & 0.0 [0.0, 1.3] \\
Idefics3-8B & Llama-3.1-8B & $\{788, 1384\}$ & 0 & 0 & 0 & 0 & 0.0 [0.0, 1.3] \\
InternVL3.5-38B & Qwen3-32B & $731$ & 100 (100) & 100 (100) & 2 ($<$0.1) & 0 ($<$0.1) & 1.7 [0.7, 3.8] \\
Qwen3-VL-32B & Qwen3-32B & $731$ & 100 (100) & 100 (100) & 47 (0.2) & 41 (0.2) & 46.7 [41.1, 52.3] \\
Qwen2.5-VL-32B & Qwen2.5-32B & $\{4675, 3094\}$ & 100 (100) & 100 (98) & 52 (0.2) & 56 (0.2) & 52.0 [46.4, 57.6] \\
InternVL3.5-4B & Qwen3-4B & $4$ & 100 (90) & 51 (1.1) & 16 (0.1) & 19 (0.1) & 16.3 [12.6, 20.9] \\
Qwen3-VL-4B & Qwen3-4B & $4$ & 100 (87) & 77 (0.6) & 75 (0.5) & 76 (0.3) & 75.3 [70.2, 79.9] \\
Qwen3-VL-8B & Qwen3-8B & $2276$ & 100 (98) & 99 (7.6) & 13 ($<$0.1) & 10 ($<$0.1) & 12.7 [9.4, 16.9] \\
InternVL3.5-8B & Qwen3-8B & $2276$ & 100 (99) & 100 (76) & 62 (0.4) & 57 (0.2) & 61.7 [56.1, 67.0] \\
InternVL3.5-14B & Qwen3-14B & $731$ & 100 (100) & 100 (96) & 69 (0.4) & 63 (0.3) & 69.0 [63.6, 74.0] \\
Molmo2-8B & Qwen3-8B & $2276$ & 100 (90) & 100 (11) & 100 (0.9) & 34 (0.1) & 100.0 [98.7, 100.0] \\
Pixtral-12B & Mistral-NeMo-12B & $\{3448, 3357\}$ & 100 (88) & 100 (0.6) & 100 (0.1) & 100 (0.1) & 99.7 [98.1, 99.9] \\
\bottomrule
\end{tabular}
\end{center}
\end{apptable}

\textbf{The channel criterion on the 25 models.}\quad The channel criterion of \citet{kang2025see}, $\phi \ge 20$ over the inherited spike dimensions, is capped at $\sqrt{D}$, so a threshold that separates tokens on one model cannot be raised to separate them on another.
A token's root mean square is $\mathrm{RMS}(\vb{x}_i) = \lVert \vb{x}_i \rVert / \sqrt{D}$, so $\phi_i = \sqrt{D}\, \max_{d \in D_{\mathrm{spike}}} \lvert \vb{x}_i[d] \rvert / \lVert \vb{x}_i \rVert = \sqrt{D\, s_i}$ with $s_i = \max_{d \in D_{\mathrm{spike}}} \vb{x}_i[d]^2 / \sum_{k} \vb{x}_i[k]^2$, the share of the token's energy on its peak channel.
One channel's square cannot exceed the sum over all $D$ channels, so $s_i \le 1$ and $\phi_i \le \sqrt{D}$, with equality when the whole token sits on one spike channel.
It was set on LLaVA-1.5-7B, where it selects under $2\%$ of the image tokens, as it does on LLaVA-1.5-13B.
\citet{kang2025see} study six models, LLaVA-1.5-7B, LLaVA-1.5-13B, LLaVA-1.5-HD-13B, VILA-13B, Qwen2-VL-7B and InternVL2-8B, and the four of those that are among our 25 are all non-spiking under the original criterion.
The tokens the criterion selects there carry an elevated value on the inherited channels and lead the image's attention without reaching the criterion. This is the reading the fifteen non-spiking models give.
In 21 of the 25 models it selects $83$ to $100\%$, because there the inherited channel is elevated on every image token, the cause \citet{yoo2026nature} give on two omni-modal models, against at most $1\%$ at $\Psi \ge 1{,}000$ in the spiking models (Table~\ref{tab:regimes}).
On Janus-Pro-7B and Idefics3-8B it selects none, their image tokens reaching at most $\phi = 19.7$ and $12.3$.
At $\phi \ge 50$ it takes $99.9\%$ of the tokens in two models, none in five, under $1.2\%$ in eleven and $2$ to $93\%$ in the other seven, spiking and non-spiking alike.

\textbf{No single threshold recovers the regimes.}\quad A threshold on $\phi$ selects a spiking model's spike token only when it sits at or below that token's $\phi$, and it selects no token of a non-spiking model only when it sits above every image token's $\phi$ there.
We measure both on the 300 census images of all 25 models, ten spiking and fifteen non-spiking.
The spike token's median $\phi$ at its peak state is $47.0$ in Qwen2.5-VL-32B and lies within $1.3$ of $\sqrt{D}$ in the other nine, from $49.9$ to $70.6$.
The largest $\phi$ of an image token in a non-spiking model runs from $12.3$ to $86.1$, and it exceeds $47.0$ in ten of the fifteen.
A single threshold would therefore have to be at most $47.0$ and above $86.1$.
A spike token scores near its own model's cap, and the cap of a narrow model lies below the ordinary tokens of a wider one, $50.6$ for Qwen3-VL-4B against the $86.1$ reached in Qwen2.5-VL-72B.
A threshold per hidden size fails on the same measurement, since Qwen2.5-VL-32B and LLaVA-1.5-13B share $D = 5{,}120$, where the first needs a threshold of at most $47.0$ and the second one above $58.7$.

\textbf{A threshold per model does not recover the regimes.}\quad Since $\phi$ is capped at $\sqrt{D}$, the criterion can be rescaled to the same fraction of each model's cap that $20$ is of LLaVA-1.5's, or calibrated per model to the share of tokens $\phi \ge 20$ selects there.
Neither reading separates the ten spiking from the fifteen non-spiking models.
Rescaled, the criterion still passes $26$ to $99\%$ of the deep image tokens outside the two LLaVA-1.5 models, Janus-Pro-7B and Idefics3-8B, where it passes at most $1.3\%$, and it passes more than $\phi \ge 20$ does wherever $D$ is smaller than LLaVA-1.5-7B's, since a smaller cap lowers the threshold.
Calibrated instead to the $1.3\%$ of deep image tokens $\phi \ge 20$ selects on LLaVA-1.5-7B, the threshold a model needs runs from $0.57$ to $0.90$ of its cap in the spiking models and from $0.09$ to $0.85$ in the non-spiking ones, so the two ranges overlap.
A threshold fitted per model also makes the criterion a ranking, which returns its share of tokens in every model and so cannot report that a model raises none, where $\Psi$ has a floor and does.

\textbf{Attention on three tasks.}\quad Table~\ref{tab:attntasks} gives, per model and task, the forwards on which an attention sink forms, the top image token's median share, the forwards on which an image token passes $0.3$ at a single layer-head pair of the raw attention, and the task's outcome on the same forwards.
Of the $4{,}184$ spiking forwards of the ten models, $3{,}529$ have an attention sink, and it sits on a spike token on all but three.
The single-pair reading is reached on nearly every forward of LLaVA-1.5-13B, on 64 to 280 of LLaVA-OV-7B's, on up to 226 of PaliGemma 2-3B's and 36 of Molmo-7B-D's, and on at most 7 per task elsewhere. What it misses is persistence.
On all 25 models, the largest share of layer-head pairs any image token holds is $1.4\%$, where the text spike token holds $49$ to $98\%$.

\begin{table}[!htbp]
\caption{\textbf{An attention sink forms on the spike token, and rarely without one.
Molmo2-8B does on a third at most.} Per LVLM and task: forwards with an attention sink, an image token at $0.3$ or more of the image's head-mean attention in a deep block, over spiking then non-spiking forwards in the spiking models, all forwards in the others; top image token's median share; forwards passing $0.3$ at one layer-head pair of raw attention; POPE and VQAv2 accuracy and CIDEr.
Read on the text queries after the image.\protect\backto{tab:attntasks}}
\label{tab:attntasks}
\begin{center}\scriptsize
\setlength{\tabcolsep}{2.5pt}
\resizebox{\textwidth}{!}{\begin{tabular}{llllrrl}
\toprule
LVLM & POPE & VQAv2 & captions & top share & single pair & outcome \\
\midrule
Qwen2.5-VL-32B & 99/99, 0/101 & 166/167, 0/130 & 169/169, 0/131 & 0.30 / 0.32 / 0.24 & 4, 7, 0 & 0.870, 0.715, 0.23 \\
Qwen3-VL-4B & 133/136, 1/64 & 231/233, 0/64 & 225/234, 0/66 & 0.25 / 0.23 / 0.21 & 0, 0, 0 & 0.915, 0.722, 0.19 \\
InternVL3.5-4B & 30/32, 0/168 & 57/58, 2/239 & 57/58, 0/242 & 0.40 / 0.38 / 0.37 & 0, 5, 0 & 0.845, 0.658, 0.61 \\
Qwen3-VL-8B & 15/16, 3/184 & 31/32, 2/265 & 30/34, 0/266 & 0.21 / 0.18 / 0.19 & 0, 0, 0 & 0.890, 0.734, 0.35 \\
InternVL3.5-8B & 107/110, 0/90 & 173/181, 1/116 & 171/182, 0/118 & 0.41 / 0.34 / 0.36 & 0, 6, 0 & 0.860, 0.676, 0.71 \\
Qwen3-VL-32B & 74/81, 0/119 & 119/127, 3/170 & 118/127, 4/173 & 0.20 / 0.19 / 0.16 & 1, 3, 6 & 0.895, 0.739, 0.15 \\
InternVL3.5-14B & 108/114, 1/86 & 184/192, 0/105 & 188/195, 2/105 & 0.30 / 0.28 / 0.25 & 0, 3, 0 & 0.865, 0.676, 0.60 \\
InternVL3.5-38B & 8/9, 0/191 & 1/3, 0/294 & 1/3, 0/297 & 0.14 / 0.07 / 0.06 & 0, 2, 2 & 0.845, 0.738, 0.80 \\
Molmo2-8B & 48/200, -- & 89/297, -- & 102/300, -- & 0.20 / 0.20 / 0.21 & 0, 1, 1 & 0.900, 0.686, 0.20 \\
Pixtral-12B & 200/200, -- & 296/296, 0/1 & 299/299, 0/1 & 0.53 / 0.56 / 0.52 & 2, 0, 0 & 0.790, 0.619, 0.12 \\
\midrule
Qwen2-VL-7B & 0/200 & 0/297 & 0/300 & 0.05 / 0.05 / 0.06 & 0, 3, 0 & 0.890, 0.721, 0.71 \\
Molmo-7B-D & 0/200 & 4/297 & 31/300 & 0.02 / 0.03 / 0.03 & 23, 36, 36 & 0.900, 0.609, 0.06 \\
LLaVA-OV-7B & 0/200 & 0/297 & 0/300 & 0.04 / 0.04 / 0.06 & 130, 64, 280 & 0.925, 0.727, 1.44 \\
Qwen2.5-VL-3B & 0/200 & 1/297 & 0/300 & 0.08 / 0.05 / 0.06 & 0, 4, 0 & 0.875, 0.716, 1.14 \\
Qwen2.5-VL-7B & 0/200 & 1/297 & 0/300 & 0.06 / 0.05 / 0.05 & 0, 2, 0 & 0.860, 0.717, 0.41 \\
Qwen2.5-Omni-7B & 1/200 & 0/297 & 0/300 & 0.06 / 0.05 / 0.05 & 0, 1, 0 & 0.885, 0.750, 0.55 \\
Qwen3-VL-2B & 0/200 & 0/297 & 0/300 & 0.05 / 0.04 / 0.08 & 0, 0, 0 & 0.915, 0.674, 0.31 \\
LLaVA-1.5-7B & 0/200 & 1/297 & 0/300 & 0.07 / 0.07 / 0.07 & 0, 0, 0 & 0.870, 0.661, 1.26 \\
LLaVA-1.5-13B & 17/200 & 50/297 & 12/300 & 0.14 / 0.14 / 0.13 & 200, 296, 299 & 0.855, 0.687, 1.32 \\
InternVL2-8B & 0/200 & 0/297 & 0/300 & 0.06 / 0.06 / 0.07 & 0, 0, 0 & 0.845, 0.675, 0.49 \\
Qwen2.5-VL-72B & 0/200 & 0/297 & 0/300 & 0.03 / 0.03 / 0.03 & 0, 4, 1 & 0.865, 0.718, 0.45 \\
GLM-4.1V-9B-Base & 1/200 & 2/297 & 0/300 & 0.08 / 0.07 / 0.10 & 1, 0, 0 & 0.895, 0.674, 0.75 \\
Janus-Pro-7B & 0/200 & 0/297 & 0/300 & 0.05 / 0.05 / 0.06 & 0, 0, 0 & 0.885, 0.682, 0.69 \\
PaliGemma 2-3B & 0/200 & 0/297 & 47/300 & 0.03 / 0.03 / 0.05 & 12, 0, 226 & 0.895, 0.769, 1.55 \\
Idefics3-8B & 0/200 & 1/297 & 1/300 & 0.09 / 0.08 / 0.11 & 4, 0, 3 & 0.875, 0.676, 0.12 \\
\bottomrule
\end{tabular}}
\end{center}
\end{table}

\FloatBarrier
\subsection{Visual switches and trigger directions}
\label{app:drive}
\backto*{app:drive}
\begin{appfigure}
\begin{center}
\includegraphics[width=\linewidth]{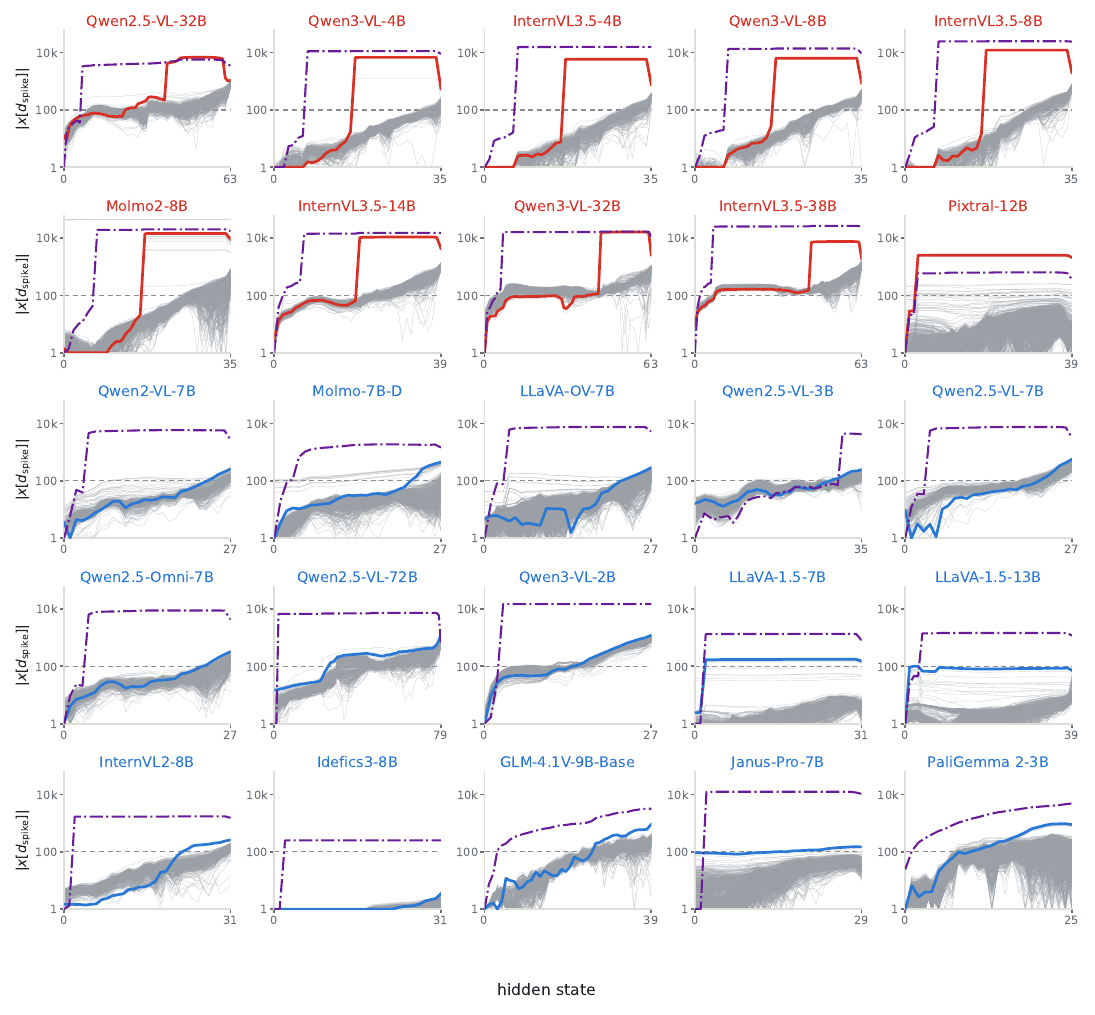}
\end{center}
\caption{\textbf{Every model's switch, or its absence, on one page.} Figure~\ref{fig:switchfour} for all 25 LVLMs, the ten spiking models first, each on the census image whose peak is that model's median.
The spike-channel magnitude across hidden states on a log scale.
Every image token is a grey line.
The spike token is red in the ten spiking models, and the largest image token is blue in the fifteen without one.
The dash-dot line is the text spike token, and the dashed line the criterion's floor at $100$.\protect\backto{fig:switchall}}
\label{fig:switchall}
\end{appfigure}

\begin{figure}[!htbp]
\begin{center}
\includegraphics[width=\linewidth]{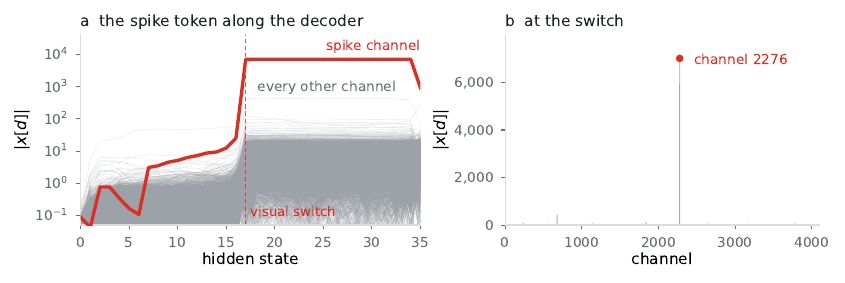}
\end{center}
\caption{\textbf{How much of the spike token is the spike.} Figure~\ref{fig:switchall} fixes the spike channel and runs over the tokens.
This one fixes the token that spikes and runs over its $4{,}096$ channels, under Qwen3-VL-8B on Figure~\ref{fig:one}'s image.
Left, every channel along the decoder on a log scale, the spike channel in red.
Right, the same token at the state the visual switch writes into, where the spike channel reaches $7{,}008$ against the next channel's $436$ and the median's $4.6$.}
\label{fig:chan}
\end{figure}

\textbf{The trigger, in full.}\quad Block $\ell$'s feed-forward maps its input $\vb{x}$ to $W \vb{h}(\vb{x})$ with $\vb{h}(\vb{x}) = \sigma(G\tilde{\vb{x}}) \odot U\tilde{\vb{x}}$ and $\tilde{\vb{x}} = \vb{x} / \mathrm{RMS}(\vb{x})$.
The gate and up matrices $G$ and $U$ already include the preceding norm's weights $\vb{g}$, folded into their columns, and $\sigma$ is SiLU in the Qwen bases.
The trigger $\vb{t}$ of Equation~\ref{eq:trigger} maximizes the write in the eruption's sign over the unit sphere, by Adam at learning rate $0.05$ for $300$ to $500$ steps from random starts and from the leading eigendirection of the sign-adjusted quadratic form of the same weights, keeping the best.
The two families of starts reach the same direction on every channel at cosine $1.000$, where random starts alone had stalled on one channel.
No image enters the optimization.
The sign $s_d$ is that of the spike token's spike channel after the switch, measured on the spiking census forwards of one LVLM per base.
It is the same on every one of those forwards, $765$ of $765$ over the five Qwen bases.
The trigger is then applied to the stored tokens entering the block, and the alignment is $\langle \vb{x} / \lVert \vb{x} \rVert, \vb{t} \rangle$.

\textbf{How the text spike is written, in full.}\quad \citet{sun2026spike} give the mechanism in three steps.
The first token attends only to itself, so its early blocks carry it toward a direction $\vb{s}^{\ast}$ set by the weights.
At the text switch, with the activation near the identity on such tokens, the output on channel $d$ is the quadratic form $\tilde{\vb{x}}^{\top} S_d \tilde{\vb{x}}$ of the gate, up and down projections with the norm's weight folded in.
In the blocks they study, one eigenvalue of $S_d$ exceeds the rest by orders of magnitude, so the write is $\lambda^{\ast} (\vb{s}^{\ast\top} \tilde{\vb{x}})^{2}$ and alignment with $\vb{s}^{\ast}$ is what makes a spike token.
They then separate the activation from the sink.
The activation acts globally, as a near-constant state carried across the layers, where the sink acts locally on the heads, and the two decouple once the pre-norm placement is changed.

\textbf{The trigger and its quadratic approximation.}\quad \citet{sun2026spike} write the SwiGLU block's output on channel $d$, with $\sigma$ taken as the identity, as $\tilde{\vb{x}}^{\top} S_d \tilde{\vb{x}}$, $S_d$ the symmetrized $\sum_i W_{d i}\, G_i U_i^{\top}$ with $\vb{g}$ folded into $G$ and $U$.
For a unit direction $\vb{u}$, $\tilde{\vb{u}} = \sqrt{D}\,\vb{u}$, so the write is $D\,\vb{u}^{\top} S_d\vb{u}$ and its maximizer in sign $s_d$ is the leading eigenvector of $s_d S_d$.
Table~\ref{tab:sun} builds the form at the visual switch of the five Qwen bases and their LVLMs.
Without $\vb{g}$ folded in, the form's eigenvectors are orthogonal to $\vb{t}$, so the fold is not optional.
At the text switch of Qwen3-8B the folded form has one eigenvalue $41$ times the next and the trigger is its eigenvector at cosine $1.000$.
At the visual switch the eigenvalue of largest magnitude is negative on every checkpoint, $1.1$ to $2.6$ times the positive one on the five Qwen bases and $20$ times on Mistral-NeMo-12B, and along that direction the real block writes $-86{,}000$ to $-608{,}000$ on one sign and almost nothing on the other.

\begin{table}[!htbp]
\caption{\textbf{The trigger and the published eigenvector agree at the visual switch, and the trigger survives adaptation.} The quadratic form at the visual switch of the six bases, the norm's weight folded in.
Columns: largest eigenvalues, cosine of the trigger with the top positive eigenvector, real write along each, share of spiking images whose spike token tops that eigenvector, and the trigger's cosine between the base and each of its LVLMs, estimated on the LVLM's own weights at the same block, channel and sign.
Third row: Qwen3-8B's text switch, top-1 read on the first token.
$^\ast$Against the most negative eigenvalue's eigenvector, the sign of Mistral-NeMo-12B's eruption.\protect\backto{tab:sun}}
\label{tab:sun}
\begin{center}\scriptsize
\setlength{\tabcolsep}{4pt}
\resizebox{\textwidth}{!}{\begin{tabular}{llrrrrrr}
\toprule
base & block, channel & $\lambda_{\lvert\max\rvert}$ & $\lambda_{+}$ & $\cos(\vb{t}, v_{+})$ & write along $v_{+}$, $\vb{t}$ & top-1 by $v_{+}$ & $\vb{t}$ across adaptations \\
\midrule
Qwen3-4B & 16, 4 & $-38.2$ & $+14.8$ & 1.000 & 37,893, 37,895 & 0.73, 0.98 & 0.995, 0.995 \\
Qwen3-8B & 16, 2276 & $-39.5$ & $+19.5$ & 1.000 & 79,630, 79,635 & 1.00, 0.85 & 0.997 to 0.998 \\
Qwen3-8B, its text spike token & 6, 2276 & $+17.6$ & $+17.6$ & 1.000 & 70,707, 70,732 & first token at cosine 0.46 & n/a \\
Qwen3-14B & 19, 731 & $-41.3$ & $+36.7$ & 1.000 & 186,333, 186,369 & 0.86 & 0.995 \\
Qwen3-32B & 43, 731 & $-118.8$ & $+71.7$ & 1.000 & 365,735, 365,776 & 0.94 & 0.938, 0.999 \\
Qwen2.5-32B & 38, 4675 & $-32.3$ & $+16.7$ & 1.000 & 85,273, 85,281 & 1.00 & 0.999 \\
Mistral-NeMo-12B & 2, 3448 & $-1.2$ & $+0.1$ & 0.999$^\ast$ & 6,662, 6,674 & 0.997$^\ast$ & 0.990 \\
\bottomrule
\end{tabular}}
\end{center}
\end{table}

\begin{table}[!htbp]
\caption{\textbf{At the switch's input the eventual spike token is aligned with the trigger, and no other token is.} Per LVLM at its visual switch's input, with its base's trigger.
Columns: spiking and non-spiking census images; the eventual spike token's cosine, median and tenth percentile; every other token's, median and 99th percentile; share of spiking images whose spike token is the top token; non-spiking images' best token, median; share of non-spiking images whose best token passes the spike tokens' tenth percentile.
$^\dagger$Five spiking images, indicative.\protect\backto{tab:trigcos}}
\label{tab:trigcos}
\begin{center}\scriptsize
\setlength{\tabcolsep}{2pt}
\begin{tabular}{llrrrrrr}
\toprule
LVLM & base, block & \shortstack[r]{spiking,\\ non-spiking} & spike & others & spike on top & \shortstack[r]{non-spiking\\ best} & \shortstack[r]{non-spiking\\ above} \\
\midrule
Qwen3-VL-4B & Qwen3-4B, 16 & 226, 74 & 0.295, 0.199 & $-0.011$, 0.015 & 0.75 & 0.022 & 0.00 \\
InternVL3.5-4B & Qwen3-4B, 16 & 49, 251 & 0.216, 0.169 & $-0.008$, 0.020 & 1.00 & 0.026 & 0.00 \\
Qwen3-VL-8B & Qwen3-8B, 16 & 38, 262 & 0.210, 0.138 & $-0.012$, 0.003 & 0.89 & 0.008 & 0.00 \\
InternVL3.5-8B & Qwen3-8B, 16 & 185, 115 & 0.220, 0.138 & $-0.015$, 0.018 & 0.82 & 0.034 & 0.00 \\
InternVL3.5-14B & Qwen3-14B, 19 & 207, 93 & 0.119, 0.061 & 0.001, 0.027 & 0.87 & 0.031 & 0.02 \\
Qwen3-VL-32B & Qwen3-32B, 43 & 140, 160 & 0.084, 0.056 & 0.002, 0.024 & 0.91 & 0.029 & 0.09 \\
Qwen2.5-VL-32B & Qwen2.5-32B, 38 & 156, 144 & 0.202, 0.189 & $-0.014$, 0.008 & 1.00 & 0.020 & 0.00 \\
InternVL3.5-38B$^\dagger$ & Qwen3-32B, 43 & 5, 295 & 0.058, 0.053 & 0.001, 0.017 & 1.00 & 0.020 & 0.01 \\
Molmo2-8B & Qwen3-8B, 16 & 300, 0 & 0.256, 0.199 & $-0.018$, 0.108 & 0.44 & -- & -- \\
Pixtral-12B & Mistral-NeMo-12B, 2 & 299, 1 & 0.620, 0.611 & $-0.021$, 0.047 & 1.00 & 0.148 & 0.00 \\
\bottomrule
\end{tabular}
\end{center}
\end{table}

\textbf{One block writes the eruption.}\quad In the nine spiking models with a later visual switch, that block writes $3{,}919$ to $15{,}631$ on the spike channel, whereas the image's largest token stood at $3$ to $165$ after the text switch, and Pixtral-12B's text switch writes $2{,}518$ (Table~\ref{tab:twoswitch}).
In the others the channel creeps up over the last third of the depth (Figure~\ref{fig:switchall}).
The writing block is 16 in the Qwen3-4B and Qwen3-8B models, 19 in Qwen3-14B, 43 in Qwen3-32B and 38 in Qwen2.5-VL-32B, and its down-projection row for the spike channel is unchanged by the adaptations, at cosine $0.997$ to the text-only base.
The base's feed-forward at that block, applied from its weights to the token as the feed-forward sees it, writes $3{,}000$ to $22{,}000$ on the spike channel for the eventual spike token and $1$ to $13$ for any other token.
The base's text does not spike through this block, since block 16 of Qwen3-8B writes at most $6.9$ on the channel for any token but the first over 100 prompts.
Molmo2-8B's eruption is written at the same block 16 of Qwen3-8B, on several tokens per image, and Pixtral-12B's at block 2 of Mistral-NeMo-12B, its text switch (\aref{app:extension}).

\textbf{The two switches on every model.}\quad Table~\ref{tab:twoswitch} reads, for each of the 25, what its text switch does to the image tokens and whether a later block writes one.
The text switch is the block where the text spike token takes its largest step.
GLM-4.1V-9B-Base and PaliGemma 2-3B build their text spike over many blocks with no step above $330$, as their bases do, so the block printed for them is their base's largest step and not an abrupt switch.
The write it makes on an image token, and the image's peak after it, are medians over the 300 census images.
The later block is the one making the largest write on an image token.

\begin{table}[!htbp]
\caption{\textbf{The text switch leaves image tokens small, and only the spiking models write one into the thousands, Pixtral-12B at the text switch itself.} Per LVLM: the text switch's block, the text spike token before and after it, that block's largest write on an image token, and the image's peak after it as $\lvert \vb{x}[d] \rvert$ and $\Psi$.
Then the largest later write's block, the share of spiking images on which it is that block, and that write.
Read on block outputs, over image tokens entering below the floor.\protect\backto{tab:twoswitch}}
\label{tab:twoswitch}
\begin{center}\scriptsize
\setlength{\tabcolsep}{3pt}
\begin{tabular}{lllrrl}
\toprule
LVLM & base LLM & text switch: text spike token & image write & image peak & later block, spiking images, write \\
\midrule
Qwen3-VL-4B & Qwen3-4B & 6: 12 $\to$ 11{,}072 & 18 & 8, 15 & 16, 100\%, 6{,}736 \\
InternVL3.5-4B & Qwen3-4B & 6: 16 $\to$ 15{,}616 & 17 & 3, 11 & 16, 100\%, 5{,}786 \\
Qwen3-VL-8B & Qwen3-8B & 6: 20 $\to$ 13{,}440 & 26 & 5, 12 & 16, 100\%, 6{,}092 \\
InternVL3.5-8B & Qwen3-8B & 6: 26 $\to$ 24{,}448 & 22 & 5, 16 & 16, 100\%, 12{,}275 \\
InternVL3.5-14B & Qwen3-14B & 6: 278 $\to$ 13{,}888 & 25 & 68, 179 & 19, 100\%, 10{,}494 \\
Qwen2.5-VL-32B & Qwen2.5-32B & 6: 36 $\to$ 3{,}280 & 95 & 165, 142 & 38, 100\%, 3{,}919 \\
Qwen3-VL-32B & Qwen3-32B & 6: 159 $\to$ 15{,}808 & 38 & 136, 127 & 43, 100\%, 15{,}631 \\
InternVL3.5-38B & Qwen3-32B & 6: 270 $\to$ 24{,}832 & 23 & 107, 182 & 43, 100\%, 8{,}423 \\
Molmo2-8B & Qwen3-8B & 6: 42 $\to$ 19{,}328 & 31 & 3, 5 & 16, 100\%, 13{,}544 \\
Pixtral-12B & Mistral-NeMo-12B & 2: 21 $\to$ 588 & 2{,}518 & 2{,}496, 32{,}271 & 20, 70\%, 24 \\
\midrule
Qwen2-VL-7B & Qwen2-7B & 3: 40 $\to$ 4{,}640 & 19 & 56, 123 & 25, 72 \\
Molmo-7B-D & Qwen2-7B & 3: 116 $\to$ 680 & 30 & 95, 296 & 23, 121 \\
LLaVA-OV-7B & Qwen2-7B & 3: 77 $\to$ 6{,}112 & 288 & 176, 265 & last, 101 \\
Qwen2.5-VL-3B & Qwen2.5-3B & 2: 41 $\to$ 2{,}544 & 17 & 35, 37 & last, 66 \\
Qwen2.5-VL-7B & Qwen2.5-7B & 3: 34 $\to$ 5{,}600 & 30 & 41, 80 & last, 136 \\
Qwen2.5-Omni-7B & Qwen2.5-7B & 3: 21 $\to$ 6{,}400 & 25 & 25, 53 & last, 102 \\
Qwen2.5-VL-72B & Qwen2.5-72B & 1: 1 $\to$ 6{,}720 & 3 & 18, 66 & last, 608 \\
Qwen3-VL-2B & Qwen3-1.7B & 2: 26 $\to$ 14{,}720 & 57 & 81, 93 & last, 392 \\
LLaVA-1.5-7B & Vicuna-7B & 1: 1 $\to$ 1{,}320 & 200 & 168, 589 & 24, 3 \\
LLaVA-1.5-13B & Vicuna-13B & 3: 45 $\to$ 1{,}400 & 52 & 65, 283 & last, 49 \\
InternVL2-8B & InternLM2.5-7B & 1: 1 $\to$ 1{,}688 & 3 & 7, 33 & last, 52 \\
GLM-4.1V-9B-Base & GLM-4-9B & 27: 1{,}624 $\to$ 1{,}848 & 175 & 270, 168 & 36, 154 \\
Janus-Pro-7B & DeepSeek-LLM-7B & 1: 1 $\to$ 12{,}288 & 30 & 78, 42 & 8, 27 \\
PaliGemma 2-3B & Gemma-2-2B & 9: 1{,}224 $\to$ 1{,}428 & 86 & 217, 154 & 18, 172 \\
Idefics3-8B & Llama-3.1-8B & 1: 0 $\to$ 249 & 1 & 1, 18 & last, 0 \\
\bottomrule
\end{tabular}
\end{center}
\end{table}

\FloatBarrier
\textbf{Two gates, what a block can write and what reaches it.}\quad On the Qwen3-4B base the two triggers of Equation~\ref{eq:trigger} are orthogonal, at cosine $0.057$, and each block writes a third of a percent of its own maximum along the other's trigger.
Table~\ref{tab:pos0} feeds one-token sequences to the language model at position 0.
There the residual entering the text switch is the embedding plus the blocks' own accumulated writes, of norm $20$ to $34$ against a text embedding's $1.1$. A small vector of any direction therefore enters the block at cosine $0.88$ with its trigger and is written on, and a vector as large as those writes keeps its own direction and is not.
The exceptions are 21 control and rare tokens, among them \texttt{<|im\_start|>}, so the chat prompts carry their text spike token on the second token.
Table~\ref{tab:pos0all} repeats the reading on the ten spiking LVLMs.
Gaussian vectors at the text norm reach the level on $99.9$ to $100\%$ in eight of them, and the projector's image tokens at their own norm reach it on at most $8.3\%$ in all ten.
Pixtral-12B, whose text norm is $0.5$, carries $99.2\%$ of its vocabulary but $9.5\%$ of the Gaussian vectors at that norm.
InternVL3.5-14B's text switch writes its chat-template start token and $22.3\%$ of its vocabulary but none of the Gaussian vectors, which its base Qwen3-14B carries after the same block.
Those vectors reach the level later, $95.7\%$ of them after block 19, its visual switch.
In Qwen3-VL-8B, Qwen3-VL-32B, InternVL3.5-38B and Qwen2.5-VL-32B, Gaussian vectors at the image norm are also carried, on $88.7$ to $98.2\%$, while the image tokens at that norm are not.
Table~\ref{tab:capacity} scans every block of the base for its largest write on the channel along any unit direction in the eruption's sign.
Table~\ref{tab:reach} gives, for the blocks whose write passes $5{,}000$, how close the spike token and the closest other image token come to each block's trigger.
Only block 16 is approached, at $+0.28$ to $+0.43$ for the spike token against $+0.02$ to $+0.09$ for the closest other token, and with the image positions filled by Gaussian vectors, by zeros, or by the real tokens scaled by $3$, no image token spikes.

\begin{table}[!htbp]
\caption{\textbf{At position 0 any small enough vector is carried onto the trigger.} Qwen3-VL-4B's language model, the base in the second row: one-token sequences of the vectors named, the share reaching the level and the spike channel after block 6.\protect\backto{tab:pos0}}
\label{tab:pos0}
\begin{center}\small
\begin{tabular}{lrrrr}
\toprule
vector at position 0 & norm & $n$ & reach the level (\%) & channel after block 6 \\
\midrule
every vocabulary token (LVLM) & 1.1 & 151,936 & 100.0 & 7,712 \\
every vocabulary token (base) & 1.1 & 151,936 & 100.0 & 10,176 \\
image tokens, the projector's outputs & 39.6 & 2,048 & 0.5 & 4 \\
the same image tokens at the text norm & 1.1 & 2,048 & 100.0 & 7,936 \\
the same at ten times the text norm & 11.1 & 2,048 & 16.4 & 18 \\
text embeddings at the image norm & 39.6 & 4,096 & 3.0 & 104 \\
Gaussian at the text norm & 1.1 & 4,096 & 100.0 & 7,904 \\
Gaussian at ten times it & 11.1 & 512 & 99.0 & 7,200 \\
Gaussian at twenty times it & 22.1 & 2,048 & 14.0 & 150 \\
Gaussian at forty times it & 44.2 & 2,048 & 0.1 & 30 \\
Gaussian at the image norm & 39.6 & 4,096 & 0.6 & 34 \\
\bottomrule
\end{tabular}
\end{center}
\end{table}

\begin{table}[!htbp]
\caption{\textbf{At position 0 small vectors reach the text spike in eight of the ten models, and image tokens in none.} Each spiking LVLM's language model on one-token sequences, as in Table~\ref{tab:pos0}: the share (\%) reaching the level after the text switch.
Columns: every vocabulary token in the LVLM and in its base, the projector's image tokens at their own norm and at the text norm, Gaussian vectors at the text norm, at twenty times it and at the image norm, then the text and the image norm.\protect\backto{tab:pos0all}}
\label{tab:pos0all}
\begin{center}\small
\setlength{\tabcolsep}{3pt}
\begin{tabular}{lrrrrrrrr}
\toprule
 & \multicolumn{2}{c}{vocabulary} & \multicolumn{2}{c}{image tokens} & \multicolumn{3}{c}{Gaussian} & \\
\cmidrule(lr){2-3}\cmidrule(lr){4-5}\cmidrule(lr){6-8}
model & LVLM & base & own norm & text norm & text norm & $20\times$ & image norm & norms \\
\midrule
Qwen3-VL-4B & 100.0 & 100.0 & 0.5 & 100.0 & 100.0 & 14.0 & 0.5 & 1.1, 39.4 \\
InternVL3.5-4B & 99.8 & 100.0 & 0.0 & 77.8 & 99.9 & 5.0 & 9.5 & 1.1, 19.9 \\
Qwen3-VL-8B & 99.9 & 99.9 & 4.4 & 96.1 & 100.0 & 30.7 & 88.7 & 1.4, 19.0 \\
InternVL3.5-8B & 95.3 & 99.9 & 0.0 & 62.5 & 100.0 & 0.1 & 0.0 & 1.4, 23.8 \\
Molmo2-8B & 99.7 & 99.9 & 0.0 & 94.2 & 100.0 & 3.4 & 0.0 & 1.5, 85.2 \\
InternVL3.5-14B & 22.3 & 99.9 & 0.0 & 2.7 & 0.0 & 0.1 & 0.0 & 1.3, 38.5 \\
Qwen3-VL-32B & 100.0 & 100.0 & 8.3 & 100.0 & 100.0 & 100.0 & 96.0 & 1.1, 88.5 \\
InternVL3.5-38B & 100.0 & 100.0 & 0.3 & 100.0 & 100.0 & 97.6 & 92.2 & 1.3, 57.6 \\
Qwen2.5-VL-32B & 98.4 & 98.4 & 0.3 & 98.6 & 100.0 & 100.0 & 98.2 & 1.3, 158.2 \\
Pixtral-12B & 99.2 & 99.2 & 0.1 & 0.8 & 9.5 & 0.7 & 0.0 & 0.5, 16.7 \\
\bottomrule
\end{tabular}
\end{center}
\end{table}

\begin{table}[!htbp]
\caption{\textbf{Many blocks could write the channel.
One writes an image token.} The Qwen3-4B base, channel 4, the ten blocks whose write along the trigger exceeds $5{,}000$: per block the largest write along any unit direction in the eruption's sign (Equation~\ref{eq:trigger}) and the write the block actually makes on an image token, on an ordinary text token and on the text spike token.\protect\backto{tab:capacity}}
\label{tab:capacity}
\begin{center}\small
\begin{tabular}{lrrrr}
\toprule
block & along the trigger & on an image token & on ordinary text & on the text spike token \\
\midrule
1 & 8,850 & 3 & 1 & 0 \\
2 & 7,635 & 3 & 0 & 4 \\
6 & 28,879 & 5 & 5 & 11,072 \\
16 & 37,895 & 7,328 & 5 & 211 \\
30 & 7,128 & 26 & 31 & -2 \\
31 & 7,333 & 26 & 25 & -3 \\
32 & 6,336 & 23 & 30 & -4 \\
33 & 8,135 & 39 & 40 & -17 \\
34 & 79,900 & 98 & 135 & -2,336 \\
35 & 263,423 & 0 & 412 & -5,952 \\
\bottomrule
\end{tabular}
\end{center}
\end{table}

\begin{table}[!htbp]
\caption{\textbf{Only the visual switch's trigger is reached.} For the blocks of the two bases whose write along the trigger passes $5{,}000$: that write, and the cosine with the block's trigger of the eventual spike token and of the closest other image token, in each adaptation of the base.
Read on the residual entering the block's feed-forward, median over the spiking images, 49 and 5 of 60 on Qwen3-4B and 6 and 40 of 60 on Qwen3-8B.
Read under $\Psi$ with the floor, tokens entering at $100$ or above excluded.\protect\backto{tab:reach}}
\label{tab:reach}
\begin{center}
\scriptsize
\begin{tabular}{llrrrrr}
\toprule
base & block & along the trigger & spike, Qwen3-VL & closest, Qwen3-VL & spike, InternVL3.5 & closest, InternVL3.5 \\
\midrule
Qwen3-4B & 1 & 8{,}850 & $-0.30$ & $-0.13$ & $-0.40$ & $-0.27$ \\
Qwen3-4B & 2 & 7{,}635 & $-0.33$ & $-0.17$ & $-0.42$ & $-0.35$ \\
Qwen3-4B & 6 & 28{,}880 & $-0.02$ & $+0.01$ & $-0.02$ & $-0.01$ \\
Qwen3-4B & 16 & 37{,}895 & $+0.40$ & $+0.09$ & $+0.42$ & $+0.04$ \\
Qwen3-4B & 30 & 7{,}128 & $0.00$ & $-0.09$ & $0.00$ & $-0.09$ \\
Qwen3-4B & 31 & 7{,}333 & $-0.01$ & $-0.09$ & $-0.01$ & $-0.08$ \\
Qwen3-4B & 32 & 6{,}336 & $-0.01$ & $-0.09$ & $-0.01$ & $-0.07$ \\
Qwen3-4B & 33 & 8{,}135 & $-0.01$ & $-0.10$ & $0.00$ & $-0.07$ \\
Qwen3-4B & 34 & 79{,}900 & $-0.02$ & $-0.11$ & $-0.02$ & $-0.08$ \\
Qwen3-4B & 35 & 263{,}423 & $-0.05$ & $-0.03$ & $-0.03$ & $-0.04$ \\
Qwen3-8B & 1 & 11{,}726 & $-0.27$ & $-0.17$ & $-0.34$ & $-0.23$ \\
Qwen3-8B & 2 & 1{,}852{,}864 & $-0.27$ & $-0.15$ & $-0.29$ & $-0.22$ \\
Qwen3-8B & 6 & 70{,}736 & $-0.01$ & $-0.01$ & $-0.02$ & $-0.01$ \\
Qwen3-8B & 16 & 79{,}635 & $+0.28$ & $+0.02$ & $+0.43$ & $+0.05$ \\
Qwen3-8B & 26 & 5{,}984 & $+0.01$ & $-0.08$ & $+0.02$ & $-0.07$ \\
Qwen3-8B & 27 & 7{,}424 & $+0.01$ & $-0.09$ & $+0.02$ & $-0.06$ \\
Qwen3-8B & 28 & 9{,}969 & $+0.02$ & $-0.10$ & $+0.02$ & $-0.06$ \\
Qwen3-8B & 29 & 14{,}990 & $+0.01$ & $-0.10$ & $+0.01$ & $-0.07$ \\
Qwen3-8B & 30 & 19{,}885 & $+0.01$ & $-0.09$ & $+0.01$ & $-0.06$ \\
Qwen3-8B & 31 & 21{,}409 & $0.00$ & $-0.09$ & $+0.01$ & $-0.05$ \\
Qwen3-8B & 32 & 20{,}724 & $+0.01$ & $-0.09$ & $+0.01$ & $-0.04$ \\
Qwen3-8B & 33 & 30{,}983 & $0.00$ & $-0.11$ & $+0.01$ & $-0.05$ \\
Qwen3-8B & 34 & 271{,}639 & $-0.01$ & $-0.12$ & $0.00$ & $-0.07$ \\
Qwen3-8B & 35 & 845{,}277 & $-0.03$ & $-0.03$ & $-0.01$ & $-0.03$ \\
\bottomrule
\end{tabular}
\end{center}
\end{table}

\textbf{The text-only base fed the adaptation's vectors.}\quad Table~\ref{tab:basefed} runs the text-only Qwen3-4B on the vectors Qwen3-VL-4B hands its language model, the input of decoder block 0 with the base's positions and without the DeepStack features. It also swaps the two models at a block, the base from block $k$ on given the adapted model's residual entering $k$.
The adapted language model spikes on that input, on another token.
The base does not, and given the adapted model's state it spikes from block 16 on three images of four, on the adaptation's own token.
Every block of the adapted language model sits at a relative Frobenius distance of $0.14$ to $0.28$ from the base's.
Table~\ref{tab:basefedov} does the same for LLaVA-OV-7B on Qwen2-7B, where the base's text switch writes the least-shared image token on $85\%$ of the images and the adapted blocks take the token out of its reach.
Table~\ref{tab:basefedall} runs the same swap on every pair. Given the state entering the visual switch, the base spikes on the LVLM's spike token on $66$ to $100\%$ of spiking images in seven pairs, on $42\%$ for Qwen3-VL-32B, and on none of InternVL3.5-38B's three.
Pixtral-12B's base spikes already on the decoder-input vectors, since its visual switch is its text switch.
Among the non-spiking LVLMs, only LLaVA-OV-7B's base spikes on the adaptation's vectors, and the other six stay at $0$ to $1\%$.

\begin{table}[!htbp]
\caption{\textbf{The base does not spike on the adaptation's vectors until its own visual switch.} Qwen3-VL-4B's decoder-0 input run through the text-only Qwen3-4B, 150 POPE images: the peak $\Psi$ over the deep layers, the share of images at or above the level, the share on which the peak is the adapted model's own spike token, and $\Psi$ at that token.
In parentheses, the same run on Qwen3-4B-Base, the pretrained checkpoint the instruction-tuned base is built from.\protect\backto{tab:basefed}}
\label{tab:basefed}
\centering\footnotesize\setlength{\tabcolsep}{3pt}
\begin{tabular}{lrrrr}
\toprule
input & peak $\Psi$ & $\ge 1{,}000$ & on the spike token & $\Psi$ there \\
\midrule
Qwen3-VL-4B as it runs & 16{,}384 & 78\% & 100\% & 16{,}384 \\
its LM, 1-D positions, no DeepStack & 13{,}243 & 73\% & 7\% & 56 \\
\quad 2-D positions, no DeepStack & 13{,}092 & 57\% & 11\% & 52 \\
\quad 1-D positions, DeepStack & 15{,}763 & 87\% & 29\% & 60 \\
\quad 1-D, from block 13, its own state & 15{,}767 & 85\% & 73\% & 14{,}415 \\
\quad from block 13, the base's state & 204 (250) & 1\% (9\%) & 1\% (1\%) & 122 (132) \\
Qwen3-4B on the decoder-0 input & 196 (241) & 0\% (3\%) & 1\% (1\%) & 110 (104) \\
\quad from block 13, the adapted state & 129 (111) & 7\% (1\%) & 5\% (3\%) & 89 (87) \\
\quad from block 14 & 196 (227) & 15\% (6\%) & 9\% (3\%) & 89 (86) \\
\quad from block 15 & 7{,}048 (311) & 52\% (38\%) & 41\% (31\%) & 96 (92) \\
\quad from block 16 & 15{,}331 (12{,}255) & 77\% (75\%) & 76\% (73\%) & 15{,}250 (11{,}937) \\
\bottomrule
\end{tabular}
\end{table}

\begin{table}[!htbp]
\caption{\textbf{Here the base spikes on the adaptation's vectors and the adaptation does not.} LLaVA-OV-7B's decoder-0 input run through the text-only Qwen2-7B, both models with one-dimensional positions.
Top: $\Psi$ after each block at the base's peak image token and at position 0, median over the 128 of 150 images on which the base spikes.
Bottom: on 150 images, the peak $\Psi$ over the deep layers and where it lands.\protect\backto{tab:basefedov}}
\label{tab:basefedov}
\centering\footnotesize
\begin{tabular}{lrrrr}
\toprule
after block & base, its token & adapted, that token & base, position 0 & adapted, position 0 \\
\midrule
2 & 6 & 11 & 10 & 107 \\
3 & 1{,}748 & 9 & 7{,}605 & 8{,}992 \\
4 & 2{,}655 & 16 & 8{,}777 & 10{,}561 \\
22 & 3{,}137 & 54 & 9{,}058 & 7{,}956 \\
\bottomrule
\end{tabular}
\medskip

\begin{tabular}{lrrrr}
\toprule
input & peak $\Psi$ & $\ge 1{,}000$ & on the base's token & $\Psi$ there \\
\midrule
LLaVA-OV-7B as it runs & 329 & 0\% & 3\% & 329 \\
Qwen2-7B on the decoder-0 input & 5{,}040 & 85\% & 100\% & 5{,}040 \\
\quad from block 1, the adapted state & 2{,}919 & 73\% & 58\% & 1{,}529 \\
\quad from block 2 & 248 & 0\% & 12\% & 95 \\
\quad from block 3 & 237 & 0\% & 12\% & 94 \\
LLaVA-OV-7B from block 3, the base's state & 1{,}018 & 50\% & 49\% & 425 \\
\quad from block 4 & 4{,}620 & 85\% & 90\% & 4{,}620 \\
\quad from block 6 & 4{,}082 & 85\% & 93\% & 4{,}082 \\
\bottomrule
\end{tabular}
\end{table}

\begin{table}[!htbp]
\caption{\textbf{Across the pairs, the base spikes on the LVLM's token only once given the state entering the visual switch.} Each LVLM with its text-only base, 150 POPE images. Columns: the LVLM's spiking images; the share of all images on which the base, run on the LVLM's decoder-input vectors, spikes, and the share of spiking images on which it does so on the LVLM's own spike token; then, for each block $k$ from which the base is given the LVLM's state (in parentheses, the last is the visual switch), the share of spiking images on which the base spikes on the LVLM's spike token. Lower block, the non-spiking LVLMs: the share of images on which the base spikes on the decoder-input vectors, and its median peak $\Psi$ there. Read under $\Psi$ with the floor, tokens entering at $100$ or above excluded.\protect\backto{tab:basefedall}}
\label{tab:basefedall}
\begin{center}
\footnotesize
\setlength{\tabcolsep}{3pt}
\resizebox{\linewidth}{!}{\begin{tabular}{llrrrrrr}
\toprule
 & & & \multicolumn{2}{c}{decoder input} & \multicolumn{3}{c}{from block $k$} \\
LVLM & base & spiking & base spikes & on its token & $S-3$ & $S-1$ & $S$ \\
\midrule
Qwen3-VL-4B & Qwen3-4B & 114/150 & 0\% & 0\% & 4\% (13) & 48\% (15) & 90\% (16) \\
InternVL3.5-4B & Qwen3-4B & 25/150 & 0\% & 0\% & 8\% (13) & 76\% (15) & 100\% (16) \\
Qwen3-VL-8B & Qwen3-8B & 13/150 & 0\% & 0\% & 0\% (13) & 31\% (15) & 100\% (16) \\
InternVL3.5-8B & Qwen3-8B & 89/150 & 0\% & 0\% & 7\% (13) & 53\% (15) & 93\% (16) \\
Molmo2-8B & Qwen3-8B & 150/150 & 3\% & 0\% & 45\% (13) & 54\% (15) & 66\% (16) \\
InternVL3.5-14B & Qwen3-14B & 94/150 & 0\% & 0\% & 0\% (16) & 30\% (18) & 87\% (19) \\
Qwen3-VL-32B & Qwen3-32B & 52/150 & 0\% & 0\% & 0\% (40) & 6\% (42) & 42\% (43) \\
InternVL3.5-38B & Qwen3-32B & 3/150 & 0\% & 0\% & 0\% (40) & 0\% (42) & 0\% (43) \\
Qwen2.5-VL-32B & Qwen2.5-32B & 78/150 & 0\% & 0\% & 97\% (35) & 99\% (37) & 99\% (38) \\
Pixtral-12B & Mistral-NeMo-12B & 150/150 & 100\% & 100\% & 100\% (1) & 100\% (2) & -- \\
\midrule
Qwen2-VL-7B & Qwen2-7B & 0/150 & 0\% & 151 & -- & -- & -- \\
LLaVA-OV-7B & Qwen2-7B & 0/150 & 85\% & 5{,}040 & -- & -- & -- \\
Molmo-7B-D & Qwen2-7B & 0/150 & 0\% & 448 & -- & -- & -- \\
Qwen2.5-VL-3B & Qwen2.5-3B & 0/150 & 1\% & 205 & -- & -- & -- \\
Qwen2.5-VL-7B & Qwen2.5-7B & 0/150 & 0\% & 134 & -- & -- & -- \\
Qwen2.5-Omni-7B & Qwen2.5-7B & 0/150 & 0\% & 140 & -- & -- & -- \\
Qwen3-VL-2B & Qwen3-1.7B & 0/150 & 0\% & 101 & -- & -- & -- \\
\bottomrule
\end{tabular}}
\end{center}
\end{table}

\textbf{Every row of every block.}\quad Table~\ref{tab:rowcap} gives the largest write of every row of every block's down projection in the 20 bases of Table~\ref{tab:heldout}, from the weights alone. Each row's largest write is the maximum of $\lvert \vb{w}_d^{\top} \vb{h}(\vb{u}) \rvert$ over unit $\vb{u}$, by Adam for $200$ steps from one random start per row and four for the leading rows.
Over the 722 blocks of the 18 bases whose feed-forward output is not normalized, the two held-out bases among them, the spike channel tops 494, and 237 can write it above $5{,}000$, its candidate blocks.
GLM-4-9B and Gemma-2-2B normalize that output, so their writes are read before the norm and left out of these counts.
Table~\ref{tab:passed} counts, for the six bases carrying a spiking LVLM, the candidate blocks between the text switch and the visual switch.
Table~\ref{tab:textmove} runs Qwen2.5-32B and Qwen2.5-VL-32B text only on the same token ids, where the LVLM's blocks 4 to 6 keep the base's largest writes to within $15\%$.

\begin{table}[!htbp]
\caption{\textbf{At the switches the spike channel's writer writes hardest of the block's rows.} Per base, at the text switch, the visual switch and the last third's top block: the writer's rank among the block's rows, four starts per row, and its ratio to the best other row.
Parentheses: a stand-in where no LVLM on the base spikes.
Last column: blocks topped by a spike row, one start per row.
$^{*}$ largest single weight on a spike row.
$^\dagger$Before the norm these two apply to the feed-forward's output.\protect\backto{tab:rowcap}}
\label{tab:rowcap}
\centering\footnotesize\setlength{\tabcolsep}{4pt}
\begin{tabular}{llllr}
\toprule
base & text switch & visual switch & last third & spike channel on top \\
\midrule
Qwen3-4B & 6: 1st, $\times 3.5^{*}$ & 16: 1st, $\times 8.6^{*}$ & 35: 1st$^{*}$ & 13 of 36 \\
Qwen3-8B & 6: 1st, $\times 5.0^{*}$ & 16: 1st, $\times 22.9^{*}$ & 34: 1st$^{*}$ & 13 of 36 \\
Qwen3-14B & 6: 1st, $\times 7.3^{*}$ & 19: 1st, $\times 6.3^{*}$ & 38: 1st & 39 of 40 \\
Qwen3-32B & 6: 1st, $\times 10.1^{*}$ & 43: 1st, $\times 6.5^{*}$ & 62: 1st$^{*}$ & 63 of 64 \\
Qwen2.5-32B & 5: 1st, $\times 1.4^{*}$ & 38: 1st, $\times 1.7^{*}$ & 60: 1st$^{*}$ & 55 of 64 \\
\midrule
Qwen3-1.7B & 2: 1st, $\times 15.0^{*}$ & (18: 1st, $\times 1.3$) & 27: 1st$^{*}$ & 24 of 28 \\
Qwen2.5-3B & 2: 6th, $\times 0.44$ & (17: 1st, $\times 2.0^{*}$) & 30: 1st$^{*}$ & 24 of 36 \\
Qwen2.5-7B & 3: 1st, $\times 3.3^{*}$ & (11: 1st, $\times 1.5^{*}$) & 26: 1st$^{*}$ & 22 of 28 \\
Qwen2.5-72B & 1: 1st, $\times 6.0^{*}$ & (45: 1st, $\times 1.3^{*}$) & 79: 1st$^{*}$ & 71 of 80 \\
Qwen2-7B & 3: 1st, $\times 2.4^{*}$ & (11: 1st, $\times 1.2^{*}$) & 26: 1st$^{*}$ & 22 of 28 \\
InternLM2.5-7B & 1: 1st, $\times 17.9^{*}$ & (19: 1st, $\times 1.6^{*}$) & 30: 1st$^{*}$ & 30 of 32 \\
Vicuna-7B & 1: 1st, $\times 17.8^{*}$ & (18: 3rd, $\times 0.50$) & 30: 1st$^{*}$ & 5 of 32 \\
Vicuna-13B & 3: 1st, $\times 13.9^{*}$ & (14: 3rd, $\times 0.86$) & 38: 1st$^{*}$ & 8 of 40 \\
Qwen3-0.6B & 2: 1st, $\times 10.1^{*}$ & (18: 1st, $\times 2.1^{*}$) & 27: 1st$^{*}$ & 25 of 28 \\
Llama-3.1-8B & 1: 1st, $\times 1.1$ & (17: 5th, $\times 0.36$) & 31: 1st & 2 of 32 \\
Qwen2.5-14B & 5: 3rd, $\times 0.49^{*}$ & (22: 1st, $\times 1.2^{*}$) & 46: 1st$^{*}$ & 37 of 48 \\
Mistral-NeMo-12B & 2: 1st, $\times 5.7^{*}$ & 2: 1st, $\times 5.7^{*}$ & 38: 1st$^{*}$ & 37 of 40 \\
DeepSeek-LLM-7B & 1: 1st, $\times 8.2^{*}$ & (18: 4047th, $\times 0.23$) & 27: 1st$^{*}$ & 4 of 30 \\
GLM-4-9B$^\dagger$ & 27: 5th, $\times 0.12$ & (25: 83rd, $\times 0.16$) & 39: 3rd & 0 of 40 \\
Gemma-2-2B$^\dagger$ & 9: 3rd, $\times 0.38$ & (9: 3rd, $\times 0.38$) & 24: 3rd & 0 of 26 \\
\bottomrule
\end{tabular}
\end{table}

\begin{table}[!htbp]
\caption{\textbf{Image tokens can pass candidate blocks before their visual switch.} The six bases carrying a spiking LVLM: the text switch, the visual switch with its largest write on the spike channel, and the candidate blocks between the two, with how many of them have the spike channel as their top row. \protect\backto{tab:passed}}
\label{tab:passed}
\centering\footnotesize\setlength{\tabcolsep}{4pt}
\begin{tabular}{lrrrr}
\toprule
base & text switch & \shortstack[r]{visual switch\\ (largest write)} & \shortstack[r]{candidate blocks\\ between} & \shortstack[r]{spike channel\\ on top} \\
\midrule
Qwen3-4B & 6 & 16 ($97{,}859$) & 0 & 0 \\
Qwen3-8B & 6 & 16 ($161{,}622$) & 0 & 0 \\
Qwen3-14B & 6 & 19 ($211{,}300$) & 1 & 1 \\
Qwen3-32B & 6 & 43 ($608{,}414$) & 15 & 15 \\
Qwen2.5-32B & 5 & 38 ($165{,}490$) & 6 & 4 \\
Mistral-NeMo-12B & 2 & 2 ($6{,}672$) & 0 & 0 \\
\bottomrule
\end{tabular}
\end{table}

\begin{table}[!htbp]
\caption{\textbf{On the same tokens the adapted model carries its text spike token elsewhere, on the same channels.} Qwen2.5-32B and Qwen2.5-VL-32B text only, on the same token ids of the 12 chat prompts under each model's template, channels 4675 and 3094: the text spike token's position, counted from 0, its text switch with the median value before and after, its peak, and the peaks at positions 1 and 2.}
\label{tab:textmove}
\centering\footnotesize
\setlength{\tabcolsep}{4pt}
\begin{tabular}{llrlrrr}
\toprule
model & template & text spike token & text switch & peak & position 1 & position 2 \\
\midrule
Qwen2.5-32B & its own & 1 & 5 ($12 \rightarrow 14{,}976$) & $18{,}176$ & $18{,}176$ & 796 \\
Qwen2.5-32B & the LVLM's & 1 & 5 ($12 \rightarrow 14{,}976$) & $18{,}176$ & $18{,}176$ & 796 \\
Qwen2.5-VL-32B & its own & 2 & 6 ($36 \rightarrow 3{,}280$) & $5{,}728$ & 452 & $5{,}728$ \\
Qwen2.5-VL-32B & the base's & 2 & 6 ($36 \rightarrow 3{,}280$) & $5{,}728$ & 452 & $5{,}728$ \\
\bottomrule
\end{tabular}
\end{table}

\textbf{The base's text switch bounds spiking and does not decide it.}\quad Each base runs its own text forward on 12 chat prompts under its own template, and from it we read its text switch and the largest write on the spike channel that a block in the middle third of its depth can make.
On the 13 bases the two rules were fixed on, the five carrying a spiking LVLM switch at block 5 or 6 and the eight whose LVLMs do not spike by block 3, and the largest write separates them too, at $97{,}858$ and above against $44{,}078$ and below (Table~\ref{tab:heldout}).
Before any held-out model was run, we fixed two rules on the 13, spiking when the switch sits at block 4 or later and spiking when that write reaches $65{,}677$.
Qwen3-0.6B switches early, and InternVL3.5-1B spikes on none of 300 images, as both rules predict.
Qwen2.5-14B switches at block 5 with a write of $136{,}076$, and InternVL3-14B spikes on none of its 300 images, so both rules fail on their spiking side.
The base supplies the block that can write, and the adaptation aligns image tokens with it or does not.
On the five further bases the first rule predicts spiking for Gemma-2-2B and GLM-4-9B, whose adaptations do not spike. Both rules predict none for Mistral-NeMo-12B, whose text spike takes a first small step at block 0 and is written at block 2, its text switch, and whose Pixtral-12B spikes at the text switch itself (\aref{app:extension}).
With the base run text only on the same tokens, its first token steps from $13$ to $744$ at block 2 and Pixtral-12B's from $21$ to $588$, and the base also writes a sentence delimiter of its prompt at block 0, to $888$.
The remaining two bases, DeepSeek-LLM-7B and Llama-3.1-8B, follow both rules.

\begin{table}[!htbp]
\centering
\caption{\textbf{An early text switch goes with no visual switch in every base but Mistral-NeMo-12B, and a late one does not decide.} Per base, from its text forward on 12 chat prompts: blocks, text switch, largest spike-channel write of a middle-third block, that block in parentheses.
Per LVLM: text switch, clean $\mathrm{VSI}_\kappa$, and share of census images, 60 or 300 per model, whose best read block holds an image token at cosine $0.2$ or more with the base's trigger.
Read blocks: text switch, each third's top block, visual switch.}
\label{tab:heldout}
\small
\setlength{\tabcolsep}{3pt}
\resizebox{\linewidth}{!}{\begin{tabular}{lrrrlrrr}
\toprule
base & blocks & text switch & largest write (block) & LVLM & its text switch & $\mathrm{VSI}_\kappa$ (\%) & past $0.2$ (\%, block) \\
\midrule
\multicolumn{8}{l}{\emph{Bases carrying a spiking LVLM}} \\
Qwen3-4B & 36 & 6 & $97{,}858$ (16) & Qwen3-VL-4B & 6 & $75.3$ & $80$ (16) \\
 & & & & InternVL3.5-4B & 6 & $16.3$ & $17$ (16) \\
Qwen3-8B & 36 & 6 & $161{,}627$ (16) & Qwen3-VL-8B & 6 & $12.7$ & $15$ (16) \\
 & & & & InternVL3.5-8B & 6 & $61.7$ & $67$ (16) \\
Qwen3-14B & 40 & 6 & $211{,}303$ (19) & InternVL3.5-14B & 6 & $69.0$ & $65$ (19) \\
Qwen3-32B & 64 & 6 & $132{,}278$ (31) & Qwen3-VL-32B & 6 & $46.7$ & $52$ (43) \\
 & & & & InternVL3.5-38B & 6 & $1.7$ & $0$ \\
Qwen2.5-32B & 64 & 5 & $168{,}995$ (38) & Qwen2.5-VL-32B & 6 & $52.0$ & $17$ (38) \\
\addlinespace[2pt]
Qwen3-8B & 36 & 6 & $161{,}627$ (16) & Molmo2-8B & 6 & $100.0$ & $100$ (16) \\
Mistral-NeMo-12B & 40 & 2 & $711$ (20) & Pixtral-12B & 2 & $99.7$ & $100$ (2) \\
\midrule
\multicolumn{8}{l}{\emph{Bases whose LVLMs do not spike}} \\
Qwen3-1.7B & 28 & 2 & $15{,}136$ (18) & Qwen3-VL-2B & 2 & $0.0$ & $0$ \\
Qwen2.5-3B & 36 & 2 & $41{,}199$ (17) & Qwen2.5-VL-3B & 2 & $0.0$ & $0$ \\
Qwen2.5-7B & 28 & 3 & $21{,}834$ (11) & Qwen2.5-VL-7B & 3 & $0.0$ & $0$ \\
 & & & & Qwen2.5-Omni-7B & 3 & $0.0$ & $0$ \\
Qwen2.5-72B & 80 & 1 & $28{,}977$ (45) & Qwen2.5-VL-72B & 1 & $0.0$ & $0$ \\
Qwen2-7B & 28 & 3 & $10{,}597$ (11) & Qwen2-VL-7B & 3 & $0.0$ & $0$ \\
 & & & & LLaVA-OV-7B & 3 & $0.0$ & $0$ \\
 & & & & Molmo-7B-D & 3 & $0.0$ & $0$ \\
InternLM2.5-7B & 32 & 1 & $44{,}078$ (19) & InternVL2-8B & 1 & $0.0$ & $0$ \\
Vicuna-7B & 32 & 1 & $202$ (18) & LLaVA-1.5-7B & 1 & $0.0$ & $73$ (1) \\
Vicuna-13B & 40 & 3 & $330$ (14) & LLaVA-1.5-13B & 3 & $0.0$ & $0$ \\
\addlinespace[2pt]
GLM-4-9B & 40 & 27 & $364$ (25) & GLM-4.1V-9B-Base & 27 & $0.0$ & $0$ \\
DeepSeek-LLM-7B & 30 & 1 & $276$ (18) & Janus-Pro-7B & 1 & $0.0$ & $0$ \\
Gemma-2-2B & 26 & 9 & $10$ (9) & PaliGemma 2-3B-mix-448 & 9 & $0.0$ & $100$ (24) \\
Llama-3.1-8B & 32 & 1 & $74$ (17) & Idefics3-8B & 1 & $0.0$ & $0$ \\
\midrule
\multicolumn{8}{l}{\emph{Held out, the rules fixed before the runs}} \\
Qwen3-0.6B & 28 & 2 & $2{,}054$ (18) & InternVL3.5-1B & n/a & $0.0$ & $0$ \\
Qwen2.5-14B & 48 & 5 & $136{,}076$ (22) & InternVL3-14B & 5 & $0.0$ & $0$ \\
\bottomrule
\end{tabular}}
\end{table}

\FloatBarrier
\section{Where the visual spike forms}
\label{app:location}
\backto*{app:location}
This appendix gives the additional analyses for Section~\ref{sec:location}.

\FloatBarrier
\subsection{The shared part and spike location}
\label{app:siting}
\backto*{app:siting}
\textbf{Token size at rest does not pick the spiking images.}\quad Within a model, three summaries of an image's token sizes fail to separate the spiking images from the rest.
The Spearman correlations with the image's smallest token, its lowest decile and its median lie between $-0.35$ and $+0.05$ in the eight spiking models whose census has both spiking and non-spiking images, while the spike token lies in the lowest quarter of its image by the shared part on $71$ to $100\%$ of spiking images (Table~\ref{tab:decile}).

\begin{table}[!htbp]
\caption{\textbf{In every model the spike token sits in the least-shared quarter of its image on at least $71\%$ of images.} The share of images whose spike token lies in the image's lowest decile and lowest quarter, by the shared part $p$ and by the norm, over the 300 census images per model.
The ten spiking models, their spiking images counted.
$^\dagger$Fewer than ten images, indicative.\protect\backto{tab:decile}}
\label{tab:decile}
\begin{center}\small
\setlength{\tabcolsep}{3pt}
\begin{tabular}{lrrrrr}
\toprule
model & $n$ spiking & lowest decile, $p$ & lowest quarter, $p$ & lowest decile, norm & lowest quarter, norm \\
\midrule
Qwen3-VL-8B & 38 & 92.1\% & 97.4\% & 97.4\% & 97.4\% \\
Qwen3-VL-4B & 226 & 67.7\% & 98.2\% & 69.5\% & 93.4\% \\
Qwen3-VL-32B & 140 & 45.7\% & 70.7\% & 49.3\% & 74.3\% \\
Qwen2.5-VL-32B & 156 & 97.4\% & 100.0\% & 0.0\% & 7.1\% \\
InternVL3.5-4B & 49 & 42.9\% & 89.8\% & 12.2\% & 53.1\% \\
InternVL3.5-8B & 185 & 60.5\% & 87.0\% & 39.5\% & 70.3\% \\
InternVL3.5-14B & 207 & 70.5\% & 91.3\% & 58.0\% & 90.3\% \\
InternVL3.5-38B$^\dagger$ & 5 & 100.0\% & 100.0\% & 80.0\% & 100.0\% \\
Molmo2-8B & 300 & 71.0\% & 94.3\% & 82.7\% & 98.7\% \\
Pixtral-12B & 299 & 85.6\% & 95.0\% & 0.0\% & 0.0\% \\
\bottomrule
\end{tabular}
\end{center}
\end{table}

\begin{figure}[!htbp]
\begin{center}
\includegraphics[width=\linewidth]{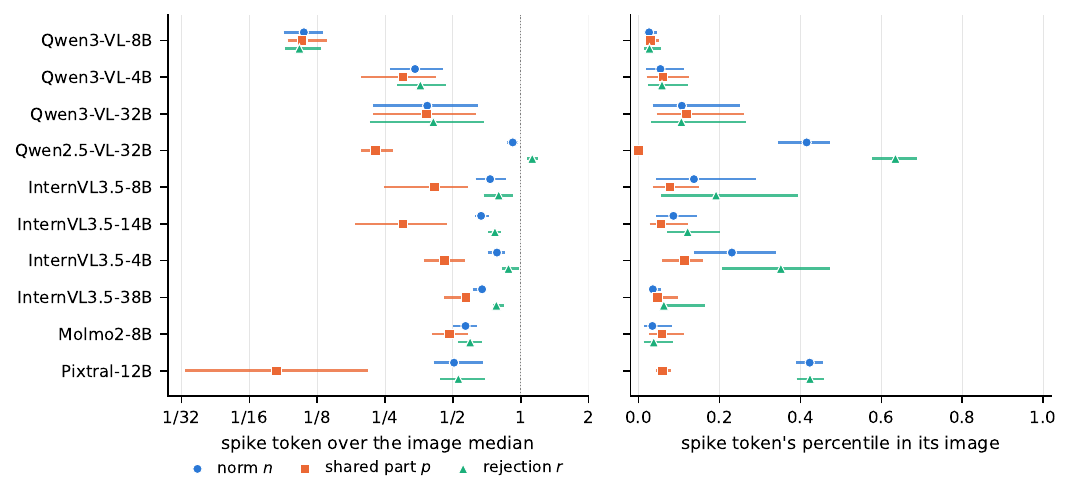}
\end{center}
\caption{\textbf{The spike token is an ordinary-sized token with an unusually small shared part.} The ten spiking models, their spiking census images, image tokens entering below the floor.
Left: the spike token over its image's median for the norm, the shared part and the rejection, median and interquartile range over spiking images.
Right: the spike token's percentile in its image by each of the three, $0$ the lowest token.\protect\backto{fig:anatomy} }
\label{fig:anatomy}
\end{figure}

\textbf{The rest of the token.}\quad The part of the token orthogonal to the common direction is the rejection $\mathrm{rej}_{\vb{m}_{-i}}(\vb{x}_i) = \vb{x}_i - p_i \hat{\vb{u}}_{-i}$, of size $r_i$, so that $\lVert \vb{x}_i\rVert^2 = p_i^2 + r_i^2$.
Figure~\ref{fig:anatomy} draws all three on the spike token, and Table~\ref{tab:locator} reads each as a locator.

\textbf{The shared part and the alignment with the trigger.}\quad Table~\ref{tab:c4corr} reads, per model, how two quantities rank the same image's tokens, the shared part at layer 0 and the cosine with the base's trigger at the visual switch's input, on the census forwards of Table~\ref{tab:trigcos}.
The correlation also holds on non-spiking images, with a median of $-0.28$ across the nine spiking models that have such images, so a low shared part predicts later trigger alignment whether or not the image produces a spike.

\begin{table}[!htbp]
\caption{\textbf{The less a token shares with its image at layer 0, the more it aligns with the trigger at the switch.} Per LVLM, the Spearman correlation across each census image's tokens between the layer-0 shared part and the cosine with the base's trigger at the visual switch's input.
Columns: median and interquartile range over spiking images, share negative, median over non-spiking images, and the mean cosine of the image's lowest tenth by the shared part, its highest tenth and the rest, pooled over spiking images.
$^\dagger$Five spiking images, indicative.}
\label{tab:c4corr}
\begin{center}\scriptsize
\setlength{\tabcolsep}{3pt}
\begin{tabular}{lrrrrr}
\toprule
LVLM & spiking & $\rho$, spiking images & negative & $\rho$, non-spiking & cosine, lowest tenth, highest tenth, rest \\
\midrule
Qwen3-VL-4B & 226 & $-0.33$ ($-0.40$, $-0.27$) & 100\% & $-0.28$ & $0.010$, $-0.014$, $-0.010$ \\
InternVL3.5-4B & 49 & $-0.54$ ($-0.61$, $-0.48$) & 100\% & $-0.50$ & $0.006$, $-0.016$, $-0.008$ \\
Qwen3-VL-8B & 38 & $-0.05$ ($-0.09$, $0.03$) & 65.8\% & $0.01$ & $-0.005$, $-0.011$, $-0.012$ \\
InternVL3.5-8B & 185 & $-0.31$ ($-0.37$, $-0.23$) & 100\% & $-0.29$ & $-0.003$, $-0.018$, $-0.015$ \\
InternVL3.5-14B & 207 & $-0.38$ ($-0.45$, $-0.31$) & 99.5\% & $-0.34$ & $0.011$, $-0.002$, $0.001$ \\
Qwen3-VL-32B & 140 & $-0.26$ ($-0.31$, $-0.19$) & 99.3\% & $-0.26$ & $0.008$, $0.001$, $0.003$ \\
Qwen2.5-VL-32B & 156 & $-0.12$ ($-0.19$, $-0.02$) & 76.9\% & $-0.16$ & $-0.008$, $-0.016$, $-0.013$ \\
InternVL3.5-38B$^\dagger$ & 5 & $-0.26$ ($-0.32$, $-0.21$) & 100\% & $-0.25$ & $0.005$, $-0.001$, $0.001$ \\
Molmo2-8B & 300 & $-0.32$ ($-0.41$, $-0.25$) & 100\% & -- & $0.005$, $-0.020$, $-0.017$ \\
Pixtral-12B & 299 & $-0.57$ ($-0.61$, $-0.54$) & 100\% & $-0.65$ & $-0.007$, $-0.028$, $-0.024$ \\
\bottomrule
\end{tabular}
\end{center}
\end{table}

\textbf{The background association and its base rate.}\quad Most of a COCO image is background, $60$ to $75\%$ of tokens under COCO instance masks.
Table~\ref{tab:bgassoc} counts the spike tokens that land there, over the spiking images of the edit set.
The association holds in all nine spiking models measured, with lifts of $+12.2$ to $+23.9$ points over each image's own base rate, and it is invariant to the background threshold.
\citet{kang2025see} report $90$ to $94\%$ on LLaVA-1.5-7B under a main-object definition of background that covers $83$ to $91\%$ of all visual tokens, a lift of $3$ to $8$ points, so we report the lift too.

\begin{figure}[!htbp]
\begin{center}
\includegraphics[width=\linewidth]{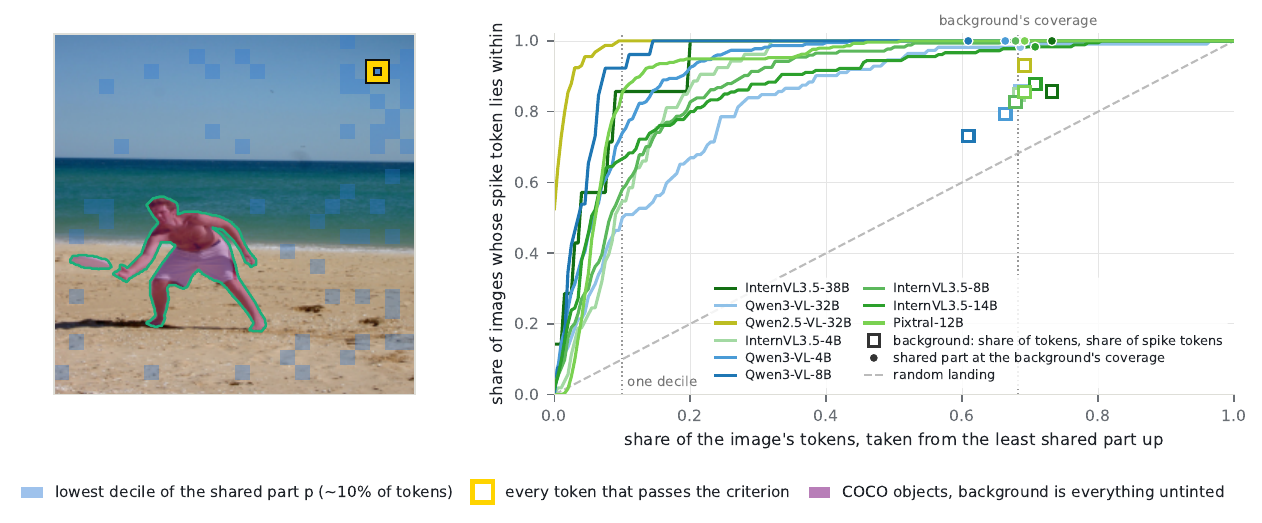}
\end{center}
\caption{\textbf{At equal coverage the shared part locates more spike tokens than background does.} Left: one image under Qwen3-VL-8B, COCO objects tinted, the shared part's lowest decile in blue, tokens meeting the criterion marked.
Right: the spiking models, their spiking images, the share whose spike token falls in the given fraction of the image's tokens, taken from the smallest shared part up. Molmo2-8B is absent because its overlapping crops have no single token grid on which to place background, and Pixtral-12B is read on the census.
Each square is background, token share against spike-token share; the dot of its colour is the curve there.\protect\backto{fig:bgdrive}}
\label{fig:bgdrive}
\end{figure}

\begin{table}[!htbp]
\caption{\textbf{The spike token lands in background more often than a random token would.} Over the spiking images of each spiking model: the share of spike tokens under no COCO object mask, the background's share of the image's tokens, the difference in points, and the $z$ of that count against a random-landing null with its one-sided $p$.
Pixtral-12B is read on the census, on its $32 \times 32$ token grid; Molmo2-8B's overlapping crops have no single grid.
$^\dagger$Fewer than ten images, indicative.\protect\backto{tab:bgassoc}}
\label{tab:bgassoc}
\begin{center}\scriptsize
\setlength{\tabcolsep}{3.5pt}
\begin{tabular}{lrrrrrr}
\toprule
model & $n$ & spike in bg & base rate & lift & $z$ & $p$ \\
\midrule
Qwen2.5-VL-32B & 158 & 93.0\% & 69.2\% & $+23.9$ pp & 7.48 & $<10^{-4}$ \\
Qwen3-VL-32B & 112 & 85.7\% & 68.6\% & $+17.2$ pp & 4.56 & $<10^{-4}$ \\
InternVL3.5-14B & 180 & 87.8\% & 70.8\% & $+17.0$ pp & 5.79 & $<10^{-4}$ \\
InternVL3.5-4B & 53 & 84.9\% & 68.3\% & $+16.6$ pp & 3.06 & 0.0011 \\
Pixtral-12B & 296 & 85.5\% & 69.2\% & $+16.2$ pp & 7.21 & $<10^{-4}$ \\
InternVL3.5-8B & 174 & 82.8\% & 67.9\% & $+14.9$ pp & 4.98 & $<10^{-4}$ \\
Qwen3-VL-4B & 227 & 79.3\% & 66.3\% & $+12.9$ pp & 4.88 & $<10^{-4}$ \\
InternVL3.5-38B$^\dagger$ & 7 & 85.7\% & 73.2\% & $+12.5$ pp & 0.78 & 0.22 \\
Qwen3-VL-8B & 26 & 73.1\% & 60.9\% & $+12.2$ pp & 1.58 & 0.06 \\
\bottomrule
\end{tabular}
\end{center}
\end{table}

\begin{table}[!htbp]
\caption{\textbf{Given background the shared part still locates the spike token.
Given the shared part, background adds little.} For a token of a spiking image, $S$ is being the spike token, $D$ being among the tenth of the image's tokens with the smallest shared part, and $B$ being background.
Columns 5 and 8 are the differences $P(D \mid S, B) - P(D \mid B)$ and $P(B \mid S, D) - P(B \mid D)$, in points.
Pixtral-12B is read on the census.
$^\dagger$Fewer than ten images, indicative.\protect\backto{tab:bgcond}}
\label{tab:bgcond}
\begin{center}
\scriptsize
\setlength{\tabcolsep}{3pt}
\resizebox{\textwidth}{!}{\begin{tabular}{lrrrrrrr}
\toprule
model & $n$ & $P(D \mid S, B)$ & $P(D \mid B)$ & the shared part adds & $P(B \mid S, D)$ & $P(B \mid D)$ & background adds \\
\midrule
Qwen3-VL-8B & 26 & 94.7\% & 11.2\% & $+83.5$ & 75.0\% & 68.5\% & $+6.5$ \\
Qwen3-VL-4B & 227 & 73.3\% & 11.7\% & $+61.6$ & 80.5\% & 77.6\% & $+2.9$ \\
Qwen3-VL-32B & 112 & 50.0\% & 11.9\% & $+38.1$ & 88.9\% & 81.5\% & $+7.4$ \\
Qwen2.5-VL-32B & 158 & 100.0\% & 12.0\% & $+88.0$ & 93.0\% & 83.5\% & $+9.5$ \\
InternVL3.5-4B & 53 & 51.1\% & 12.0\% & $+39.1$ & 82.1\% & 84.1\% & $-1.9$ \\
InternVL3.5-8B & 174 & 55.6\% & 12.2\% & $+43.3$ & 82.5\% & 85.0\% & $-2.6$ \\
InternVL3.5-14B & 180 & 67.1\% & 11.8\% & $+55.3$ & 89.1\% & 85.2\% & $+3.9$ \\
InternVL3.5-38B$^\dagger$ & 7 & 83.3\% & 11.9\% & $+71.4$ & 83.3\% & 89.1\% & $-5.8$ \\
Pixtral-12B & 296 & 83.4\% & 9.9\% & $+73.5$ & 84.1\% & 69.0\% & $+15.1$ \\
\bottomrule
\end{tabular}}
\end{center}
\end{table}

\textbf{The asymmetry in conditional-probability form.}\quad For a token in a spiking forward, let $S$ mean it is the spike token, $D$ that it is among the tenth of the image's tokens with the smallest shared part and $B$ that it is in background.
Given background, the spike token is in that decile at $P(D \mid S, B) = 0.50$ to $1.00$ against $P(D \mid B) = 0.10$ to $0.12$ for a random background token.
Given the decile, it is background at $P(B \mid S, D) = 0.75$ to $0.93$ against $P(B \mid D) = 0.69$ to $0.89$ (Table~\ref{tab:bgcond}).
The shared part is the selection rule and background its correlate.

\textbf{What each part of the token contributes.}\quad Table~\ref{tab:locator} reads the token's size alone, its direction alone with the size divided out, the rejection and the shared part as locators of the spike token against the other tokens of the same image.
Neither size nor direction works everywhere.
The direction alone carries Qwen2.5-VL-32B and collapses in Qwen3-VL-8B and Qwen3-VL-32B. The norm alone is the better of the two in six models and fails in the one the direction carries.
The shared part beats both in five of the ten.

\begin{table}[!htbp]
\caption{\textbf{The shared part beats both the token's size and its direction in five of the ten models, and neither of those two holds everywhere.} The ten spiking models on their spiking census forwards, each score read at the language model's input.
Positives are the spike tokens, counted in the second column.
AUC is the within-image area, higher better; FPR@95 the share of non-spike tokens above the threshold covering $95\%$ of spike tokens.
Last row: median over the ten models.
$^\dagger$Five spiking images, indicative.\protect\backto{tab:locator}}
\label{tab:locator}
\begin{center}\scriptsize
\setlength{\tabcolsep}{4pt}
\begin{tabular}{lrrrrr}
\toprule
model & spike tokens & shared part $p$ & direction & norm & rejection $r$ \\
 & & AUC / FPR@95 & AUC / FPR@95 & AUC / FPR@95 & AUC / FPR@95 \\
\midrule
Qwen3-VL-4B & 366 & 0.926 / 20\% & 0.721 / 81\% & 0.919 / 27\% & 0.900 / 37\% \\
InternVL3.5-4B & 55 & 0.880 / 28\% & 0.894 / 35\% & 0.755 / 53\% & 0.647 / 69\% \\
Qwen3-VL-8B & 39 & 0.957 / 14\% & 0.572 / 98\% & 0.950 / 9\% & 0.943 / 13\% \\
InternVL3.5-8B & 282 & 0.873 / 41\% & 0.811 / 79\% & 0.792 / 64\% & 0.727 / 81\% \\
InternVL3.5-14B & 341 & 0.880 / 51\% & 0.845 / 71\% & 0.866 / 44\% & 0.820 / 53\% \\
Qwen2.5-VL-32B & 156 & 0.988 / 6\% & 0.991 / 3\% & 0.590 / 57\% & 0.381 / 79\% \\
Qwen3-VL-32B & 162 & 0.818 / 63\% & 0.529 / 97\% & 0.834 / 58\% & 0.833 / 55\% \\
InternVL3.5-38B$^\dagger$ & 5 & 0.940 / 10\% & 0.868 / 32\% & 0.935 / 19\% & 0.851 / 46\% \\
Molmo2-8B & 2{,}013 & 0.905 / 29\% & 0.658 / 88\% & 0.935 / 22\% & 0.927 / 26\% \\
Pixtral-12B & 403 & 0.763 / 80\% & 0.838 / 77\% & 0.494 / 86\% & 0.488 / 87\% \\
\midrule
median & & 0.893 / 28\% & 0.824 / 78\% & 0.850 / 48\% & 0.826 / 54\% \\
\bottomrule
\end{tabular}
\end{center}
\end{table}

\textbf{Siting under noise.}\quad Under Gaussian noise at $\sigma = 0.45$ on 150 images per model, the spike token sits at the 0th to 18th percentile of its forward's tokens by the shared part in nine of the ten spiking models, as on the clean images, while the number of spiking images rises, from 13 to 113 in Qwen3-VL-8B and from 24 to 134 in InternVL3.5-4B.
In Molmo2-8B, which spikes on every image clean and noised, it sits near the middle, at the 45th percentile clean and the 57th noised.

\FloatBarrier
\textbf{What the shared part reads, in full.}\quad Table~\ref{tab:decode} gives, for eight images of the edit set under Qwen3-VL-4B, the model's description with the real image tokens and with every image position filled by the image's mean token, its most-shared token, its least-shared token or its spike token, the DeepStack features removed.
In Qwen3-VL-4B, the mean token gives the scene, the most-shared token a salient object of it, and the least-shared token a placeholder.
Across the ten spiking models, a description built from the spike token names the scene on $0$ to $12\%$ of images, and on one of InternVL3.5-38B's three, against $94$ to $100\%$ with the real tokens, and it reads as a repetitive pattern only in the three Qwen3-VL models (Table~\ref{tab:decodesum}).
The layer-0 logit lens of those tokens reads the same way.

For the Qwen3-VL-8B example in Figure~\ref{fig:one}, the most-shared token's cosine with the trigger at the switch input is $-0.002$, while the spike token's is $0.26$, the largest in the image.
Their two highest layer-0 logit-lens words are \emph{weather} and \emph{alley} for the most-shared token, and \emph{hap} and \emph{compass} for the spike token.

\begin{table}[!htbp]
\caption{\textbf{In Qwen3-VL-4B the spike token decodes to a repetitive pattern rather than to its own position's content.} Qwen3-VL-4B, eight images on the $16 \times 16$ token grid the model reads, the most-shared token outlined in blue, the least-shared in grey and the spike token in red.
Beside each, the description with the real tokens and with every image position filled by the image's mean token or its most-shared, least-shared or spike token, each followed, in blue italics, by the three highest tokens of that vector's layer-0 logit lens.
The eighth image does not spike, and its red cell and last column read its peak token.\protect\backto{tab:decode}}
\label{tab:decode}
\begin{center}
\scriptsize
\setlength{\tabcolsep}{2pt}
\begin{tabular}{@{}c@{\hspace{4pt}}p{2.15cm}p{2.15cm}p{2.15cm}p{2.15cm}p{2.15cm}@{}}
\toprule
image & real tokens & the mean token & \textcolor{sharedc}{\rule{4.5pt}{4.5pt}}~the most-shared token & \textcolor{leastc}{\rule{4.5pt}{4.5pt}}~the least-shared token & \textcolor{spikec}{\rule{4.5pt}{4.5pt}}~the spike token \\
\midrule
\decodeimg{1} & A vintage steam locomotive numbered 71 is illuminated at dusk, standing proudly on\ldots & a dark, vintage-style steam locomotive with multiple large driving wheels and a prominent\ldots \textcolor{lens}{\emph{lens: kai (a name), end, seat}} & a long, dark street illuminated by a series of evenly spaced, warm orange\ldots \textcolor{lens}{\emph{lens: lighting, lamp fixture, lamp}} & a repetitive, monotonous pattern of the same placeholder or empty square icon, stacked\ldots \textcolor{lens}{\emph{lens: more, @example, lé}} & a repetitive, identical pattern of a single small, indistinct square or pixelated block,\ldots \textcolor{lens}{\emph{lens: yne, a set, more}} \\
\decodeimg{2} & A colorful plate of stir-fried chicken, broccoli, carrots, and cauliflower served with a\ldots & a close-up of a plate of food, likely a stir-fry or curry, featuring\ldots \textcolor{lens}{\emph{lens: dining table, dining, bar}} & a single fork with multiple tines, repeatedly duplicated in a row, creating a\ldots \textcolor{lens}{\emph{lens: yd, clip, kiss}} & a repetitive, monotonous pattern of the word ``tonon'' in a uniform, unchanging font\ldots \textcolor{lens}{\emph{lens: ayne, jou, jon}} & a close-up of a single, slightly blurry, spiral-shaped food item, possibly a cooked\ldots \textcolor{lens}{\emph{lens: shape, Shapes, shape}} \\
\decodeimg{3} & A young boy in a Star Wars t-shirt walks a brown and white\ldots & a black dog with a white patch on its chest, standing on a\ldots \textcolor{lens}{\emph{lens: qi (qilin), column, trs}} & a person wearing a beige, long-sleeved, button-up shirt and matching beige pants, standing\ldots \textcolor{lens}{\emph{lens: brown, uder, kiss}} & a series of identical, empty, and slightly blurred square frames arranged in a\ldots \textcolor{lens}{\emph{lens: lé, more, yne}} & a repetitive, visually striking pattern of the word ``save'' written in a bold,\ldots \textcolor{lens}{\emph{lens: aid, ej, aid}} \\
\decodeimg{4} & A person holds a white Nintendo Wii remote in front of a monitor\ldots & a row of identical white, rectangular objects with a small circular detail on\ldots \textcolor{lens}{\emph{lens: kai (a name), áo, roe deer}} & a row of identical white Wii controllers, each with a slightly different angle\ldots \textcolor{lens}{\emph{lens: pale, transparent, simple}} & a long horizontal line of repeated text reading ``MAX'' in a bold, sans-serif\ldots \textcolor{lens}{\emph{lens: thwart, lesson, yla}} & a single, repeated, and identical pattern of a small, square-shaped icon or symbol,\ldots \textcolor{lens}{\emph{lens: more, plain, wid}} \\
\decodeimg{5} & A cozy, well-lit hotel room features two neatly made beds with plush towels,\ldots & a row of identical dark brown wooden doors with a simple, modern design,\ldots \textcolor{lens}{\emph{lens: beds, lodging, beside}} & a stack of neatly folded, dark-colored towels arranged in a repeating pattern, with\ldots \textcolor{lens}{\emph{lens: OfClass, pillow, .same}} & a repetitive, stacked pattern of identical, empty, square-shaped placeholders, creating a visually uniform\ldots \textcolor{lens}{\emph{lens: more, composition, a group}} & a single, stylized, repeating pattern of a dark green leaf or leaf-like shape\ldots \textcolor{lens}{\emph{lens: lodging, hotel, bar}} \\
\decodeimg{6} & A freshly baked pizza with tomatoes, spinach, and onions sits on a metal\ldots & a close-up of a round, thin-crust pizza with a golden-brown surface, topped with\ldots \textcolor{lens}{\emph{lens: end, bar, canteen}} & a stack of identical, empty, square coasters arranged in a neat, repeating pattern. \textcolor{lens}{\emph{lens: beside, porter, thick}} & a repetitive, static pattern of the same word ``tomato'' written multiple times in\ldots \textcolor{lens}{\emph{lens: persuasion, goosefoot, ying (Britain)}} & a single, slightly blurry, unremarkable photograph of a plain white surface with no\ldots \textcolor{lens}{\emph{lens: bar, dining table, beer}} \\
\decodeimg{7} & A well-arranged buffet counter displays an assortment of fresh bread, sandwiches, cheeses, meats,\ldots & a long, flat surface with multiple small, square white plates arranged in a\ldots \textcolor{lens}{\emph{lens: on the table, end, kitchen}} & a repeating pattern of white, creamy substances, likely whipped cream or yogurt, arranged\ldots \textcolor{lens}{\emph{lens: dawn, oples, one's own}} & a repeating, static pattern of the same digital timestamp ``2023-04-05 12:34:56'' across multiple\ldots \textcolor{lens}{\emph{lens: Davies, composite, they}} & a repetitive sequence of the word ``fast'' written in a simple, sans-serif font,\ldots \textcolor{lens}{\emph{lens: breakfast, breakfast, mornings}} \\
\decodeimg{8} & A rider on a white horse performs a dramatic jumping maneuver in front\ldots & A group of people in matching dark jackets and white shirts are seated\ldots \textcolor{lens}{\emph{lens: bar, -sk, driver}} & a single, repeated, and highly stylized depiction of a woman with curly reddish-brown\ldots \textcolor{lens}{\emph{lens: shape, Hair, a batch}} & a repeating, stylized pattern of the number ``3'' in a dark, solid color\ldots \textcolor{lens}{\emph{lens: buffer, /topics, despre}} & a serene, slightly blurry landscape of a tree-lined field under a pale sky,\ldots \textcolor{lens}{\emph{lens: insert, edException, hari}} \\
\bottomrule
\end{tabular}
\end{center}
\end{table}

\begin{table}[!htbp]
\caption{\textbf{The spike token rarely carries the scene.} Per spiking model, up to 50 spiking census images, the whole spiking pool for Qwen3-VL-8B and InternVL3.5-38B: the description with the real image tokens and with every image position filled by the mean, most-shared, least-shared or spike token at the language model's input. Each cell gives the share of descriptions naming a repetitive pattern, then the share naming the scene, with at least two content words shared with COCO's five reference captions.\protect\backto{tab:decodesum}}
\label{tab:decodesum}
\begin{center}
\footnotesize
\setlength{\tabcolsep}{4pt}
\begin{tabular}{lrrrrrr}
\toprule
 & & \multicolumn{5}{c}{repetitive, scene (\%)} \\
model & images & real & mean & most-shared & least-shared & spike \\
\midrule
Qwen3-VL-4B & 50 & 10, 100 & 38, 74 & 66, 20 & 100, 0 & 98, 4 \\
Qwen3-VL-8B & 34 & 3, 94 & 53, 56 & 74, 15 & 100, 0 & 100, 6 \\
Qwen3-VL-32B & 50 & 10, 100 & 46, 32 & 70, 20 & 98, 0 & 66, 12 \\
Qwen2.5-VL-32B & 50 & 8, 96 & 18, 56 & 32, 40 & 30, 2 & 4, 0 \\
InternVL3.5-4B & 50 & 4, 96 & 16, 32 & 26, 8 & 8, 16 & 2, 10 \\
InternVL3.5-8B & 50 & 2, 100 & 6, 46 & 28, 20 & 4, 4 & 6, 6 \\
InternVL3.5-14B & 50 & 8, 100 & 14, 40 & 22, 18 & 4, 8 & 0, 8 \\
InternVL3.5-38B & 3 & 33, 100 & 0, 100 & 33, 67 & 33, 0 & 0, 33 \\
Molmo2-8B & 50 & 6, 98 & 20, 18 & 28, 18 & 12, 0 & 6, 8 \\
Pixtral-12B & 50 & 4, 98 & 14, 38 & 0, 10 & 0, 0 & 0, 0 \\
\bottomrule
\end{tabular}
\end{center}
\end{table}

\FloatBarrier
\subsection{Selection and alignment of the spike token}
\label{app:formation}
\backto*{app:formation}
We give three readings on the spiking models whose base trigger is stored, over the census images under the description prompt, with eager attention.
The trigger is the base's, estimated as  \aref{app:drive} describes.

\textbf{Dependence on the earlier text spike.}\quad
In the nine models with a later visual switch, blocking attention from the future visual spike token to the text spike token leaves it spiking on only $0$ to $5\%$ of images in six models.
InternVL3.5-8B retains $25\%$, Molmo2-8B $15\%$, and InternVL3.5-38B $33\%$ of three images (Table~\ref{tab:formcausal}).
The text spike token's value is small and nearly input-independent, so the attention it receives contributes little content while changing what the remaining attention can read.
Pixtral-12B has no earlier text spike to attend to because its visual and text switches coincide.

\begin{table}[!htbp]
\caption{\textbf{The choice of spike token is settled late.} The spike token's whole state is exchanged with a random other image token's at the input of one block, every other token untouched.
Columns: the spiking census images, the visual switch, and the share of images on which the moved state spikes at its new slot, exchanged at the language model's input, near block 12, and one block before the switch.
Qwen3-VL-32B is exchanged from block 4, its visual features entering over the first three blocks.
Pixtral-12B's switch is block 2. InternVL3.5-38B spikes on 5 of its 300 census images, 3 of which enter this reading.\protect\backto{tab:swapdepth}}
\label{tab:swapdepth}
\begin{center}
\begin{tabular}{lrrrrr}
\toprule
model & images & switch & at the input & near block 12 & one block before \\
\midrule
Qwen3-VL-4B & 100 & 16 & 0.35 & 0.61 & 0.86 \\
Qwen3-VL-8B & 33 & 16 & 0.15 & 0.36 & 0.73 \\
Qwen3-VL-32B & 60 & 43 & 0.22 & 0.20 & 0.77 \\
InternVL3.5-4B & 49 & 16 & 0.35 & 0.57 & 0.98 \\
InternVL3.5-8B & 80 & 16 & 0.11 & 0.33 & 0.63 \\
InternVL3.5-14B & 80 & 19 & 0.24 & 0.44 & 0.89 \\
Qwen2.5-VL-32B & 80 & 38 & 0.85 & 0.93 & 1.00 \\
InternVL3.5-38B & 3 & 43 & 0.00 & 0.00 & 0.00 \\
Molmo2-8B & 60 & 16 & 0.55 & 0.88 & 0.93 \\
Pixtral-12B & 60 & 2 & 1.00 & -- & 1.00 \\
\bottomrule
\end{tabular}
\end{center}
\end{table}

\begin{table}[!htbp]
\caption{\textbf{The alignment with the trigger is written in the last blocks.} Medians over spiking census forwards under the description prompt, up to 80 per model.
Columns: forwards, the visual switch $S$, the spike token's cosine with the trigger entering blocks $S-4$, $S-1$ and $S$, its attention and feed-forward writes along the trigger at $S-1$, and its attention write at $S$.
Qwen2.5-VL-32B's spike token is already aligned at $S-4$, and Pixtral-12B's switch at block 2 leaves no block $S-4$.\protect\backto{tab:aimorigin}}
\label{tab:aimorigin}
\begin{center}
\small
\begin{tabular}{lrrrrrrrr}
\toprule
 & & & \multicolumn{3}{c}{cosine with the trigger} & \multicolumn{2}{c}{writes at $S-1$} & at $S$ \\
model & forwards & $S$ & $S-4$ & $S-1$ & $S$ & attention & feed-forward & attention \\
\midrule
Qwen3-VL-4B & 80 & 16 & -0.01 & +0.01 & +0.27 & 3.8 & 9.5 & 2.4 \\
Qwen3-VL-8B & 35 & 16 & -0.02 & +0.00 & +0.22 & 4.8 & 11.3 & 0.8 \\
Qwen3-VL-32B & 60 & 43 & +0.01 & +0.03 & +0.09 & 0.5 & 8.6 & 20.1 \\
InternVL3.5-4B & 49 & 16 & -0.01 & +0.06 & +0.22 & -0.6 & 6.4 & 7.5 \\
InternVL3.5-8B & 80 & 16 & -0.01 & +0.00 & +0.21 & 3.6 & 12.6 & 15.5 \\
InternVL3.5-14B & 80 & 19 & -0.00 & +0.03 & +0.12 & 0.4 & 10.3 & 18.7 \\
Qwen2.5-VL-32B & 80 & 38 & +0.13 & +0.14 & +0.21 & -1.8 & 28.5 & -10.2 \\
InternVL3.5-38B & 3 & 43 & -0.00 & +0.02 & +0.06 & 1.7 & 6.9 & 16.6 \\
Molmo2-8B & 60 & 16 & -0.02 & +0.06 & +0.26 & 0.7 & 18.7 & 14.4 \\
Pixtral-12B & 60 & 2 & -- & +0.62 & +0.62 & -0.1 & -0.5 & -0.1 \\
\bottomrule
\end{tabular}
\end{center}
\end{table}

\begin{table}[!htbp]
\caption{\textbf{What the eruption needs.} The share of spiking census images whose spike token still spikes under one intervention.
Columns: images; the spike token's head-mean attention to the text spike token, averaged over the blocks up to the text switch and over those after it up to $S$, median over images; its cosine with the trigger at the input the switch's feed-forward reads, clean and with that attention blocked, median over images; the spike token denied the text spike token from the text switch to $S$; that token's massive entries zeroed; the spike token denied, at $S-1$ and $S$, the prompt position and five image tokens whose values aim most along the trigger; as many random positions; and the trigger component removed from its feed-forward writes, its attention writes, and both. n/a: Pixtral-12B's visual switch is its text switch.\protect\backto{tab:formcausal}}
\label{tab:formcausal}
\begin{center}
\footnotesize
\setlength{\tabcolsep}{4pt}
\resizebox{\linewidth}{!}{\begin{tabular}{lrccrrrrrrr}
\toprule
 & & attention to the & alignment, & \multicolumn{3}{c}{denied} & control & \multicolumn{3}{c}{trigger component removed} \\
model & images & text spike, before, after & clean, blocked & text spike & its entries & sources & random & feed-forward & attention & both \\
\midrule
Qwen3-VL-4B & 60 & 0.05, 0.65 & 0.60, -0.01 & 0.05 & 0.35 & 0.05 & 1.00 & 0.42 & 0.90 & 0.00 \\
Qwen3-VL-8B & 33 & 0.12, 0.60 & 0.55, 0.00 & 0.00 & 0.27 & 0.12 & 1.00 & 0.55 & 0.91 & 0.00 \\
Qwen3-VL-32B & 60 & 0.31, 0.77 & 0.51, -0.03 & 0.00 & 0.18 & 0.45 & 0.95 & 0.87 & 0.90 & 0.02 \\
InternVL3.5-4B & 49 & 0.10, 0.67 & 0.58, 0.03 & 0.00 & 0.82 & 0.86 & 1.00 & 0.02 & 1.00 & 0.00 \\
InternVL3.5-8B & 60 & 0.17, 0.61 & 0.63, 0.11 & 0.25 & 0.32 & 0.23 & 1.00 & 0.72 & 0.80 & 0.00 \\
InternVL3.5-14B & 60 & 0.33, 0.63 & 0.46, 0.06 & 0.02 & 0.05 & 0.90 & 1.00 & 0.88 & 0.70 & 0.25 \\
Qwen2.5-VL-32B & 60 & 0.09, 0.79 & 0.34, 0.02 & 0.00 & 0.72 & 1.00 & 1.00 & 0.47 & 1.00 & 0.33 \\
InternVL3.5-38B & 3 & 0.43, 0.39 & 0.32, 0.19 & 0.33 & 0.00 & 0.00 & 1.00 & 0.67 & 0.33 & 0.00 \\
Molmo2-8B & 60 & 0.26, 0.51 & 0.60, 0.05 & 0.15 & 1.00 & 0.95 & 1.00 & 0.27 & 1.00 & 0.00 \\
Pixtral-12B & 60 & n/a & 0.96, n/a & n/a & n/a & 1.00 & 1.00 & 1.00 & 1.00 & 1.00 \\
\bottomrule
\end{tabular}}
\end{center}
\end{table}

\begin{table}[!htbp]
\caption{\textbf{Without the text spike token, the eventual spike token reads larger values and loses its state.} The eventual spike token denied the text spike token from the block after the text switch to the visual switch $S$, on spiking census images, medians over images.
Columns: images; its head-mean attention on image tokens at the first denied block, clean and denied; the attention-weighted norm of the values it reads there over the text spike token's, denied; its attention writes over the denied blocks, denied over clean; its state entering $S$, cosine with the clean state; the share of images on which it still spikes.\protect\backto{tab:overwrite}}
\label{tab:overwrite}
\begin{center}
\footnotesize
\setlength{\tabcolsep}{4pt}
\begin{tabular}{lrrrrrrr}
\toprule
 & & \multicolumn{2}{c}{attention on image tokens} & & & & \\
model & images & clean & denied & values read & writes & state & still spikes \\
\midrule
Qwen3-VL-4B & 60 & 0.26 & 0.93 & 3.55 & 1.50 & 0.65 & 0.05 \\
Qwen3-VL-8B & 34 & 0.34 & 0.94 & 3.68 & 1.53 & 0.70 & 0.00 \\
Qwen3-VL-32B & 60 & 0.20 & 0.49 & 2.74 & 2.25 & 0.55 & 0.00 \\
InternVL3.5-4B & 50 & 0.23 & 0.92 & 3.73 & 1.70 & 0.63 & 0.00 \\
InternVL3.5-8B & 60 & 0.24 & 0.91 & 3.98 & 1.53 & 0.71 & 0.25 \\
InternVL3.5-14B & 60 & 0.47 & 0.60 & 1.89 & 1.28 & 0.61 & 0.02 \\
Qwen2.5-VL-32B & 60 & 0.11 & 0.53 & 6.48 & 2.43 & 0.30 & 0.00 \\
InternVL3.5-38B & 3 & 0.24 & 0.51 & 5.34 & 0.96 & 0.98 & 0.33 \\
Molmo2-8B & 60 & 0.31 & 0.83 & 4.85 & 1.37 & 0.73 & 0.15 \\
\bottomrule
\end{tabular}
\end{center}
\end{table}

\FloatBarrier
\subsection{Controlling the spike location}
\label{app:relocation}
\begin{figure}[!htbp]
\begin{center}
\includegraphics[width=\linewidth]{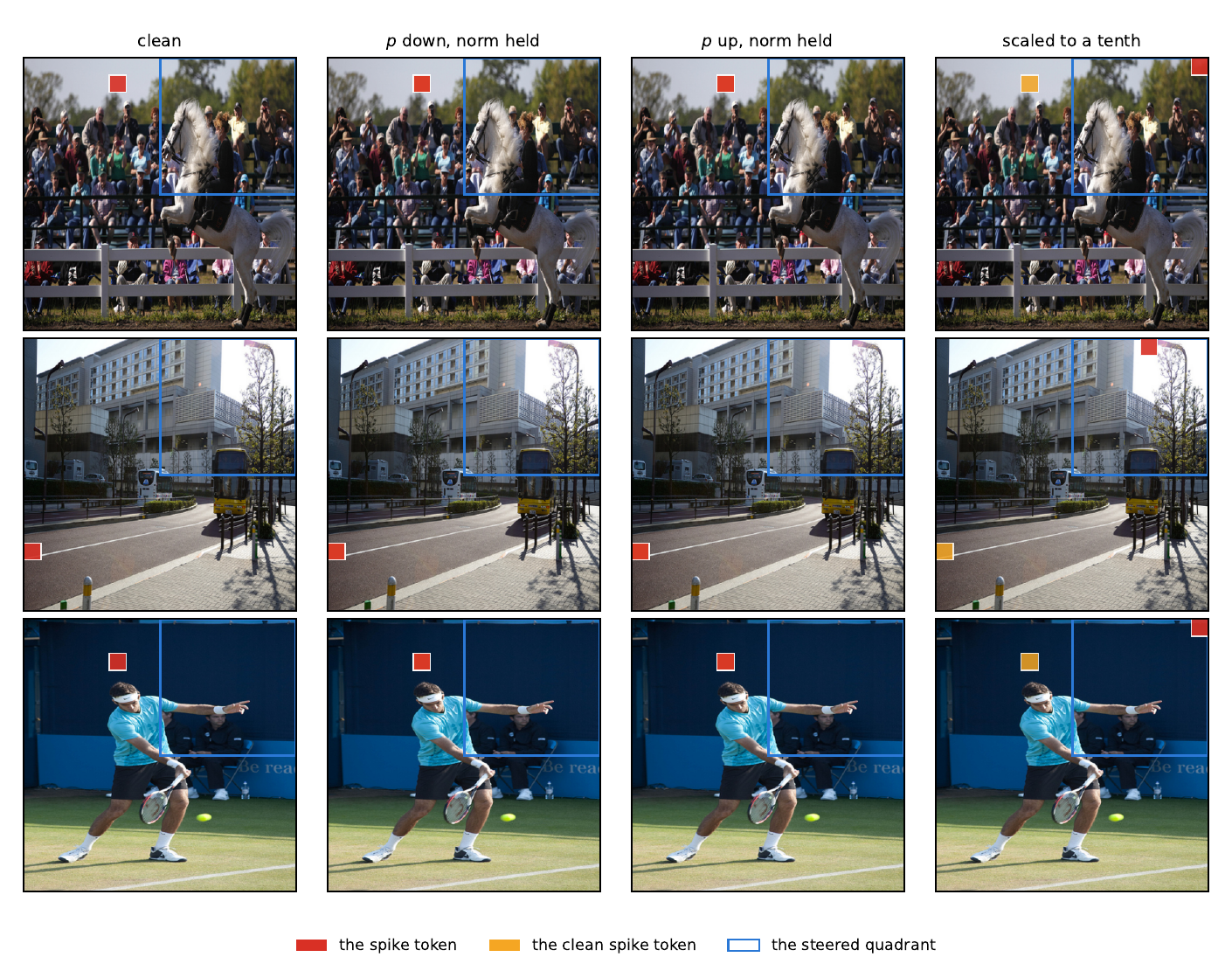}
\end{center}
\caption{\textbf{Moving where the spike lands.} InternVL3.5-8B on three images whose clean spike token sits outside the outlined quadrant.
Columns: the clean forward, then the quadrant's tokens with their shared part cut to a tenth and with it quadrupled, the norm restored after each, and scaled to a tenth.
The red square is the spike token, the amber square is the clean one, and the outline is the target quadrant.\protect\backto{fig:steer}}
\label{fig:steer}
\end{figure}

\begin{figure}[!htbp]
\begin{center}
\includegraphics[width=\linewidth]{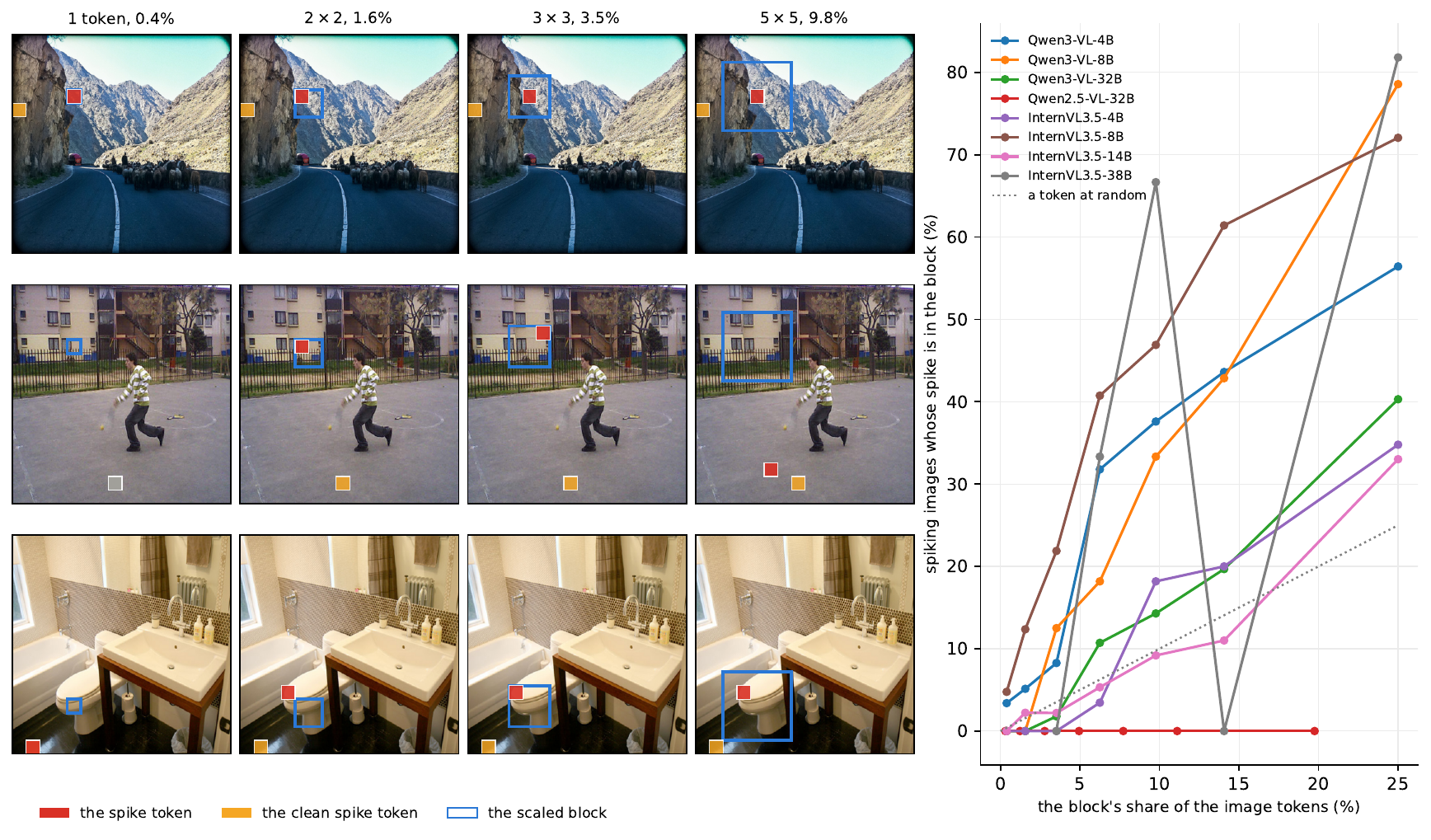}
\end{center}
\caption{\textbf{The spike can be placed in a tenth of the image, or in one token.} Left, InternVL3.5-8B on three images whose clean spike sits outside the block.
Columns: a block of 1 token, $2\times2$, $3\times3$ and $5\times5$ of a $16\times16$ grid, scaled to a tenth at the input.
Red is the spike token, amber the clean one, grey a peak without a spike.
Right, the share of spiking images whose spike lands in the block, against the block's share of the tokens.\protect\backto{fig:blocks}}
\label{fig:blocks}
\end{figure}

\begin{figure}[!htbp]
\begin{center}
\includegraphics[width=\linewidth]{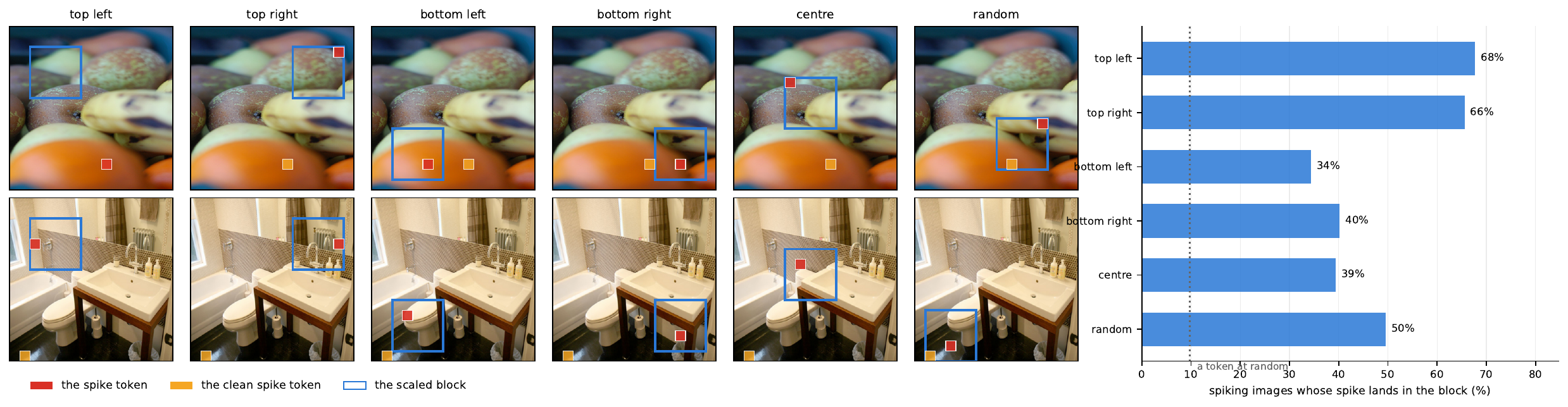}
\end{center}
\caption{\textbf{The place is ours to choose.} InternVL3.5-8B on two images, the same block of $25$ tokens moved to the centre of each quadrant, to the image's centre, and to a position drawn at random, scaled to a tenth at the input.
Red is the spike token, amber the clean one, and the outline the block.
Right, the share of spiking images whose spike lands in the block at each position, over 150 images, the dotted line a token landing at random.\protect\backto{fig:positions}}
\label{fig:positions}
\end{figure}

\backto*{app:relocation}
\textbf{Deciding the site.}\quad We describe here the edits behind Figure~\ref{fig:blocks} and the tables below.
With $\hat{\vb{m}}$ the unit mean of the image's tokens at layer 0 and $p_i = \vb{x}_i \cdot \hat{\vb{m}}$, the chosen region's tokens are replaced by
\begin{equation}
\vb{x}_i' = \lVert \vb{x}_i \rVert \frac{\vb{x}_i - c\, p_i \hat{\vb{m}}}{\lVert \vb{x}_i - c\, p_i \hat{\vb{m}} \rVert}, \quad c \in \{0.9, -3\}, \qquad \text{or} \qquad \vb{x}_i' = \tfrac{1}{10} \vb{x}_i .
\label{eq:steer}
\end{equation}
The first form moves the shared part and then restores each token's own norm, down at $c = 0.9$ and up at $c = -3$, so the part that remains is not exactly a tenth or four times the original. The second lowers the shared part and the norm together.
The mean here includes the token itself, where the locator of Section~\ref{sec:location} leaves it out.
On a quadrant, the second form puts the spike inside it on $33$ to $82\%$ of spiking images in seven of the nine models tested, all but Molmo2-8B, whose overlapping crops have no single token grid, and the first does so in an eighth, on $68\%$. The clean forward puts it there on $9$ to $19\%$ and Pixtral-12B stays at its clean $9\%$ under every edit.
Raising the shared part at fixed norm keeps the spike out, at $0$ to $15\%$ in seven of the nine (Table~\ref{tab:steer}, Figure~\ref{fig:steer}).
InternVL3.5-38B spikes on none of the 150 clean images, so its cells rest on 2 to 11 images and read as creation rather than relocation.
A block of $25$ tokens, a tenth of the image, takes the spike on $47$ and $38\%$ in the two strongest models against a $9.8\%$ null, and one scaled token still takes it on $5\%$ against $0.4\%$ (Table~\ref{tab:blocks}).
Where the block takes the spike, its place can be chosen. Moved to the centre of each quadrant, the image's centre or a random position, the same block takes the spike on $9$ to $68\%$ of spiking images in the three Qwen3-VL models and InternVL3.5-8B, against $10\%$ at random, and on at most $22\%$ in InternVL3.5-4B and InternVL3.5-14B (Table~\ref{tab:positions}, Figure~\ref{fig:positions}). In Qwen2.5-VL-32B and Pixtral-12B, whose spike the scaled block does not move, it takes the spike at no position.
Which edit works, by contrast, is not free.
The direction locates Qwen2.5-VL-32B's spike at rest, at an AUC of $0.991$ with an FPR@95 of $3\%$ against the norm's $0.590$ and $57\%$ (Table~\ref{tab:locator}), and it is the model the fixed-norm edit moves.
Scaling is the edit that works in the other seven, and the norm is the better locator at rest in five of them.

\begin{table}[!htbp]
\caption{\textbf{The relocation interventions.} Per model, 150 images and one quadrant per image, the share of spiking images whose spike token lies in that quadrant, with the number of spiking images in parentheses.
The quadrant is the one whose tokens have the largest median norm on the clean image, so the clean share lies under the $25\%$ of a uniform landing.
Columns: the clean forward, then the quadrant's shared part cut to a tenth and quadrupled, the norm restored after each, and its tokens scaled to a tenth.
Read under $\Psi$ with the floor, tokens entering at $100$ or above excluded.\protect\backto{tab:steer}}
\label{tab:steer}
\begin{center}
\footnotesize
\setlength{\tabcolsep}{5pt}
\begin{tabular}{lrrrr}
\toprule
 & & $p$ down & $p$ up & scaled \\
model & clean & norm held & norm held & to a tenth \\
\midrule
Qwen3-VL-4B & 0.17 (116) & 0.28 (125) & 0.15 (115) & 0.56 (140) \\
Qwen3-VL-8B & 0.11 (9) & 0.30 (10) & 0.40 (10) & 0.79 (28) \\
Qwen3-VL-32B & 0.11 (57) & 0.10 (58) & 0.35 (71) & 0.40 (67) \\
Qwen2.5-VL-32B & 0.11 (72) & 0.68 (109) & 0.00 (64) & 0.00 (64) \\
InternVL3.5-4B & 0.19 (27) & 0.29 (31) & 0.05 (22) & 0.35 (23) \\
InternVL3.5-8B & 0.11 (82) & 0.28 (87) & 0.11 (92) & 0.72 (136) \\
InternVL3.5-14B & 0.14 (87) & 0.21 (94) & 0.14 (85) & 0.33 (106) \\
InternVL3.5-38B & -- (0) & 0.75 (4) & 0.00 (2) & 0.82 (11) \\
Pixtral-12B & 0.09 (150) & 0.09 (150) & 0.09 (150) & 0.09 (150) \\
\bottomrule
\end{tabular}
\end{center}
\end{table}

\begin{table}[!htbp]
\caption{\textbf{The spike comes to the block in proportion to its size.} Per model, 150 images, a $k \times k$ block around the centre of the target quadrant scaled to a tenth at the language model's input, and the share of spiking images whose spike lands inside it.
Columns: the block's size in tokens.
The last row is the share a token landing at random gives on the $16\times16$ grid of seven of the models.
The grid is $18\times18$ in Qwen2.5-VL-32B and $32\times32$ in Pixtral-12B, where the largest block covers $20\%$ and $6\%$ of the tokens.
$^\dagger$At most eleven spiking images per size and none at one token, indicative.
Read under $\Psi$ with the floor, tokens entering at $100$ or above excluded.\protect\backto{tab:blocks}}
\label{tab:blocks}
\begin{center}
\footnotesize
\setlength{\tabcolsep}{5pt}
\begin{tabular}{lrrrrrr}
\toprule
model & 1 & 9 & 16 & 25 & 36 & 64 \\
\midrule
Qwen3-VL-4B & 0.03 & 0.08 & 0.32 & 0.38 & 0.44 & 0.56 \\
Qwen3-VL-8B & 0.00 & 0.13 & 0.18 & 0.33 & 0.43 & 0.79 \\
Qwen3-VL-32B & 0.00 & 0.02 & 0.11 & 0.14 & 0.20 & 0.40 \\
Qwen2.5-VL-32B & 0.00 & 0.00 & 0.00 & 0.00 & 0.00 & 0.00 \\
InternVL3.5-4B & 0.00 & 0.00 & 0.03 & 0.18 & 0.20 & 0.35 \\
InternVL3.5-8B & 0.05 & 0.22 & 0.41 & 0.47 & 0.61 & 0.72 \\
InternVL3.5-14B & 0.00 & 0.02 & 0.05 & 0.09 & 0.11 & 0.33 \\
InternVL3.5-38B$^\dagger$ & -- & 0.00 & 0.33 & 0.67 & 0.00 & 0.82 \\
Pixtral-12B & 0.00 & 0.00 & 0.00 & 0.00 & 0.00 & 0.00 \\
\midrule
a token at random & 0.004 & 0.035 & 0.063 & 0.098 & 0.141 & 0.250 \\
\bottomrule
\end{tabular}
\end{center}
\end{table}

\begin{table}[!htbp]
\caption{\textbf{Where the block moves the spike, its place can be chosen.} Per model, 150 images, a $5\times5$ block of image tokens scaled to a tenth at the language model's input and placed at the centre of each quadrant, the grid's centre or a random position: the share of spiking forwards whose spike token lies inside the block, with their number in parentheses, and the share a token landing at random gives. Read under $\Psi$ with the floor, tokens entering at $100$ or above excluded. Molmo2-8B has no row, its overlapping crops having no single token grid.\protect\backto{tab:positions}}
\label{tab:positions}
\begin{center}
\footnotesize
\setlength{\tabcolsep}{3pt}
\resizebox{\linewidth}{!}{\begin{tabular}{llrrrrrrr}
\toprule
model & grid & top left & top right & bottom left & bottom right & centre & random & at random \\
\midrule
Qwen3-VL-4B & $16 \times 16$ & 0.64 (139) & 0.60 (136) & 0.18 (120) & 0.16 (118) & 0.17 (121) & 0.33 (125) & 0.098 \\
Qwen3-VL-8B & $16 \times 16$ & 0.68 (22) & 0.67 (27) & 0.20 (10) & 0.25 (12) & 0.44 (16) & 0.21 (14) & 0.098 \\
Qwen3-VL-32B & $16 \times 16$ & 0.46 (65) & 0.19 (58) & 0.09 (55) & 0.16 (62) & 0.28 (67) & 0.22 (58) & 0.098 \\
Qwen2.5-VL-32B & $18 \times 18$ & 0.00 (72) & 0.00 (72) & 0.00 (72) & 0.00 (72) & 0.00 (72) & 0.00 (72) & 0.077 \\
InternVL3.5-4B & $16 \times 16$ & 0.00 (26) & 0.07 (28) & 0.22 (36) & 0.21 (34) & 0.00 (26) & 0.08 (26) & 0.098 \\
InternVL3.5-8B & $16 \times 16$ & 0.68 (127) & 0.66 (122) & 0.34 (99) & 0.40 (107) & 0.39 (99) & 0.50 (115) & 0.098 \\
InternVL3.5-14B & $16 \times 16$ & 0.09 (100) & 0.14 (94) & 0.11 (92) & 0.11 (99) & 0.00 (87) & 0.12 (101) & 0.098 \\
InternVL3.5-38B & $16 \times 16$ & 0.00 (1) & 0.33 (3) & 1.00 (7) & -- (0) & 0.00 (1) & 0.50 (2) & 0.098 \\
Pixtral-12B & $32 \times 32$ & 0.00 (150) & 0.00 (150) & 0.00 (150) & 0.00 (150) & 0.00 (150) & 0.01 (150) & 0.024 \\
\bottomrule
\end{tabular}}
\end{center}
\end{table}

\FloatBarrier
\section{Visual spikes are brittle to image perturbations}
\label{app:psink}
\backto{app:psink}
This appendix gives the additional analyses for Section~\ref{sec:brittleness}.

\FloatBarrier
\subsection{Common corruptions}
\label{app:corruptions}
\backto{app:corruptions}
\textbf{The corruption panel.}\quad Table~\ref{tab:corrpanel} gives six operators at one grade each on the POPE manifest's 200 questions, read as Table~\ref{tab:noiseroster} is.
Every operator creates the state on 0 to 149 questions that carried none and abolishes it on 0 to 43 that did, $3{,}152$ creations against $476$ abolitions over the ten models and six operators.
Four of the 60 model--operator cells run the other way, two of them on one image, and 50 of the 60 move at least one question.
Incidence alone can hide that traffic.
Under a brightness rise, Qwen3-VL-32B moves from $0.41$ to $0.39$ while 39 questions gain the state and 43 lose it, so the share is steady and two fifths of the questions have changed sides.
Under the same six operators and 200 questions, none of the $16{,}800$ corrupted forwards of fourteen non-spiking models spikes, and the highest peak is $\Psi = 899$.
Among the forwards that keep the state, the spike sits on a different image token on $85$ to $100\%$ of them in Qwen3-VL-4B, Qwen3-VL-8B and Qwen3-VL-32B, and on $6$ to $14\%$ in Qwen2.5-VL-32B, which puts it back where it was.
Accuracy falls most under Gaussian noise, by $8.5$ to $22.0$ points, and least under JPEG compression and a brightness rise, by at most $5.5$ points.

\begin{figure}[!htbp]
\begin{center}
\includegraphics[width=\linewidth]{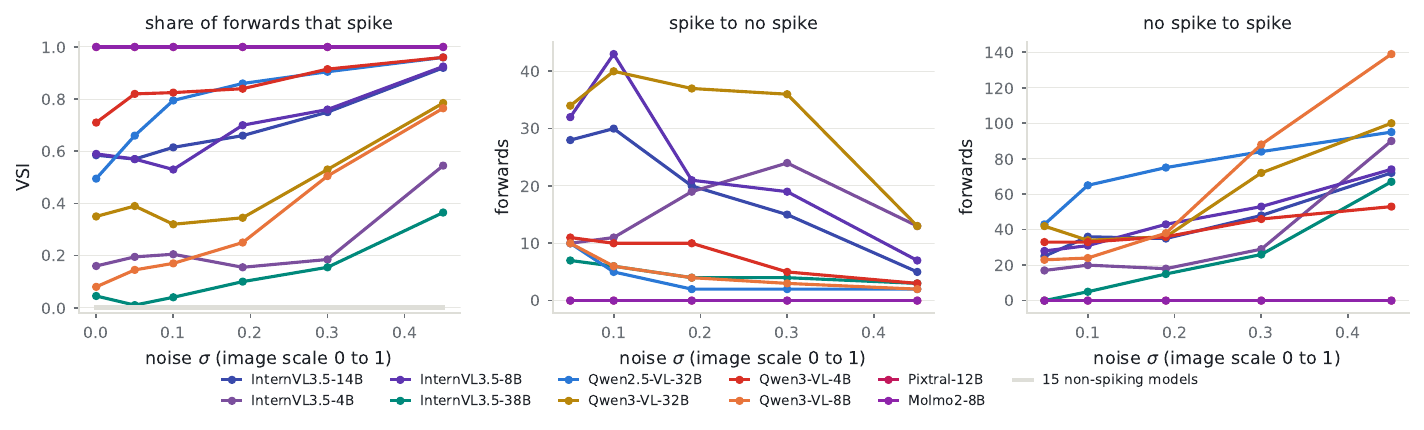}
\end{center}
\caption{\textbf{The corruption sweep on the same 200 forwards per model.} Left, the share of forwards that spike, clean and at the five noise grades.
Middle, the forwards that spike clean and not under the noise.
Right, the forwards that spike under the noise and not clean.
Eight spiking models rise overall with the grade, Molmo2-8B and Pixtral-12B sit at every forward, and the fifteen non-spiking models at none.
Table~\ref{tab:noiseroster} carries the numbers.\protect\backto{fig:noisesweep}}
\label{fig:noisesweep}
\end{figure}

\begin{table}[!htbp]
\caption{\textbf{Destroying the image raises the incidence in every spiking LVLM below saturation and creates it in none of the others.} Per LVLM, share of the same 200 POPE forwards that spike, clean and under Gaussian noise of standard deviation $\sigma$ on the $0$ to $1$ image scale, then POPE accuracy clean and at $\sigma = 0.45$.
Read on the 200 POPE questions of the evaluation manifest, not the 300-image census, so clean shares differ from Table~\ref{tab:fullroster} by a few points.
The six models of \aref{app:extension} close each group. The separate run of Table~\ref{tab:corrpanel} has clean shares within six points of these.\protect\backto{tab:noiseroster}}
\label{tab:noiseroster}
\begin{center}\scriptsize
\setlength{\tabcolsep}{3.5pt}
\begin{tabular}{llllllll}
\toprule
 & \multicolumn{6}{c}{spikes} & \\
\cmidrule(lr){2-7}
LVLM & clean & $\sigma\,0.05$ & $0.10$ & $0.19$ & $0.30$ & $0.45$ & accuracy, clean to $0.45$ \\
\midrule
Qwen3-VL-4B & 71.0\% & 82.0\% & 82.5\% & 84.0\% & 91.5\% & 96.0\% & 90.5\% to 79.0\% \\
Qwen3-VL-8B & 8.0\% & 14.5\% & 17.0\% & 25.0\% & 50.5\% & 76.5\% & 89.0\% to 78.0\% \\
Qwen3-VL-32B & 35.0\% & 39.0\% & 32.0\% & 34.5\% & 53.0\% & 78.5\% & 89.5\% to 78.0\% \\
Qwen2.5-VL-32B & 49.5\% & 66.0\% & 79.5\% & 86.0\% & 90.5\% & 96.0\% & 87.0\% to 69.5\% \\
InternVL3.5-4B & 16.0\% & 19.5\% & 20.5\% & 15.5\% & 18.5\% & 54.5\% & 84.0\% to 79.0\% \\
InternVL3.5-8B & 59.0\% & 57.0\% & 53.0\% & 70.0\% & 76.0\% & 92.5\% & 86.0\% to 79.0\% \\
InternVL3.5-14B & 58.5\% & 57.0\% & 61.5\% & 66.0\% & 75.0\% & 92.0\% & 86.5\% to 72.0\% \\
InternVL3.5-38B & 4.5\% & 1.0\% & 4.0\% & 10.0\% & 15.5\% & 36.5\% & 84.5\% to 78.0\% \\
\addlinespace[2pt]
Molmo2-8B & 100.0\% & 100.0\% & 100.0\% & 100.0\% & 100.0\% & 100.0\% & 90.0\% to 79.0\% \\
Pixtral-12B & 100.0\% & 100.0\% & 100.0\% & 100.0\% & 100.0\% & 100.0\% & 78.5\% to 71.5\% \\
\midrule
Qwen3-VL-2B & 0.0\% & 0.0\% & 0.0\% & 0.0\% & 0.0\% & 0.0\% & 92.0\% to 82.0\% \\
Qwen2.5-VL-3B & 0.0\% & 0.0\% & 0.0\% & 0.0\% & 0.0\% & 0.0\% & 87.5\% to 71.0\% \\
Qwen2.5-VL-7B & 0.0\% & 0.0\% & 0.0\% & 0.0\% & 0.0\% & 0.0\% & 87.5\% to 73.0\% \\
Qwen2.5-VL-72B & 0.0\% & 0.0\% & 0.0\% & 0.0\% & 0.0\% & 0.0\% & 86.5\% to 69.5\% \\
Qwen2-VL-7B & 0.0\% & 0.0\% & 0.0\% & 0.0\% & 0.0\% & 0.0\% & 89.0\% to 79.0\% \\
Qwen2.5-Omni-7B & 0.0\% & 0.0\% & 0.0\% & 0.0\% & 0.0\% & 0.0\% & 88.0\% to 72.0\% \\
InternVL2-8B & 0.0\% & 0.0\% & 0.0\% & 0.0\% & 0.0\% & 0.0\% & 84.5\% to 72.5\% \\
LLaVA-OV-7B & 0.0\% & 0.0\% & 0.0\% & 0.0\% & 0.0\% & 0.0\% & 92.5\% to 79.5\% \\
LLaVA-1.5-7B & 0.0\% & 0.0\% & 0.0\% & 0.0\% & 0.0\% & 0.0\% & 87.0\% to 80.5\% \\
LLaVA-1.5-13B & 0.0\% & 0.0\% & 0.0\% & 0.0\% & 0.0\% & 0.0\% & 86.0\% to 81.0\% \\
Molmo-7B-D & 0.0\% & 0.0\% & 0.0\% & 0.0\% & 0.0\% & 0.0\% & 90.0\% to 79.5\% \\
\addlinespace[2pt]
GLM-4.1V-9B-Base & 0.0\% & 0.0\% & 0.0\% & 0.0\% & 0.0\% & 0.0\% & 89.5\% to 81.0\% \\
Janus-Pro-7B & 0.0\% & 0.0\% & 0.0\% & 0.0\% & 0.0\% & 0.0\% & 88.0\% to 77.5\% \\
PaliGemma 2-3B-mix-448 & 0.0\% & 0.0\% & 0.0\% & 0.0\% & 0.0\% & 0.0\% & 91.0\% to 76.0\% \\
Idefics3-8B & 0.0\% & 0.0\% & 0.0\% & 0.0\% & 0.0\% & 0.0\% & 87.5\% to 71.0\% \\
\bottomrule
\end{tabular}
\end{center}
\end{table}

\begin{table}[!htbp]
\caption{\textbf{Six kinds of corruption all switch the state.} Per model and operator, the POPE manifest's 200 questions at one grade each, read as in Table~\ref{tab:noiseroster} on a separate run.
Columns: the share of forwards that spike clean and corrupted, the number of questions that gain and lose the state, the share of the forwards spiking in both whose spike sits on a different image token, the share whose yes-or-no decision is unchanged, and accuracy clean and corrupted.\protect\backto{tab:corrpanel}}
\label{tab:corrpanel}
\begin{center}
\footnotesize
\setlength{\tabcolsep}{4.0pt}\renewcommand{\arraystretch}{0.95}
\resizebox{\linewidth}{!}{\begin{tabular}{llrrrrrrrr}
\toprule
 & & \multicolumn{2}{c}{spikes} & \multicolumn{2}{c}{questions} & spike & answer & \multicolumn{2}{c}{accuracy} \\
model & operator & clean & corrupted & gained & lost & moved & unchanged & clean & corrupted \\
\midrule
Qwen3-VL-4B & Gaussian noise, $\sigma\,0.45$ & 0.68 & 0.97 & 61 & 3 & 0.99 & 0.83 & 0.915 & 0.795 \\
 & shot noise & 0.68 & 0.96 & 57 & 2 & 0.97 & 0.87 & 0.915 & 0.810 \\
 & Gaussian blur & 0.68 & 0.97 & 60 & 3 & 0.98 & 0.91 & 0.915 & 0.855 \\
 & JPEG & 0.68 & 0.83 & 41 & 12 & 0.91 & 0.93 & 0.915 & 0.870 \\
 & brightness & 0.68 & 0.79 & 37 & 16 & 0.85 & 0.96 & 0.915 & 0.895 \\
 & contrast & 0.68 & 0.95 & 56 & 3 & 0.90 & 0.95 & 0.915 & 0.870 \\
Qwen3-VL-8B & Gaussian noise, $\sigma\,0.45$ & 0.08 & 0.82 & 149 & 2 & 0.86 & 0.89 & 0.885 & 0.790 \\
 & shot noise & 0.08 & 0.60 & 105 & 2 & 1.00 & 0.90 & 0.885 & 0.840 \\
 & Gaussian blur & 0.08 & 0.54 & 94 & 3 & 1.00 & 0.96 & 0.885 & 0.880 \\
 & JPEG & 0.08 & 0.27 & 45 & 7 & 0.89 & 0.94 & 0.885 & 0.845 \\
 & brightness & 0.08 & 0.16 & 24 & 9 & 1.00 & 0.95 & 0.885 & 0.855 \\
 & contrast & 0.08 & 0.50 & 88 & 4 & 1.00 & 0.96 & 0.885 & 0.880 \\
Qwen3-VL-32B & Gaussian noise, $\sigma\,0.45$ & 0.41 & 0.75 & 83 & 14 & 0.93 & 0.82 & 0.895 & 0.785 \\
 & shot noise & 0.41 & 0.60 & 71 & 33 & 1.00 & 0.86 & 0.895 & 0.815 \\
 & Gaussian blur & 0.41 & 0.67 & 78 & 25 & 0.98 & 0.94 & 0.895 & 0.880 \\
 & JPEG & 0.41 & 0.56 & 59 & 29 & 0.94 & 0.94 & 0.895 & 0.880 \\
 & brightness & 0.41 & 0.39 & 39 & 43 & 0.92 & 0.95 & 0.895 & 0.870 \\
 & contrast & 0.41 & 0.69 & 81 & 24 & 0.98 & 0.95 & 0.895 & 0.885 \\
Qwen2.5-VL-32B & Gaussian noise, $\sigma\,0.45$ & 0.50 & 0.96 & 94 & 2 & 0.14 & 0.74 & 0.870 & 0.705 \\
 & shot noise & 0.50 & 0.91 & 84 & 2 & 0.13 & 0.85 & 0.870 & 0.795 \\
 & Gaussian blur & 0.50 & 1.00 & 100 & 0 & 0.09 & 0.90 & 0.870 & 0.790 \\
 & JPEG & 0.50 & 0.75 & 70 & 19 & 0.09 & 0.93 & 0.870 & 0.850 \\
 & brightness & 0.50 & 0.60 & 40 & 20 & 0.06 & 0.95 & 0.870 & 0.845 \\
 & contrast & 0.50 & 0.98 & 97 & 0 & 0.08 & 0.91 & 0.870 & 0.810 \\
InternVL3.5-4B & Gaussian noise, $\sigma\,0.45$ & 0.16 & 0.87 & 145 & 4 & 0.93 & 0.79 & 0.845 & 0.705 \\
 & shot noise & 0.16 & 0.61 & 99 & 10 & 0.91 & 0.82 & 0.845 & 0.765 \\
 & Gaussian blur & 0.16 & 0.13 & 16 & 23 & 0.56 & 0.88 & 0.845 & 0.855 \\
 & JPEG & 0.16 & 0.22 & 25 & 13 & 0.63 & 0.91 & 0.845 & 0.865 \\
 & brightness & 0.16 & 0.21 & 21 & 11 & 0.43 & 0.93 & 0.845 & 0.870 \\
 & contrast & 0.16 & 0.39 & 52 & 6 & 0.42 & 0.88 & 0.845 & 0.870 \\
InternVL3.5-8B & Gaussian noise, $\sigma\,0.45$ & 0.60 & 0.98 & 76 & 0 & 0.98 & 0.76 & 0.855 & 0.720 \\
 & shot noise & 0.60 & 0.95 & 71 & 2 & 0.98 & 0.81 & 0.855 & 0.750 \\
 & Gaussian blur & 0.60 & 0.87 & 66 & 12 & 0.95 & 0.87 & 0.855 & 0.825 \\
 & JPEG & 0.60 & 0.84 & 58 & 10 & 0.94 & 0.88 & 0.855 & 0.840 \\
 & brightness & 0.60 & 0.67 & 40 & 26 & 0.84 & 0.96 & 0.855 & 0.850 \\
 & contrast & 0.60 & 0.86 & 60 & 9 & 0.92 & 0.90 & 0.855 & 0.880 \\
InternVL3.5-14B & Gaussian noise, $\sigma\,0.45$ & 0.59 & 0.96 & 75 & 1 & 0.93 & 0.75 & 0.860 & 0.640 \\
 & shot noise & 0.59 & 0.90 & 66 & 4 & 0.92 & 0.77 & 0.860 & 0.710 \\
 & Gaussian blur & 0.59 & 0.93 & 70 & 3 & 0.82 & 0.86 & 0.860 & 0.790 \\
 & JPEG & 0.59 & 0.68 & 38 & 21 & 0.75 & 0.88 & 0.860 & 0.805 \\
 & brightness & 0.59 & 0.66 & 29 & 16 & 0.63 & 0.92 & 0.860 & 0.825 \\
 & contrast & 0.59 & 0.81 & 53 & 9 & 0.76 & 0.88 & 0.860 & 0.835 \\
InternVL3.5-38B & Gaussian noise, $\sigma\,0.45$ & 0.04 & 0.61 & 116 & 1 & 1.00 & 0.77 & 0.850 & 0.700 \\
 & shot noise & 0.04 & 0.43 & 81 & 3 & 1.00 & 0.81 & 0.850 & 0.780 \\
 & Gaussian blur & 0.04 & 0.59 & 110 & 0 & 0.57 & 0.92 & 0.850 & 0.850 \\
 & JPEG & 0.04 & 0.18 & 32 & 3 & 1.00 & 0.90 & 0.850 & 0.835 \\
 & brightness & 0.04 & 0.04 & 7 & 6 & 1.00 & 0.96 & 0.850 & 0.835 \\
 & contrast & 0.04 & 0.18 & 33 & 4 & 0.67 & 0.93 & 0.850 & 0.865 \\
Molmo2-8B & Gaussian noise, $\sigma\,0.45$ & 1.00 & 1.00 & 0 & 0 & 0.99 & 0.83 & 0.900 & 0.790 \\
 & shot noise & 1.00 & 1.00 & 0 & 0 & 1.00 & 0.87 & 0.900 & 0.805 \\
 & Gaussian blur & 1.00 & 1.00 & 0 & 0 & 0.96 & 0.95 & 0.900 & 0.885 \\
 & JPEG & 1.00 & 1.00 & 0 & 0 & 0.99 & 0.95 & 0.900 & 0.880 \\
 & brightness & 1.00 & 1.00 & 0 & 0 & 0.97 & 0.97 & 0.900 & 0.895 \\
 & contrast & 1.00 & 1.00 & 0 & 0 & 0.99 & 0.94 & 0.900 & 0.875 \\
Pixtral-12B & Gaussian noise, $\sigma\,0.45$ & 1.00 & 1.00 & 0 & 1 & 0.20 & 0.79 & 0.785 & 0.700 \\
 & shot noise & 1.00 & 1.00 & 0 & 0 & 0.14 & 0.84 & 0.785 & 0.760 \\
 & Gaussian blur & 1.00 & 1.00 & 0 & 0 & 0.02 & 0.85 & 0.785 & 0.750 \\
 & JPEG & 1.00 & 1.00 & 0 & 0 & 0.04 & 0.88 & 0.785 & 0.830 \\
 & brightness & 1.00 & 1.00 & 0 & 1 & 0.04 & 0.92 & 0.785 & 0.785 \\
 & contrast & 1.00 & 1.00 & 0 & 0 & 0.02 & 0.89 & 0.785 & 0.795 \\
\bottomrule
\end{tabular}}
\end{center}
\end{table}

\FloatBarrier
\subsection{The spike attack}
\label{app:attack}
\backto*{app:attack}
\textbf{The objective.}\quad Equation~\ref{eq:attack} uses two measurements at the visual switch.

First, the alignment $c_i(\vb{x})=\cos(\vb{x}_i^{(\ell^{\ast})},\vb{t})$
measures how closely image token $i$'s residual at the input of block $\ell^{\ast}$ aligns with the trigger direction. It provides a gradient before the switch has written a spike. Second, the value $v_i(\vb{x})=s_d\,\vb{x}_i^{(\ell^{\ast}+1)}[d]$
measures the token's signed activation on spike channel $d$ at the switch output. The model's write sign $s_d\in\{-1,+1\}$ makes a massive activation score high regardless of its sign. This term provides a gradient once the spike is being written. The sign outside the brackets in Equation~\ref{eq:attack} instead sets the attack's direction.

We cap the value term at $2{,}000$ so that an already massive token does not dominate the objective. Scaling it by $1/1000$ puts a value of $1{,}000$ on the scale of a cosine of $1$.

\textbf{From the maximum to the log-sum-exp.}\quad What the attack wants is the most aligned token and the largest value, $\max_i c_i$ and $\max_i v_i$.
A maximum over tokens passes its gradient to one token only, so Equation~\ref{eq:attack} replaces each maximum by a soft maximum, $\tau \operatorname{logsumexp}_{i \in I} (a_i / \tau) = \tau \log \sum_{i \in I} e^{a_i / \tau}$, which lies between $\max_i a_i$ and $\max_i a_i + \tau \log \lvert I \rvert$ and passes the gradient to every token in proportion to $e^{a_i / \tau}$.
The alignment term uses $\tau_1 = 0.02$ and the value term $\tau_2 = 50$, so on 256 tokens each soft maximum sits within $0.11$ and $277$ of its maximum, and both temperatures sit below the scale of what they smooth, a cosine of $0.3$ at the switch's input and a value in the thousands at its output.

\textbf{The steps and the stop.}\quad Starting from $\vb*{\delta} = 0$, each step moves $\vb*{\delta}$ by $\alpha\, \mathrm{sign}(\nabla_{\vb*{\delta}})$ with $\alpha = \nicefrac{\varepsilon}{4}$, up the objective to create and down it to abolish, then projects $\vb*{\delta}$ onto the $L_\infty$ ball of radius $\varepsilon$ and $\vb{x} + \vb*{\delta}$ onto the pixel box, for at most 300 steps on the Qwen and InternVL3.5 models.
The forward at each step gives the reading.
Creation succeeds at the first step at which some image token's value at the switch's output reaches $100$ and $1{,}000$ times the state's median, and abolition at the first step at which no image token meets the criterion in any deep state.
That step is kept and the answer is read from it.
Nothing about the output enters the objective, the step or the choice of step.

\textbf{Budget and steps.}\quad Budgets are counted in pixel levels from one upward, and the two sides that stay far below saturation at two levels, 
InternVL3.5-38B's creating side and Molmo2-8B's abolishing side, are followed further.
The four Qwen models and InternVL3.5-4B, 8B and 14B are run at $\varepsilon$ of one and two pixel levels, with a signed step of $\nicefrac{\varepsilon}{4}$ and at most 300 steps.
InternVL3.5-38B's creating side does not saturate and is followed to $\nicefrac{64}{255}$ with a step of $\nicefrac{0.25}{255}$.
Each of its runs is one search from the clean image, and a run at a larger budget can miss an image that a smaller one found, since success comes as a jump after a flat stretch and the path differs once a pixel meets the smaller budget.
A success at a smaller budget lies within every larger one. Each budget's cell therefore counts an image once a run succeeds at that budget or a smaller one, and keeps that perturbation.
Read this way, it reaches $53\%$ of its 40 images at one pixel level, $58\%$ at two, $70\%$ at four, $80\%$ at eight and $85\%$ at $\nicefrac{64}{255}$, and 6 of the 40 images spike at no budget.
Molmo2-8B and Pixtral-12B are run under two schedules, 60 steps of $\nicefrac{\varepsilon}{4}$ and 600 steps of $\nicefrac{0.25}{255}$, at the budgets of Table~\ref{tab:epssweep}.
The free-form runs of Table~\ref{tab:attackfreeform} are at one pixel level for the eight Qwen and InternVL3.5 models, with a step of $\nicefrac{0.25}{255}$.
Pixtral-12B's run caps the budget at $\varepsilon = \nicefrac{4}{255}$ with a step of $\nicefrac{1}{255}$ and stops at the first step whose criterion flips, and $580$ of its $595$ successful attacks carry four pixel levels.

\begin{figure}[!htbp]
\begin{center}
\includegraphics[width=\linewidth]{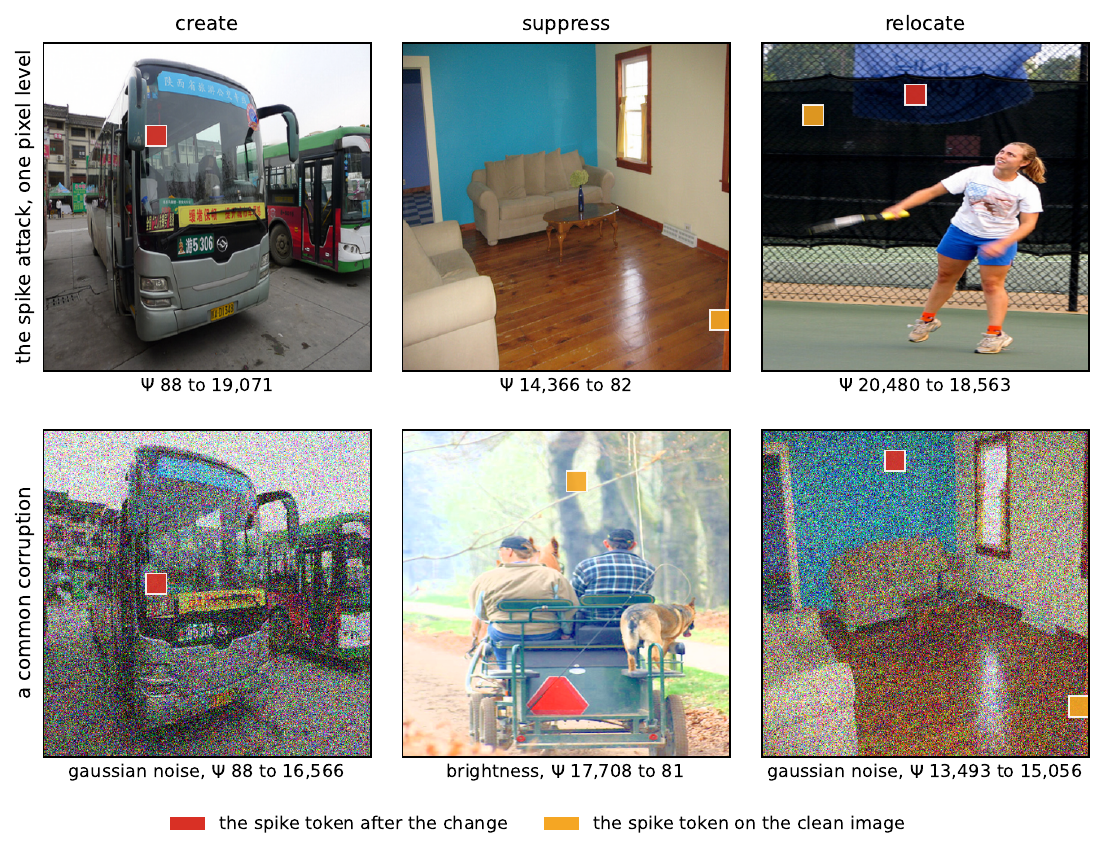}
\end{center}
\caption{\textbf{Two causes, three effects.} Qwen3-VL-4B.
Top row, the trigger-guided attack at one pixel level, which the page cannot show.
Bottom row, a common corruption at the grade of Table~\ref{tab:corrpanel}.
Columns: the state created where the clean image had none, suppressed where it stood, and relocated to another token.
The red square is the token carrying the spike after the change, and the amber square the token that carried it on the clean image.
Under each panel, the criterion's peak before and after.\protect\backto{fig:cimg}}
\label{fig:cimg}
\end{figure}

\begin{table}[!htbp]
\caption{\textbf{One pixel level over the image already switches the massive state.} Per LVLM and budget, 40 POPE images a direction.
\emph{Create} and \emph{abolish} are success rates over the eligible images in parentheses, \emph{no change} the share of answers that hold, then median KL and $L_\infty$.
Up to 300 steps of $\nicefrac{\varepsilon}{4}$, of $\nicefrac{0.25}{255}$ for InternVL3.5-38B, whose cells count each image at the smallest budget where a run succeeds.
Molmo2-8B and Pixtral-12B are read on the tensor, $\ast$ after 60 steps and unstarred 600, others on the saved image.\protect\backto{tab:epssweep}}
\label{tab:epssweep}
\begin{center}\small
\setlength{\tabcolsep}{4pt}
\begin{tabular}{lllllll}
\toprule
LVLM & budget & create & abolish & no change & median KL & median $L_\infty$ \\
\midrule
Qwen3-VL-4B & $\nicefrac{1}{255}$ & 100\% (40) & 95\% (40) & 99\% (80) & 1e{-}6 & 1 / 1 \\
Qwen3-VL-4B & $\nicefrac{2}{255}$ & 100\% (40) & 93\% (40) & 100\% (80) & 3e{-}6 & 1 / 2 \\
Qwen3-VL-8B & $\nicefrac{1}{255}$ & 95\% (40) & 100\% (16) & 93\% (56) & 3e{-}4 & 1 / 1 \\
Qwen3-VL-8B & $\nicefrac{2}{255}$ & 93\% (40) & 100\% (16) & 95\% (56) & 3e{-}4 & 2 / 1 \\
Qwen3-VL-32B & $\nicefrac{1}{255}$ & 100\% (40) & 100\% (40) & 100\% (80) & 1e{-}2 & 1 / 1 \\
Qwen3-VL-32B & $\nicefrac{2}{255}$ & 100\% (40) & 100\% (40) & 99\% (80) & 1e{-}2 & 2 / 1 \\
Qwen2.5-VL-32B & $\nicefrac{1}{255}$ & 100\% (40) & 100\% (40) & 98\% (80) & 1e{-}4 & 1 / 1 \\
Qwen2.5-VL-32B & $\nicefrac{2}{255}$ & 100\% (40) & 100\% (40) & 100\% (80) & 3e{-}4 & 2 / 1 \\
InternVL3.5-4B & $\nicefrac{1}{255}$ & 80\% (40) & 100\% (27) & 100\% (67) & 3e{-}5 & 1 / 1 \\
InternVL3.5-4B & $\nicefrac{2}{255}$ & 85\% (40) & 100\% (27) & 99\% (67) & 6e{-}5 & 2 / 2 \\
InternVL3.5-8B & $\nicefrac{1}{255}$ & 100\% (40) & 98\% (40) & 100\% (80) & 5e{-}5 & 1 / 1 \\
InternVL3.5-8B & $\nicefrac{2}{255}$ & 100\% (40) & 100\% (40) & 100\% (80) & 2e{-}4 & 2 / 2 \\
InternVL3.5-14B & $\nicefrac{1}{255}$ & 100\% (40) & 95\% (40) & 100\% (80) & 2e{-}5 & 1 / 1 \\
InternVL3.5-14B & $\nicefrac{2}{255}$ & 100\% (40) & 98\% (40) & 100\% (80) & 3e{-}5 & 2 / 2 \\
Pixtral-12B$^{\ast}$ & $\nicefrac{1}{255}$ & -- & 80\% (40) & 100\% (40) & 2e{-}2 & -- / 1 \\
Pixtral-12B$^{\ast}$ & $\nicefrac{2}{255}$ & -- & 98\% (40) & 95\% (40) & 4e{-}2 & -- / 2 \\
Pixtral-12B & $\nicefrac{1}{255}$ & -- & 100\% (40) & 95\% (40) & 2e{-}2 & -- / 1 \\
\midrule
Molmo2-8B$^{\ast}$ & $\nicefrac{1}{255}$ & -- & 0\% (40) & 100\% (40) & 7e{-}5 & -- / -- \\
Molmo2-8B$^{\ast}$ & $\nicefrac{2}{255}$ & -- & 0\% (40) & 100\% (40) & 3e{-}4 & -- / -- \\
Molmo2-8B & $\nicefrac{2}{255}$ & -- & 35\% (40) & 98\% (40) & 3e{-}4 & -- / 2 \\
Molmo2-8B & $\nicefrac{4}{255}$ & -- & 88\% (40) & 93\% (40) & 9e{-}4 & -- / 4 \\
Molmo2-8B & $\nicefrac{8}{255}$ & -- & 98\% (40) & 98\% (40) & 2e{-}3 & -- / 8 \\
Molmo2-8B & $\nicefrac{16}{255}$ & -- & 100\% (20) & 95\% (20) & 5e{-}3 & -- / 16 \\
Molmo2-8B & $\nicefrac{64}{255}$ & -- & 95\% (20) & 95\% (20) & 3e{-}3 & -- / 17 \\
InternVL3.5-38B & $\nicefrac{1}{255}$ & 53\% (40) & 100\% (7) & 100\% (47) & 2e{-}5 & 1 / 1 \\
InternVL3.5-38B & $\nicefrac{2}{255}$ & 58\% (40) & -- & 100\% (40) & 9e{-}5 & 1 / -- \\
InternVL3.5-38B & $\nicefrac{4}{255}$ & 70\% (40) & -- & 100\% (40) & 3e{-}5 & 1 / -- \\
InternVL3.5-38B & $\nicefrac{8}{255}$ & 80\% (40) & -- & 100\% (40) & 2e{-}5 & 1 / -- \\
InternVL3.5-38B & $\nicefrac{64}{255}$ & 85\% (40) & -- & 100\% (40) & 2e{-}5 & 1 / -- \\
\bottomrule
\end{tabular}
\end{center}\end{table}

\begin{table}[!htbp]
\caption{\textbf{On the free-form tasks the attack costs the answer a few points.} The eight Qwen and InternVL3.5 models at one pixel level in both directions, and Pixtral-12B below the rule on the abolishing side only, under a cap of four pixel levels.
Each image is attacked once, with the VQAv2 prompt, and both tasks are scored on that image. Pixtral-12B's captions come from their own attack at the same budget.
Columns: the budget, the $n$ VQAv2 questions and captions attacked, the pool less the forwards whose clean side moved, then the clean and the attacked score.
Qwen3-VL-8B's abolishing side covers 23 of its 32 spiking questions.
Every attempt is scored. The attack succeeds on $90$ to $100\%$ of attempts in seven models and on $66$ of InternVL3.5-38B's $101$.
VQAv2 is the official accuracy, leave one annotator out, and captioning is corpus CIDEr-D against COCO's five references.\protect\backto{tab:attackfreeform}}
\label{tab:attackfreeform}
\begin{center}\small
\begin{tabular}{llrrrrr}
\toprule
 & & & \multicolumn{2}{c}{VQAv2 accuracy} & \multicolumn{2}{c}{captioning CIDEr} \\
\cmidrule(lr){4-5}\cmidrule(lr){6-7}
model & budget & $n$ & clean & attacked & clean & attacked \\
\midrule
Qwen3-VL-4B & $\nicefrac{1}{255}$ & 264, 264 & 72.0\% & 70.4\% & 0.191 & 0.184 \\
Qwen3-VL-8B & $\nicefrac{1}{255}$ & 277, 277 & 73.9\% & 70.9\% & 0.348 & 0.319 \\
Qwen2.5-VL-32B & $\nicefrac{1}{255}$ & 293, 293 & 72.6\% & 70.1\% & 0.232 & 0.200 \\
Qwen3-VL-32B & $\nicefrac{1}{255}$ & 258, 258 & 75.1\% & 75.7\% & 0.143 & 0.136 \\
InternVL3.5-4B & $\nicefrac{1}{255}$ & 151, 151 & 63.8\% & 58.9\% & 0.588 & 0.578 \\
InternVL3.5-8B & $\nicefrac{1}{255}$ & 166, 166 & 65.2\% & 62.3\% & 0.699 & 0.664 \\
InternVL3.5-14B & $\nicefrac{1}{255}$ & 182, 182 & 63.4\% & 62.2\% & 0.565 & 0.587 \\
InternVL3.5-38B & $\nicefrac{1}{255}$ & 101, 101 & 75.0\% & 76.3\% & 0.841 & 0.739 \\
\midrule
Pixtral-12B & $\nicefrac{4}{255}$ & 296, 299 & 61.6\% & 60.4\% & 0.123 & 0.234 \\
\bottomrule
\end{tabular}
\end{center}
\end{table}

\FloatBarrier
\section{Preventive intervention}
\label{app:prevent}
\backto*{app:prevent}
Tables~\ref{tab:prevent} and~\ref{tab:preventfree} give the per-model intervention results on clean POPE inputs and free-form tasks.
Tables~\ref{tab:preventcorr} and~\ref{tab:preventatk} give the results on corrupted inputs and images created by the spike attack.
Table~\ref{tab:baselines} in the main text compares the trigger with less specific edits.

\textbf{Conditional intervention.}\quad
A conditional variant applies Equation~\ref{eq:null} only to tokens whose alignment $\vb{t}^{\top}\bar{\vb{x}}_i/\lVert\bar{\vb{x}}_i\rVert$ exceeds a threshold chosen on held-out questions, one that leaves no more spiking questions than full removal.
Table~\ref{tab:prevent} reports the number of image tokens modified.

\begin{table}[!htbp]
\caption{\textbf{Removing the trigger component prevents the eruption.} Per model, 200 evaluation questions.
Columns: share spiking and share reaching the attention-sink threshold, each without and with the intervention, and again under Gaussian noise at $\sigma = 0.15$; clean accuracy, and the intervened one where it differs; spike rate under a random direction of the same rank; image tokens the conditional form modifies; median peak $\Psi$ clean and under it, over questions that spike clean.
Each model is run with the trigger of its own base.\protect\backto{tab:prevent}}
\label{tab:prevent}
\begin{center}
\footnotesize
\setlength{\tabcolsep}{3.4pt}
\resizebox{\linewidth}{!}{\begin{tabular}{lrrrrrrrrrl}
\toprule
 & \multicolumn{2}{c}{spikes} & \multicolumn{2}{c}{attention sink} & \multicolumn{2}{c}{noised spikes} & & random & tokens & peak $\Psi$ \\
model & clean & nulled & clean & nulled & clean & nulled & accuracy & direction & modified & clean to nulled \\
\midrule
Qwen3-VL-4B & 0.71 & 0.00 & 0.68 & 0.00 & 0.85 & 0.00 & 0.910 & 0.71 & 1 & 16{,}343 to 90 \\
Qwen3-VL-8B & 0.09 & 0.00 & 0.08 & 0.00 & 0.27 & 0.00 & 0.885 & 0.09 & 1 & 9{,}376 to 120 \\
Qwen3-VL-32B & 0.35 & 0.00 & 0.38 & 0.00 & 0.34 & 0.00 & 0.890 & 0.35 & 1 & 16{,}024 to 495 \\
Qwen2.5-VL-32B & 0.49 & 0.00 & 0.49 & 0.00 & 0.95 & 0.00 & 0.870 & 0.49 & 1 & 4{,}004 to 208 \\
InternVL3.5-4B & 0.16 & 0.03 & 0.14 & 0.00 & 0.14 & 0.03 & 0.845 & 0.16 & 2 & 10{,}372 to 208 \\
InternVL3.5-8B & 0.60 & 0.33 & 0.59 & 0.00 & 0.67 & 0.35 & 0.855 & 0.60 & 2 & 22{,}985 to 1{,}276 \\
InternVL3.5-14B & 0.61 & 0.17 & 0.60 & 0.00 & 0.66 & 0.15 & 0.860 & 0.61 & 2 & 13{,}882 to 454 \\
InternVL3.5-38B & 0.05 & 0.00 & 0.00 & 0.00 & 0.10 & 0.00 & 0.845 to 0.850 & 0.05 & 1 & 7{,}578 to 313 \\
Molmo2-8B & 1.00 & 0.46 & 0.25 & 0.00 & 1.00 & 0.39 & 0.900 & 1.00 & 9 & 22{,}105 to 874 \\
Pixtral-12B & 1.00 & 0.00 & 1.00 & 1.00 & 1.00 & 0.00 & 0.770 to 0.820 & 1.00 & 3 & 17{,}270 to 664 \\
\bottomrule
\end{tabular}}
\end{center}
\end{table}

\begin{table}[!htbp]
\caption{\textbf{The intervention holds on free-form answers.} Per model, 297 VQAv2 questions and 300 COCO captions, greedy decoding, with Equation~\ref{eq:null} applied to every image token.
Columns: the share of forwards that spike without and with the intervention on each task, VQA accuracy, corpus CIDEr against COCO's references, the share of VQAv2 answers whose text is identical to the original up to case, punctuation and articles, and the same share for captions under the intervention and under a random direction of the same rank.\protect\backto{tab:preventfree}}
\label{tab:preventfree}
\begin{center}
\footnotesize
\setlength{\tabcolsep}{3.8pt}
\resizebox{\linewidth}{!}{\begin{tabular}{lrrrrrrrrrrr}
\toprule
 & \multicolumn{2}{c}{spikes, VQAv2} & \multicolumn{2}{c}{spikes, captions} & \multicolumn{2}{c}{VQA accuracy} & \multicolumn{2}{c}{CIDEr} & answer & \multicolumn{2}{c}{caption unchanged} \\
model & clean & nulled & clean & nulled & clean & nulled & clean & nulled & unchanged & nulled & random \\
\midrule
Qwen3-VL-4B & 0.79 & 0.00 & 0.79 & 0.00 & 0.729 & 0.723 & 0.191 & 0.194 & 0.99 & 0.79 & 0.82 \\
Qwen3-VL-8B & 0.11 & 0.00 & 0.12 & 0.00 & 0.736 & 0.737 & 0.361 & 0.369 & 0.99 & 0.81 & 0.86 \\
Qwen3-VL-32B & 0.41 & 0.00 & 0.41 & 0.00 & 0.741 & 0.741 & 0.145 & 0.145 & 1.00 & 0.87 & 0.88 \\
Qwen2.5-VL-32B & 0.56 & 0.00 & 0.56 & 0.00 & 0.720 & 0.720 & 0.234 & 0.227 & 0.98 & 0.79 & 0.81 \\
InternVL3.5-4B & 0.16 & 0.04 & 0.16 & 0.04 & 0.655 & 0.651 & 0.588 & 0.598 & 0.97 & 0.86 & 0.89 \\
InternVL3.5-8B & 0.62 & 0.34 & 0.62 & 0.33 & 0.666 & 0.666 & 0.720 & 0.720 & 0.98 & 0.87 & 0.89 \\
InternVL3.5-14B & 0.68 & 0.20 & 0.68 & 0.19 & 0.672 & 0.674 & 0.595 & 0.594 & 0.99 & 0.85 & 0.88 \\
InternVL3.5-38B & 0.01 & 0.00 & 0.01 & 0.00 & 0.738 & 0.738 & 0.800 & 0.795 & 1.00 & 0.85 & 0.86 \\
Molmo2-8B & 1.00 & 0.39 & 1.00 & 0.40 & 0.683 & 0.680 & 0.208 & 0.216 & 0.99 & 0.53 & 0.88 \\
Pixtral-12B & 1.00 & 0.00 & 1.00 & 0.00 & 0.619 & 0.645 & 0.125 & 0.283 & 0.84 & 0.13 & 0.63 \\
\bottomrule
\end{tabular}}
\end{center}
\end{table}

\begin{table}[!htbp]
\caption{\textbf{The intervention holds under corruption.} Ten models, 50 evaluation questions each, the six operators of Table~\ref{tab:corrpanel} at the same grades.
Columns: the share of images that spike without and with the intervention, then accuracy without and with it.
Read under $\Psi$ with the floor, tokens entering at $100$ or above excluded.\protect\backto{tab:preventcorr}}
\label{tab:preventcorr}
\begin{center}
\scriptsize
\setlength{\tabcolsep}{3pt}
\begin{tabular}[t]{llrrrr}
\toprule
 & & \multicolumn{2}{c}{spikes} & \multicolumn{2}{c}{accuracy} \\
model & image & plain & interv. & plain & interv. \\
\midrule
Qwen3-VL-4B & clean & 0.58 & 0.00 & 0.86 & 0.86 \\
 & Gaussian noise & 0.98 & 0.00 & 0.80 & 0.80 \\
 & shot noise & 0.94 & 0.00 & 0.82 & 0.82 \\
 & Gaussian blur & 0.94 & 0.00 & 0.90 & 0.90 \\
 & JPEG & 0.76 & 0.00 & 0.84 & 0.84 \\
 & brightness & 0.78 & 0.00 & 0.84 & 0.84 \\
 & contrast & 0.98 & 0.00 & 0.84 & 0.84 \\
\midrule
Qwen3-VL-8B & clean & 0.12 & 0.00 & 0.84 & 0.84 \\
 & Gaussian noise & 0.88 & 0.00 & 0.76 & 0.78 \\
 & shot noise & 0.60 & 0.00 & 0.80 & 0.80 \\
 & Gaussian blur & 0.62 & 0.00 & 0.84 & 0.84 \\
 & JPEG & 0.40 & 0.00 & 0.82 & 0.82 \\
 & brightness & 0.16 & 0.00 & 0.80 & 0.80 \\
 & contrast & 0.52 & 0.00 & 0.82 & 0.82 \\
\midrule
Qwen3-VL-32B & clean & 0.46 & 0.00 & 0.84 & 0.84 \\
 & Gaussian noise & 0.72 & 0.00 & 0.80 & 0.80 \\
 & shot noise & 0.56 & 0.00 & 0.80 & 0.80 \\
 & Gaussian blur & 0.66 & 0.00 & 0.86 & 0.86 \\
 & JPEG & 0.54 & 0.00 & 0.84 & 0.84 \\
 & brightness & 0.46 & 0.00 & 0.86 & 0.86 \\
 & contrast & 0.68 & 0.00 & 0.84 & 0.84 \\
\midrule
Qwen2.5-VL-32B & clean & 0.52 & 0.00 & 0.86 & 0.86 \\
 & Gaussian noise & 1.00 & 0.00 & 0.80 & 0.80 \\
 & shot noise & 0.90 & 0.00 & 0.84 & 0.84 \\
 & Gaussian blur & 1.00 & 0.00 & 0.80 & 0.80 \\
 & JPEG & 0.88 & 0.00 & 0.88 & 0.88 \\
 & brightness & 0.62 & 0.00 & 0.84 & 0.84 \\
 & contrast & 1.00 & 0.00 & 0.84 & 0.86 \\
\midrule
InternVL3.5-4B & clean & 0.16 & 0.00 & 0.88 & 0.88 \\
 & Gaussian noise & 0.80 & 0.36 & 0.66 & 0.68 \\
 & shot noise & 0.52 & 0.10 & 0.78 & 0.78 \\
 & Gaussian blur & 0.20 & 0.04 & 0.86 & 0.88 \\
 & JPEG & 0.22 & 0.04 & 0.94 & 0.94 \\
 & brightness & 0.26 & 0.08 & 0.90 & 0.90 \\
 & contrast & 0.52 & 0.06 & 0.84 & 0.86 \\
\bottomrule
\end{tabular}\hfill
\begin{tabular}[t]{llrrrr}
\toprule
 & & \multicolumn{2}{c}{spikes} & \multicolumn{2}{c}{accuracy} \\
model & image & plain & interv. & plain & interv. \\
\midrule
InternVL3.5-8B & clean & 0.68 & 0.48 & 0.86 & 0.86 \\
 & Gaussian noise & 0.96 & 0.78 & 0.70 & 0.72 \\
 & shot noise & 0.92 & 0.58 & 0.74 & 0.74 \\
 & Gaussian blur & 0.88 & 0.72 & 0.82 & 0.80 \\
 & JPEG & 0.86 & 0.58 & 0.82 & 0.82 \\
 & brightness & 0.74 & 0.50 & 0.88 & 0.90 \\
 & contrast & 0.84 & 0.56 & 0.90 & 0.90 \\
\midrule
InternVL3.5-14B & clean & 0.62 & 0.18 & 0.90 & 0.90 \\
 & Gaussian noise & 0.98 & 0.18 & 0.70 & 0.70 \\
 & shot noise & 0.88 & 0.14 & 0.66 & 0.66 \\
 & Gaussian blur & 0.94 & 0.20 & 0.80 & 0.80 \\
 & JPEG & 0.70 & 0.18 & 0.86 & 0.86 \\
 & brightness & 0.78 & 0.20 & 0.82 & 0.82 \\
 & contrast & 0.78 & 0.12 & 0.84 & 0.84 \\
\midrule
InternVL3.5-38B & clean & 0.04 & 0.00 & 0.82 & 0.82 \\
 & Gaussian noise & 0.60 & 0.00 & 0.68 & 0.68 \\
 & shot noise & 0.54 & 0.00 & 0.78 & 0.78 \\
 & Gaussian blur & 0.56 & 0.02 & 0.80 & 0.80 \\
 & JPEG & 0.14 & 0.00 & 0.80 & 0.80 \\
 & brightness & 0.04 & 0.00 & 0.78 & 0.80 \\
 & contrast & 0.18 & 0.00 & 0.84 & 0.84 \\
\midrule
Pixtral-12B & clean & 1.00 & 0.00 & 0.80 & 0.78 \\
 & Gaussian noise & 0.98 & 0.00 & 0.70 & 0.70 \\
 & shot noise & 1.00 & 0.00 & 0.76 & 0.80 \\
 & Gaussian blur & 1.00 & 0.00 & 0.72 & 0.74 \\
 & JPEG & 1.00 & 0.00 & 0.82 & 0.80 \\
 & brightness & 1.00 & 0.00 & 0.78 & 0.82 \\
 & contrast & 1.00 & 0.00 & 0.78 & 0.82 \\
\midrule
Molmo2-8B & clean & 1.00 & 0.44 & 0.86 & 0.86 \\
 & Gaussian noise & 1.00 & 0.58 & 0.84 & 0.86 \\
 & shot noise & 1.00 & 0.52 & 0.82 & 0.84 \\
 & Gaussian blur & 1.00 & 0.54 & 0.86 & 0.86 \\
 & JPEG & 1.00 & 0.56 & 0.88 & 0.88 \\
 & brightness & 1.00 & 0.40 & 0.88 & 0.88 \\
 & contrast & 1.00 & 0.60 & 0.88 & 0.88 \\
\bottomrule
\end{tabular}
\end{center}
\end{table}

\begin{table}[!htbp]
\caption{\textbf{The intervention holds on the attack's own images.} Every image on which the creating attack succeeded at one pixel level, read from the 8-bit file its success was judged on. Molmo2-8B and Pixtral-12B spike on nearly every clean image, so the attack has no spike to create in them.
Columns: the images, the share that spike clean, under the attack, and under the attack with Equation~\ref{eq:null} in place, then the accuracy of each, then the largest criterion value over the images under the attack and under the intervention.
Read under $\Psi$ with the floor, tokens entering at $100$ or above excluded.\protect\backto{tab:preventatk}}
\label{tab:preventatk}
\begin{center}
\footnotesize
\setlength{\tabcolsep}{3.5pt}
\begin{tabular}{lrrrrrrrrr}
\toprule
 & & \multicolumn{3}{c}{spikes} & \multicolumn{3}{c}{accuracy} & \multicolumn{2}{c}{peak $\Psi$} \\
model & $n$ & clean & attacked & intervened & clean & attacked & intervened & attacked & intervened \\
\midrule
Qwen3-VL-4B & 40 & 0.00 & 0.95 & 0.00 & 0.93 & 0.90 & 0.90 & 19{,}626 & 100 \\
Qwen3-VL-8B & 40 & 0.03 & 0.83 & 0.00 & 0.88 & 0.78 & 0.78 & 23{,}121 & 156 \\
Qwen3-VL-32B & 40 & 0.10 & 0.93 & 0.00 & 0.80 & 0.80 & 0.80 & 21{,}749 & 691 \\
Qwen2.5-VL-32B & 40 & 0.10 & 1.00 & 0.00 & 0.88 & 0.88 & 0.88 & 5{,}542 & 304 \\
InternVL3.5-4B & 40 & 0.00 & 0.80 & 0.28 & 0.88 & 0.88 & 0.88 & 23{,}740 & 2{,}009 \\
InternVL3.5-8B & 40 & 0.00 & 1.00 & 0.53 & 0.88 & 0.88 & 0.88 & 29{,}747 & 3{,}206 \\
InternVL3.5-14B & 40 & 0.00 & 1.00 & 0.30 & 0.93 & 0.93 & 0.93 & 23{,}740 & 3{,}308 \\
InternVL3.5-38B & 40 & 0.00 & 0.53 & 0.00 & 0.80 & 0.80 & 0.80 & 21{,}124 & 479 \\
\bottomrule
\end{tabular}
\end{center}
\end{table}

\FloatBarrier
\section{Six models in detail}
\label{app:extension}
\backto{app:extension}
This appendix gives what is particular to six of the 25 models, five of them on bases no other model shares (Table~\ref{tab:fullroster}).
They are Molmo2-8B \citep{ai2026molmo2} on Qwen3-8B, Pixtral-12B \citep{agrawal2024pixtral} on Mistral-NeMo-12B \citep{mistral2024nemo}, GLM-4.1V-9B-Base \citep{glmv2025thinking} on GLM-4-9B \citep{glm2024chatglm}, Janus-Pro-7B \citep{chen2025janus} on DeepSeek-LLM-7B \citep{deepseek2024llm}, PaliGemma 2-3B-mix-448 \citep{steiner2024paligemma2} on Gemma-2-2B \citep{gemma2024gemma2}, and Idefics3-8B \citep{laurencon2024idefics3} on Llama-3.1-8B \citep{grattafiori2024llama3}.
The tables give their rows beside the other models', and Molmo2-8B has no row in the two background tables, its overlapping crops sharing no single token grid.
Their bases were probed as the others were (Table~\ref{tab:heldout}).

\textbf{Census and corruption.}\quad Two spike under the criterion and four do not.
Molmo2-8B spikes on every census image and Pixtral-12B on all but one, and the four others on none, with peaks below $240$.
Noise does not change their incidence because both already spike on every forward.
Their locations need not be stable, however. Across the six corruptions, Molmo2-8B relocates the spike on $96$ to $100\%$ of forwards and Pixtral-12B on $2$ to $20\%$ (Table~\ref{tab:corrpanel}).

\textbf{Where the two spike.}\quad Pixtral-12B spikes at the text switch of Mistral-NeMo-12B, block 2, the block that writes the base's text spike.
Its image token activates that block by direction, entering at cosine $0.6$ with the block's trigger, whereas the base's text token activates it by size (Section~\ref{sec:mech}).
The token sits at one position too, the second cell of the last row, on $96\%$ of images, and no image token aligns with a later block of that base.
Molmo2-8B spikes at block 16 of Qwen3-8B, the visual switch of Qwen3-VL-8B and InternVL3.5-8B. It spikes on several tokens per image, a median of seven at the peak state, written from $1.7$ to $8{,}192$ in the median at positions that change with the image.
It also hands the language model three image tokens already massive on the spike channel, at $45{,}000$ over a layer median below one, at the same three positions on every image, and they are massive on thousands of channels, the spike channel ranking fourth of $4{,}096$.
No block writes them, and their value moves by under $4\%$ through the 36 blocks.
Read on the blocks' outputs alone, the criterion picks them first, since the base's spike channel is among the thousands such a token carries.
They are ViT-emerged tokens in the partition of \citet{choi2026sinks}, not spike tokens (\aref{app:measure}), and every reading of Molmo2-8B is on its block-written spikes, which enter below $2$.
On the other eight spiking models the spike token enters with at most $14$ on the channel.
Pixtral-12B also hands in five to seven high-norm tokens at $100$ to $400$ on its channel on every image, each with a norm two hundred times an ordinary token's, and Molmo-7B-D and Janus-Pro-7B hand in a few at $100$ to $240$.
None of them is a spike token.

\textbf{Attention.}\quad Pixtral-12B's spike token is an attention sink on every spiking forward of the three tasks, and the top image token takes $0.52$ to $0.56$ of the image's attention in the median (Table~\ref{tab:attntasks}).
Molmo2-8B forms an attention sink on $24$ to $34\%$ of forwards, with a median top share of $0.20$ to $0.21$, and every sink it forms sits on a block-written spike token, never on a carried-in one.
Of the four that do not spike, PaliGemma 2-3B forms an attention sink on $47$ of its $300$ captioning forwards and the other three on at most two forwards per task.

\textbf{Attack and intervention.}\quad Every clean forward of these two models spikes, so only abolition applies.
Under 60 steps of $\nicefrac{\varepsilon}{4}$, Pixtral-12B's state goes on $32$ of $40$ images at one pixel level and on $39$ at two, and under 600 steps of $\nicefrac{0.25}{255}$ on all $40$ at one (Table~\ref{tab:epssweep}).
Molmo2-8B's does not move under that schedule, its peak going from $\Psi = 22{,}361$ to $17{,}351$ in the median, and it moves under the finer step of Table~\ref{tab:epssweep}.
Removing the trigger component at the switch (Table~\ref{tab:prevent}) takes Pixtral-12B's write at block 2 to zero and its peak from $\Psi = 17{,}270$ to $664$, under the level, on every one of the 200 questions, with the answer held on $93\%$ of them.
Its attention sink stays on every forward, so the two phenomena part under a manipulation of the activation alone.
On Molmo2-8B it removes $96\%$ of the write, $\Psi$ from $22{,}105$ to $874$, the spike from $55\%$ of forwards and the attention sink from every forward, with every answer unchanged.
Accuracy holds at $0.900$ on Molmo2-8B and rises from $0.770$ to $0.820$ on Pixtral-12B.

\textbf{Two readings to note.}\quad The largest-write column of Table~\ref{tab:heldout} maximizes Equation~\ref{eq:trigger} with each block's pre-feed-forward norm and activation included.
Gemma-2 and GLM-4 also normalize the feed-forward's output before the residual add, so their writes are read before that norm (Table~\ref{tab:rowcap}).
PaliGemma 2 attends over the image and the prefix in both directions, so its image-token states depend on the question, whereas the causal decoders' do not.

\FloatBarrier
\section{Limitations}
\label{app:limits}
\backto*{app:limits}
Our study is restricted to adapter-based LVLMs with a separately released text-only base that already carries a text spike. This restriction lets us separate what the model inherits from its base from what visual adaptation changes. It excludes natively multimodal models, for which we cannot make the same comparison.

We study image-conditioned generation with one image presented before the text. We do not test interleaved text and images, multiple images, or video, where later visual tokens are processed in the context of earlier visual and textual states. The analysis is also limited to vision and language; we do not examine massive activations in audio, speech, or other modalities even though the study includes an omni-model.

Our analysis explains how a trained model produces a visual spike, but not how training creates the trajectory that reaches the switch. A base can contain several blocks capable of writing a spike, yet we cannot predict from the base alone whether visual adaptation will activate one of them. The base and adaptation comparisons also do not isolate which component of the adaptation, the encoder, the projector, the data, or the training, makes the switch reachable. Likewise, the shared part restricts the eventual spike token before the language model runs, but we do not trace how the vision encoder, projector, or training objective produces this structure.

The empirical results are limited to the datasets, prompts, corruptions, and perturbation budgets studied here. The separation between spiking and non-spiking models is sharp across the 25 LVLMs under these measurements.
Our measurements do not prove that a non-spiking model can never spike, but they make it unlikely on natural images. Its image tokens reach none of its base's read blocks beyond the text switch, except in PaliGemma 2-3B (Table~\ref{tab:heldout}), and none of the fifteen non-spiking models spikes under Gaussian noise (Table~\ref{tab:noiseroster}). The split may also not hold for unseen architectures.
Some measurements rely on few naturally spiking images. InternVL3.5-38B spikes on five of its 300 census images, so the body names it beside its ranges rather than inside them and the tables mark its cells indicative. Up to 35 of 200 forwards change classification across attention implementations. The spike attack is white-box and requires access to the visual switch and trigger. It shows that small pixel changes can control the state under this threat model.
The intervention is evaluated on the saved attack images, not against an adaptive attack optimized through the intervention.
High answer preservation on POPE also does not establish invariance on every task, and free-form scores show larger variation in some cases.

The intervention strongly suppresses the spike with little change in task behaviour in eight of the ten spiking models. In Molmo2-8B and Pixtral-12B, however, free-form outputs change more, with captions unchanged on $53\%$ and $13\%$ of forwards against $79$ to $87\%$ in the other eight (Table~\ref{tab:preventfree}). We therefore cannot assume that the same intervention will preserve behaviour across models.

\FloatBarrier
\section{Future work}
\label{app:future}
\backto*{app:future}
One direction is to understand how training brings image tokens to an inherited spike trigger. The base supplies the switch and its trigger direction, while whether image-token trajectories reach them differs with the visual adaptation. Identifying the roles of the vision encoder, projector, multimodal data, and optimization may allow us to predict which adaptations will spike. The early shared-part signal raises a related question: what structure do the encoder and projector place in tokens that share little with the rest of the image, and why do these tokens later become spike candidates?

A second direction is robustness by design. Our intervention acts at inference time and leaves the trained weights unchanged. It suppresses the spike on the inputs tested here, but the underlying model remains susceptible to perturbations that create, move, or remove it. Training could instead aim to prevent small image changes from steering tokens into or out of the trigger while preserving the base model's useful computation.

Interleaved images, multi-image reasoning, and video offer further tests of the mechanism. Video stands out because repeated content across frames may affect where massive activations form, which tokens become attention sinks, and what these states do with redundant visual information. Natively multimodal models and other modalities may follow different trajectories altogether.

The functional role of visual spikes remains open. Removing them has little effect on behaviour in eight of the ten spiking models we study, but changes free-form outputs substantially in Molmo2-8B and Pixtral-12B. Understanding when a spike supports useful computation, and how that role relates to attention sinks, would clarify when preventing it is desirable.

\end{document}